\documentclass[]{bytedance_seed}

\usepackage[toc,page,header]{appendix}

\usepackage{minitoc}
\usepackage{solarized-light}
\usepackage{multirow}
\usepackage{graphicx}
\usepackage{array}
\usepackage{colortbl}
\usepackage{makecell}
\usepackage{framed}
\usepackage{hyperref}
\usepackage{amsmath}
\usepackage{amsfonts}
\usepackage{algorithm}
\usepackage{algorithmic}
\usepackage[colorinlistoftodos]{todonotes}
\usepackage{longtable}
\usepackage{hhline}
\usepackage{fancyvrb}
\usepackage{fvextra}
\usepackage{CJKutf8}
\usepackage{multicol}
\usepackage{cleveref}
\usepackage{tablefootnote}
\usepackage{threeparttable}
\usepackage{tabularx}
\usepackage{mdframed}
\usepackage{subcaption}
\usepackage{wrapfig}
\usepackage{pifont}
\usepackage[usestackEOL]{stackengine}
\usepackage[numbers]{natbib}
\usepackage{bibunits}
\newif\ifseparateappendixbibliography
\separateappendixbibliographyfalse % Use \separateappendixbibliographytrue for separate reference lists.
\newcommand{\commentout}[1]{}
\renewcommand{\paragraph}[1]{\noindent\textbf{#1.}\hspace*{1em}}
\renewcommand\affiliation[2][]{\addtolist[#1]{#2}{\affiliationlist}{\affiliationformat}{\qquad}}
\usepackage{enumitem}
\setlist[itemize,enumerate]{label=$\diamond$,leftmargin=15pt}

\RequirePackage{xspace}
\makeatletter
\DeclareRobustCommand\onedot{\futurelet\@let@token\@onedot}
\def\@onedot{\ifx\@let@token.\else.\null\fi\xspace}

\def\eg{\emph{e.g}\onedot}

\makeatother

\title{MIRA: Medical Image Reflection for Agentic Diagnosis}

\author[1]{Shengzhi Wang}
\author[5]{Jun Yang}
\author[1]{Kai Wu}
\author[{2,\spadesuit}]{Xiaozhong Ji}
\author[3]{Yiwen Ye}
\author[3]{Ziyang Chen}
\author[1]{Mingliang Xiong}
\author[1]{Wen Fang}
\author[1]{Mingqing Liu}
\author[1]{Mengyuan Xu}
\author[4]{Miaoxuan Shan}
\author[1]{Caiyan Liu}
\author[1]{Bin He}
\author[{1,\dagger}]{Qingwen Liu}

\affiliation[1]{Tongji University}
\affiliation[2]{Nanjing University}
\affiliation[3]{Northwestern Polytechnical University}
\affiliation[4]{People's Public Security University of China}
\affiliation[5]{École Normale Supérieure -- PSL}

\abstract{
Recent advances in ``thinking with images'' demonstrate that visual information can be actively leveraged during multimodal reasoning. However, indiscriminate tool use may introduce noisy observations, propagate misleading evidence, and increase reasoning cost, making it crucial for medical agents to decide when external tools are actually necessary. In medical diagnosis, where errors can be fatal, merely acquiring additional observations is therefore insufficient; a reliable agent should verify whether tool actions are appropriate, whether the acquired evidence supports the current hypothesis, and whether weak or contradictory evidence calls for corrective reasoning. In this paper, we introduce MIRA (\textbf{M}edical \textbf{I}mage \textbf{R}eflection for \textbf{A}gentic Diagnosis), a medical visual diagnostic framework that enables diagnosis-oriented reasoning through autonomous evidence search and reflective verification. MIRA dynamically executes image-processing operations (e.g., zooming, grounding, and pointing), searches the web, and verifies tool use and reasoning against visual or retrieved evidence. To equip the model with this capability, we design a two-stage training strategy. First, Supervised Fine-Tuning (SFT) on a carefully curated medical tool-use and reflection dataset is conducted to teach the model tool use and reflection abilities; subsequently, Reinforcement Learning (RL) is applied to further optimize the model's decision-making. Specifically, we design a tool-augmented Monte Carlo Tree Search (MCTS) data engine for SFT training, which systematically explores the reasoning space by generating diverse diagnostic hypotheses, and performs joint verification of visual grounding accuracy and semantic reasoning consistency at each node during tree expansion. In the RL stage, we propose an online reflective principle evolution mechanism: failure cases are automatically distilled into reflective principles and injected into subsequent rollouts, with effective principles selected through a trial-and-verification loop, providing global guidance for tool use and reflection. Experiments across multiple medical visual reasoning benchmarks show that MIRA consistently improves over its Qwen3-VL-8B backbone and achieves competitive performance among strong open-source and medical-specific VLMs, while narrowing the gap to larger closed-source systems. Qualitative analyses further demonstrate that MIRA learns to re-examine visual evidence, correct premature conclusions, and adapt tool-use strategies, suggesting that evidence-grounded verification and corrective tool use can improve the reliability of medical visual diagnostic agents. Project page: \url{https://MIRA-VL.github.io/}.
}

\date{\today}

\begin{document}
\maketitle

\newpage
{\hypersetup{linkcolor=[HTML]{4C4C4C}}\tableofcontents}
\newpage

\ifseparateappendixbibliography
\begin{bibunit}[plainnat]
\fi
\section{Introduction}
\label{sec:intro}

\begin{wrapfigure}{r}{0.40\linewidth}
  \centering
  \vspace{-0.8em}
  \includegraphics[width=\linewidth]{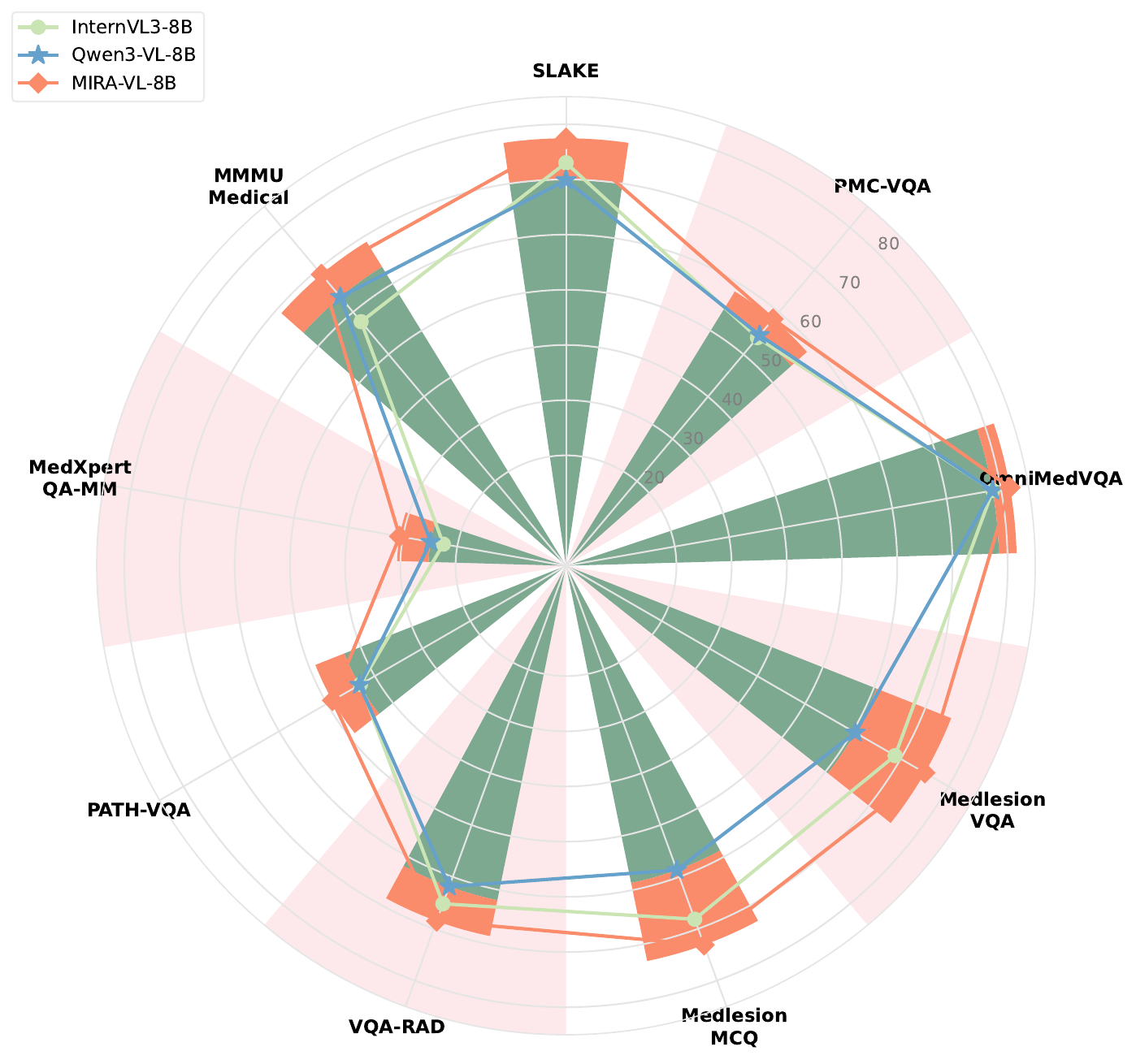}
  \caption{Benchmark performance of MIRA. The agentic medical image reflection framework enables MIRA to achieve consistent improvements over the Qwen3-VL-8B backbone across diverse medical VQA benchmarks. By leveraging active tool-use, evidence-grounded verification, and reflection-driven self-correction, MIRA shows clear gains on perception-intensive lesion tasks and expert-level medical reasoning benchmarks. The improvements across broad medical VQA, lesion VQA, radiology, pathology, and diagnostic reasoning tasks further validate the effectiveness of our two-stage training approach.}
  \label{fig:intro_radar}
  \vspace{-1.0em}
\end{wrapfigure}

Recent years have witnessed considerable interest in enabling medical Large Vision-Language Models (LVLMs) to reason beyond direct image-to-answer prediction~\citep{llavamed2023,moor2023med,medpalm2023,medgemini2024,chen2024huatuogptvisioninjectingmedicalvisual,lu2024multimodal,lingshu2024}. Existing approaches to improve medical multimodal reasoning generally fall into two categories. One category focuses on language-level reasoning, where supervised fine-tuning or reinforcement learning is used to elicit explicit chain-of-thought trajectories for diagnosis~\citep{lai2025medr1reinforcementlearninggeneralizable,pan2025medvlm,mullappilly2026medix}. These methods improve the structure and interpretability of textual rationales, but the visual observation usually remains fixed after the image is encoded once. As a result, the model may produce fluent reasoning while missing subtle lesions, attending to irrelevant regions, or hallucinating unsupported visual findings~\citep{gu2026medvh}. The other category enables models to think with images, allowing visual information to be dynamically incorporated into the reasoning process through localized inspection or grounding operations~\citep{wu2024v,su2025openthinkimg,wang2025pixel,zheng2025deepeyes,hong2025deepeyesv2,wu2025vtool}. In medicine, recent tool-augmented and grounding-oriented agents further adapt this paradigm to clinical image analysis, encouraging models to seek additional visual evidence rather than relying solely on a single static observation~\citep{jiang2025incentivizing,yan2026medreasoner,deria2026medmo,shi2026medxiaohe,chen2026meissa,li2026medscope,lu2026medvistagym,liu2026medsam}. Yet more tool use is not necessarily better: unnecessary or poorly targeted tool calls can introduce noisy observations, distract the reasoning process, and increase inference cost. Thus, an effective medical visual agent must not only know how to use tools, but also decide when tool use is warranted and whether the acquired evidence truly supports the current diagnostic hypothesis. Beyond acquiring extra observations, medical diagnosis also requires judging whether newly acquired evidence is reliable, whether it supports or contradicts the current hypothesis, and whether the diagnostic path should be revised. In this work, we use reflection in an operational sense: the model verifies tool actions and intermediate conclusions against visual or retrieved evidence, identifies unreliable reasoning paths, and seeks corrective evidence before finalizing or revising a diagnosis.

To address these challenges, we propose MIRA (\textbf{M}edical \textbf{I}mage \textbf{R}eflection for \textbf{A}gentic Diagnosis), a medical visual diagnostic framework that converts static medical VQA into an iterative process of evidence seeking, tool-use verification, and self-correction. MIRA is guided by four core principles:
\begin{itemize}
    \item \textbf{Evidence-grounded visual interaction:} MIRA uses a lightweight tool suite, including \texttt{ZOOM}, \texttt{GROUNDING}, \texttt{POINT}, \texttt{ROTATE}, \texttt{MEASURE}, and \texttt{SEARCH}, to actively inspect regions, mark spatial evidence, quantify visual relations, and retrieve external medical knowledge.
    \item \textbf{Tool-use reliability:} Instead of treating tool calls as automatically trustworthy, MIRA emphasizes whether the tool arguments, selected regions, and returned observations are visually plausible and diagnostically useful.
    \item \textbf{Failure-aware reflection:} MIRA learns to identify weak evidence, wrong tool use, and over-confident reasoning, and to acquire corrective evidence before producing or revising a diagnosis.
    \item \textbf{Continual principle evolution:} MIRA further summarizes recent failed rollouts into reusable diagnostic principles and accepts such principles only when they improve held-out validation reward.
\end{itemize}

To realize this vision, our technical roadmap comprises the following key components:
\begin{itemize}
    \item \textbf{Medical instruction data:} We first build a diverse medical instruction corpus from the training splits of four public medical VQA datasets, including PMC-VQA~\citep{pmcvqa2021}, SLAKE~\citep{slake2021}, VQA-RAD~\citep{vqarad2018}, and PathVQA~\citep{pathvqa2020}, together with an in-house high-quality medical image dataset collected from Internet sources. For all public datasets that are also used in evaluation, we strictly use only their training splits for data construction, trajectory synthesis, reflection construction, and model training, while the held-out test splits are reserved exclusively for final evaluation and are never used for training, validation, hyperparameter selection, or checkpoint selection.
    \item \textbf{MCTS-guided tool-use trajectory construction:} Since static VQA supervision does not teach a model when to inspect, where to look, or how to verify visual evidence, we convert each medical QA instance into an explicit evidence-seeking process using a tool-augmented Monte Carlo Tree Search (MCTS) data engine. Given an image and question, MCTS explores diverse diagnostic hypotheses, expands possible tool-use actions, verifies spatial actions by rendering proposed boxes or points on the image, and scores intermediate states according to both visual plausibility and semantic usefulness. This process yields answer-correct, tool-valid, and evidence-aware trajectories for cold-start supervised fine-tuning.
    \item \textbf{Reflection data construction:} Beyond successful trajectories, we further synthesize reflection supervision from both failed and successful search branches. For failed or unreliable trajectories, a teacher model identifies unstable observations, weak evidence, or reasoning leaps, and proposes corrective evidence-gathering actions. For successful trajectories, we refine intention-level thoughts so that each tool call is motivated by a clear diagnostic goal and interpreted according to the actual tool output. This data teaches the model not only how to use tools, but also how to question and revise its own diagnostic path.
    \item \textbf{On-policy RL and reflection memory optimization:} Finally, we optimize MIRA with GRPO on fresh multi-turn tool-augmented rollouts. We design a composite reward that evaluates answer correctness, output format, and consistency between the final diagnosis and the preceding evidence. To make failures reusable across cases, we introduce a validation-gated reflection memory that distills repeated failed rollouts into generalizable diagnostic principles and accepts them only when they improve held-out validation reward. This encourages selective rather than excessive tool use, preserving evidence-seeking behavior when additional information is needed while avoiding unnecessary noisy interactions.
\end{itemize}

MIRA effectively fulfills our vision of medical image reflection by combining active evidence acquisition with explicit verification and failure-aware self-correction. It can assess whether additional visual or external evidence is needed, select suitable tools, inspect fine-grained regions, verify spatial claims, and revise diagnostic hypotheses when the accumulated evidence is insufficient or contradictory. Moreover, MCTS-guided data construction and reflection supervision provide high-quality tool-use and correction trajectories for cold-start training, while the validation-gated reflection memory turns repeated failures into reusable diagnostic principles during on-policy RL. Through approximately \textbf{1200} GPU hours of RL training, MIRA further strengthens its reflection and tool-use abilities, yielding more reliable evidence-grounded diagnostic behavior. We conduct comprehensive evaluations across multiple medical visual reasoning benchmarks covering both general medical VQA and more challenging diagnostic settings. As summarized in Figure~\ref{fig:intro_radar}, experimental results demonstrate that MIRA substantially improves over its base LVLM, performs competitively among strong open-source and medical-specific models, and reduces the performance gap to larger closed-source systems. Qualitative analyses further show transparent evidence-seeking behaviors, reliable tool-grounded reasoning, and image-grounded correction of premature conclusions. Furthermore, the constructed trajectory data, tool environment, training recipes, and model checkpoints have been released to facilitate future research on reliable medical visual agents.

\section{MIRA-SFT Cold Start}
\label{sec:method}

In this section, we introduce the tool design, data composition, data construction, and training strategies employed for the cold start of MIRA.

\subsection{Tool Design}
\label{subsec:tool_design}

\paragraph{Motivation}
Our tool design is intentionally lightweight. Rather than relying on heavy external vision modules~\citep{zhu2024medical,ravi2025sam,liu2024grounding} to solve medical perception on behalf of the model, we provide a compact set of basic visual operations that encourages the LVLM to actively decide where to inspect, what evidence to collect, and how to verify its own hypotheses. This design aims to stimulate the model's intrinsic visual exploration ability while keeping the reasoning process transparent and auditable.

\paragraph{Available Tools for Image Analysis}
Based on this motivation, we equip MIRA with six tools that support active medical image inspection and external evidence acquisition:
\begin{itemize}
    \item \textbf{SEARCH}, a non-visual knowledge tool that retrieves and summarizes relevant medical evidence from external sources. It is used when internal model knowledge is insufficient or when clinical definitions, differential diagnoses, or disease-specific cues need to be verified.
    \item \textbf{GROUNDING}, a bounding-box-based localization tool. Given one or more model-proposed regions, it draws boxes on the image or crops a selected region, enabling the model to explicitly verify suspected lesions, anatomical structures, or distributed abnormalities.
    \item \textbf{POINT}, a point-rendering tool for marking fine-grained visual evidence. It can annotate landmarks, connect ordered points into contours, and make spatial claims such as lesion boundaries or anatomical positions visually checkable.
    \item \textbf{ZOOM}, a fine-grained inspection tool that enlarges either the whole image or a selected bounding-box region. It adaptively expands small crops to preserve context, allowing the model to examine subtle morphology while retaining surrounding visual cues.
    \item \textbf{ROTATE}, an orientation adjustment tool that rotates medical images when acquisition or display direction is non-standard, making anatomical axes and visual details easier to interpret.
    \item \textbf{MEASURE}, a relative measurement tool that compares a target segment with a reference segment. It returns the target length, reference length, and their ratio, supporting quantitative judgments when absolute physical calibration is unavailable.
\end{itemize}
All tools are exposed through a unified function-calling interface. The model emits JSON-formatted tool arguments, and the executor returns either an updated image, textual evidence, or both. Spatial tools follow a normalized $0$--$999$ coordinate convention, which is converted to pixel coordinates during execution. More implementation details and tool schemas are provided in Appendix~\ref{sec:tool_use_details}.

\subsection{Golden Tool-Use Trajectory Synthesis}
\label{subsec:golden_tool_use_trajectory_synthesis}

\paragraph{Medical instruction data}
Our medical instruction data is constructed from the training splits of four public medical VQA datasets, including PMC-VQA~\citep{pmcvqa2021}, SLAKE~\citep{slake2021}, VQA-RAD~\citep{vqarad2018}, and PathVQA~\citep{pathvqa2020}, together with an in-house high-quality medical image dataset collected from Internet sources. For all public datasets that are also used in evaluation, we strictly use only their training splits for data construction, trajectory synthesis, reflection construction, and model training. The held-out test splits are reserved exclusively for final evaluation and are never used for training, validation, hyperparameter selection, or checkpoint selection.

The collected datasets provide standard image-question-answer triples, but they do not contain tool-use trajectories. Directly fine-tuning on such static VQA data cannot teach the model tool-augmented diagnostic reasoning. Therefore, we synthesize golden tool-use trajectories that convert each static VQA instance into an explicit evidence-seeking diagnostic process.

To bypass the cold-start exploration problem, we employ frozen, highly capable teacher LVLMs and utilize Monte Carlo Tree Search (MCTS)~\citep{kocsis2006bandit} to systematically explore the diagnostic reasoning space. In our implementation, each instance is searched with 20 simulations, a maximum depth of 4, a branching factor of 3, and an exploration constant of $c_{\mathrm{puct}}=1.4$. We use a cascaded teacher strategy: Seed 1.8 is used first, while unresolved samples are subsequently retried with Gemini 2.5 Pro and GPT-5.2 to improve coverage and trajectory quality. As illustrated in Figure~\ref{fig:mcts}, we construct reasoning trees for medical visual queries and extract successful trajectories as golden tool-use supervision for the base model.

We formalize tool-use trajectory synthesis as a tree search process. Given an image $I$ and a question $Q$, each node $n$ represents the accumulated interaction history, including thoughts, tool calls, tool observations, and generated images. Expanding a node asks the teacher LVLM to propose the next diagnostic step, which can be either a tool call or a final answer. During search, MCTS selects the next node according to a UCB-style score:
\begin{equation}
    n^*=\arg\max_{n\in\mathcal{C}(p)}\left(Q(n)+c_{puct}\sqrt{\frac{\log(N(p)+1)}{N(n)}}\right),
\end{equation}
where $\mathcal{C}(p)$ denotes the valid children of parent node $p$, $Q(n)$ is the accumulated value of node $n$, and $N(\cdot)$ denotes visit count. After expansion, non-terminal tool-use steps are scored by a unified verifier, while terminal answers are scored by a judge against the ground-truth answer. The resulting reward is backpropagated to update the values of all nodes along the selected path.

During data construction, we encounter several practical challenges. To address them, we design the following MCTS strategies:
\begin{itemize}
    \item \textbf{Diverse clinical exploration.} Naive rollouts often collapse into a single reasoning direction or repeatedly inspect the same region. We therefore introduce a teacher-generated thought-spark mechanism, which proposes multiple clinically plausible directions based on the current path, sibling attempts, and failed branches. This encourages the search to explore different diagnostic hypotheses instead of following a single linear trajectory.
    \item \textbf{Tool-call validity verification.} LVLM-generated tool calls may contain wrong regions, invalid coordinates, or visually irrelevant actions. We address this by using a unified verifier that renders proposed boxes or points on the image, checks whether the action is visually plausible and diagnostically useful, and prunes impossible actions before they enter the tree.
    \item \textbf{Dense intermediate supervision.} Relying only on final-answer correctness produces sparse feedback, especially for multi-step diagnostic trajectories. We therefore use the verifier score as an intermediate reward for non-terminal nodes and combine it with the final judge score for terminal nodes. This guides the search toward useful evidence-gathering actions even before a final answer is reached.
    \item \textbf{Post-tool evidence grounding.} In multi-branch search, the model may ignore long tool responses, especially textual evidence returned by \texttt{SEARCH}. We explicitly carry forward tool outputs and require the teacher LVLM to summarize the newly acquired evidence after each executed tool. This grounding step ensures that subsequent expansions are conditioned on the actual tool observation rather than the previous hypothesis alone.
\end{itemize}
These designs enable MCTS to synthesize golden trajectories that are answer-correct, tool-valid, evidence-aware, and suitable for supervised cold-start training. More implementation details are provided in Appendix~\ref{sec:mcts_details}.

\subsection{Reflection Thinking Pattern Synthesis}
\label{subsec:reflection_thinking_pattern_synthesis}
Golden trajectories show how to reach the correct diagnosis with valid tools. They mainly cover successful paths. In practice, a medical visual agent also needs to know when a path is unreliable. It should notice weak visual evidence, wrong tool use, missing evidence, and early answers. To teach this ability, we build two types of reflection data: failure-driven correction data and textual intention data. As shown in Figure~\ref{fig:mcts_train_reflection}, the reflection data construction process converts unreliable or incomplete diagnostic paths into evidence-grounded correction samples. The first type is built from wrong MCTS trajectories. We ask the teacher to reflect on the unreliable observations or reasoning leaps in the trajectory and propose a corrective evidence-gathering action. The second type is built from golden trajectories. We ask the teacher to check whether each thought matches the next action and the returned evidence, and to rewrite unclear thoughts when needed.

\begin{figure*}[t]
  \centering
  \includegraphics[trim=40 170 40 130, clip, width=\linewidth]{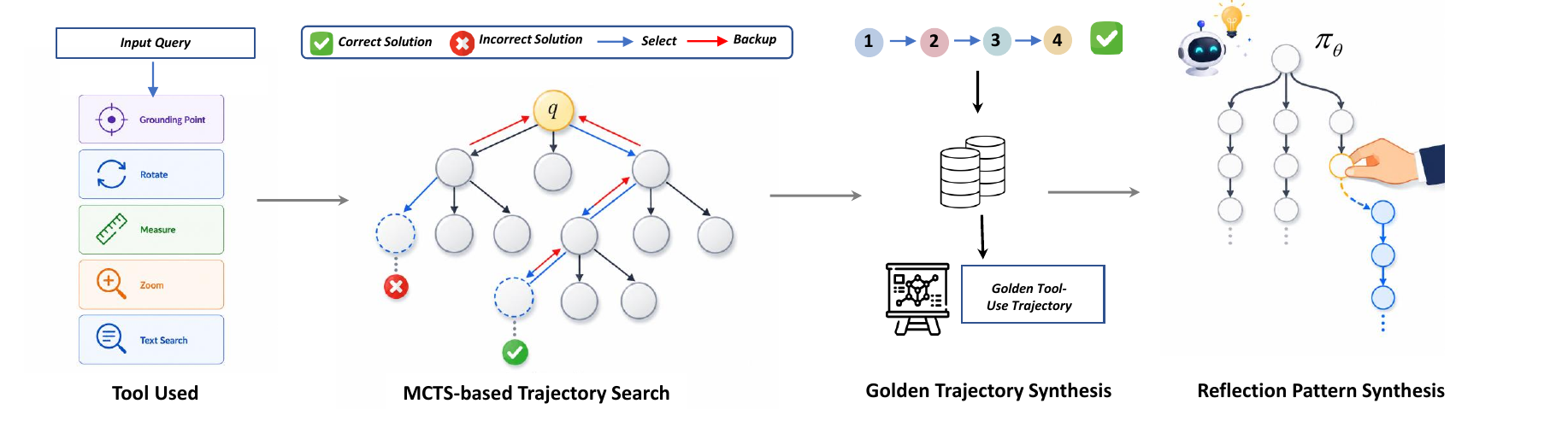}
  \caption{Reflection thinking pattern synthesis. The pipeline uses MCTS trajectories to construct both failure-driven correction data and textual intention reflection data, enabling the model to identify unreliable diagnostic paths, acquire corrective evidence, and ground revised reasoning in actual tool observations.}
  \label{fig:mcts_train_reflection}
\end{figure*}

\paragraph{Failure-driven correction reflection}
Given an image-question pair $(I,Q)$, a path in the MCTS tree is written as
\begin{equation}
    \tau=(I,Q,o_0,t_1,a_1,o_1,\ldots,t_L,a_L,o_L,y),
\end{equation}
where $t_i$ is the thought at step $i$, $a_i$ is a tool call or final answer, $o_i$ is the tool result or text state, and $y$ is the predicted answer. We collect trajectories that are judged to be wrong or infeasible at terminal nodes. Specifically, a trajectory is treated as a wrong trajectory when its final answer fails verification, the terminal state is marked as impossible, or the final judge rejects the answer:
\begin{equation}
    \mathcal{D}_{\mathrm{wrong}}
    = \{(I,Q,\tau^{-},y,\hat{y}) \mid \mathrm{Judge}(\tau^{-},\hat{y})=\mathrm{wrong}\},
\end{equation}
where $\hat{y}$ is the reference answer. To obtain more informative correction samples, we keep wrong trajectories that contain tool use and rank them by search scores, value estimates, neighboring-branch quality, trajectory length, and tool-use statistics. This selection prioritizes hard negative trajectories that appear plausible during search but eventually lead to an incorrect diagnosis.

For each selected wrong trajectory $\tau^{-}$, we construct answer-conditioned reflection data. The reference answer is used to determine the correction direction. A teacher LVLM (e.g., GPT-5.2 or Gemini 2.5 Pro) first receives the image, question, reference answer, and a summarized wrong trajectory, and then generates a corrective reflection and the next corrective action:
\begin{equation}
    (r,t^{+},a^{+},\bar{o}^{+})
    = \pi_{\mathrm{tea}}^{(1)}(P_{\mathrm{draft}}, I,Q,\tau^{-},\hat{y}),
\end{equation}
where $r$ is a natural-language reflection that identifies unstable observations or over-confident reasoning in the wrong trajectory, $t^{+}$ is the corrected next thought, $a^{+}$ is the corrective action, and $\bar{o}^{+}$ describes the type of evidence expected from this action. The corrective action usually corresponds to a new evidence-acquisition step, such as rechecking a key image region or retrieving relevant medical knowledge when needed.

We then execute the proposed corrective action to obtain a real observation:
\begin{equation}
    o^{+}=\mathcal{T}(a^{+};I),
\end{equation}
where $\mathcal{T}$ denotes the corresponding tool execution process. In the second stage, the teacher LVLM generates the final training target conditioned on the actual tool result:
\begin{equation}
    \mathcal{D}_{\mathrm{corr}}
    = \pi_{\mathrm{tea}}^{(2)}(P_{\mathrm{final}}, I,Q,\tau^{-},\hat{y},r,t^{+},a^{+},o^{+}).
\end{equation}
Each final sample contains the reflection, corrected next thought, executed corrective action, visual evidence, optional knowledge evidence, an evidence-sufficiency flag, a tool-grounded evidence summary, and the corrected final answer:
\begin{equation}
    \mathcal{D}_{\mathrm{corr}}
    = \{(I,Q,\tau^{-},r,t^{+},a^{+},o^{+},e_{\mathrm{vis}},e_{\mathrm{know}},s,\hat{y})\}.
\end{equation}
Here $e_{\mathrm{vis}}$ describes evidence supported by the image or visual observations, $e_{\mathrm{know}}$ denotes optional external knowledge evidence, and $s$ indicates whether the current evidence is sufficient to support the corrected answer. When the evidence is insufficient, the sample preserves this uncertainty rather than forcing the observation into a definitive conclusion. This data teaches the model to recognize unreliable diagnostic paths, actively acquire corrective evidence, and ground its revised answer in actual observations.

\paragraph{Textual intention reflection}
We further check thoughts in golden trajectories. A thought may not explain why a tool is needed, may not match the tool arguments, or may draw a conclusion before the tool result is checked. Given a golden trajectory
\begin{equation}
    \tau^{+}=(I,Q,o_0,t_1,a_1,o_1,\ldots,t_L,a_L,o_L,\hat{y}),
\end{equation}
we ask a teacher LVLM (e.g., GPT-5.2 or Gemini 2.5 Pro) to inspect each thought-action pair $(t_i,a_i)$ under the previous context $\tau^{+}_{1:i-1}$ and the returned observation $o_i$. The teacher checks whether $t_i$ explains the purpose of $a_i$, matches the tool arguments, and is supported by the available evidence. If a logic gap is found, the teacher rewrites the thought while keeping the valid action unchanged. Otherwise, the thought is kept unchanged. This gives an intention-level dataset:
\begin{equation}
    \mathcal{D}_{\mathrm{int}}
    =\{(I,Q,\tau^{+}_{1:i-1}, t_i^{-}, a_i, o_i, t_i^{+})\},
\end{equation}
where $t_i^{-}$ is the original thought and $t_i^{+}$ is the checked or corrected thought. This data teaches the model to state a clear diagnostic goal before using a tool or giving an answer.

Based on these two datasets, the reflection SFT objective is
\begin{equation}
\begin{aligned}
    \mathcal{L}_{\mathrm{ref\text{-}sft}}
    = &-\mathbb{E}_{\mathcal{D}_{\mathrm{corr}}}
    \log \pi_{\theta}(r,t^{+},a^{+},e_{\mathrm{vis}},e_{\mathrm{know}},s,\hat{y}\mid I,Q,\tau^{-}) \\
    &-\lambda_{\mathrm{int}}\mathbb{E}_{\mathcal{D}_{\mathrm{int}}}
    \log \pi_{\theta}(t_i^{+},a_i\mid I,Q,\tau^{+}_{1:i-1}).
\end{aligned}
\end{equation}
Together with the golden tool-use data, these reflection samples teach the model to check evidence, revise weak paths, and act with a clear goal.

% \paragraph{Supervised Fine-Tuning with Reflection}
% After the MCTS process completes, we extract the explored trajectories to construct our SFT dataset. Rather than merely cloning the optimal paths, our strategy extracts both the successful diagnostic trajectories and the erroneous, abandoned paths. These incorrect paths are explicitly formatted as reflection data (e.g., recognizing a failed tool call and proposing a correction). By training on both positive demonstrations and negative reflections, the base LVLM internalizes the fundamental syntax of hypothesis-verification-backtracking. This equips the model with basic self-correction capabilities, providing a robust policy initialization for the subsequent reinforcement learning stage. Detailed examples of the constructed data format can be found in Appendix~\ref{sec:appendix_qualitative_cases}.
\section{MIRA-RL}
\label{sec:mira_rl}

\begin{wrapfigure}{r}{0.48\textwidth}
  \centering
  \vspace{-8pt}
  \includegraphics[trim=300 190 400 130, clip, width=0.47\textwidth]{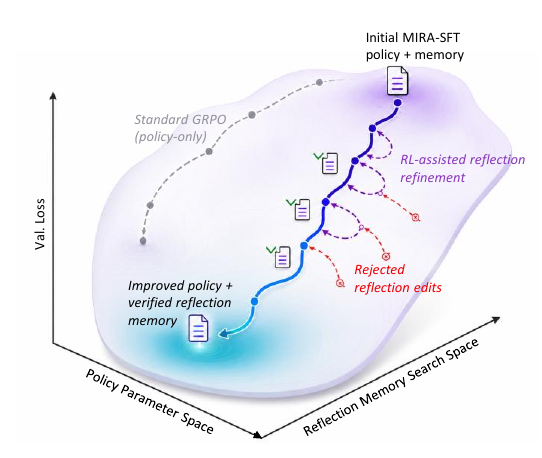}
  \caption{On-Policy Reflection Memory Optimization. The model solves tasks with the current diagnostic memory. Failed rollouts are summarized into candidate memory edits. A validation set then tests these edits. Only edits that improve the reward are kept, and rejected edits are used as feedback for later updates.}
  \label{fig:rl_training}
  \vspace{-10pt}
\end{wrapfigure}

We further optimize MIRA-SFT with online reinforcement learning and carefully designed rewards. This helps the agent learn when and how to use tools, combine multiple tools, and move beyond simply imitating the fixed trajectories used in SFT. MIRA-RL has three main components: on-policy GRPO training~\citep{shao2024deepseekmath}, a composite reward for answer accuracy and trajectory quality, and a validation-gated memory update that turns recent failures into reusable diagnostic principles.

% \paragraph{Relation to Prior Work}
% Natural-language reflection and experience memory have been explored for language agents: Reflexion~\citep{shinn2023reflexion} maintains instance-level verbal feedback that does not enter policy-gradient training, ExpeL~\citep{zhao2023expel} extracts experience offline from expert trajectories, and prompt-space optimization methods such as OPRO~\citep{yang2023opro}, Automatic Prompt Engineer~\citep{pryzant2023automatic}, and TextGrad~\citep{yuksekgonul2024textgrad} optimize prompts through score-guided search or natural-language critiques but do not couple this search with on-policy RL. MIRA-RL differs in three respects: (i) the memory is a single cross-instance state refined online during GRPO, rather than per-instance feedback or an offline-optimized prompt; (ii) memory proposals are generated from multi-turn medical tool-use failures rather than text-only reasoning errors; and (iii) candidate memories are filtered by a frozen-policy validation gate that is disjoint from the rollout distribution used for policy updates. We also emphasize that the RL memory differs from the instance-level corrective trajectories constructed during MIRA-SFT: the latter are supervised training examples for individual cases, whereas the former is a compact, cross-instance prompt state shared across subsequent RL rollouts.

\subsection{Preliminary of GRPO Training}
\label{subsec:preliminary_of_grpo_training}
MIRA-RL uses GRPO~\citep{shao2024deepseekmath} to update a policy that can call visual tools during reasoning. For each image-question pair $(I,Q)$, the rollout policy samples a group of multi-turn trajectories
\begin{equation}
\mathcal{T}_{I,Q}^{t}=\{\tau_i\}_{i=1}^{G},\qquad
\tau_i\sim\pi_{\theta_t^{-}}(\cdot\mid I,Q,m_t).
\end{equation}
Each trajectory contains model thoughts, optional \texttt{<tool\_call>} actions, returned \texttt{<tool\_response>} results, and a final answer. When the model emits a tool call, we parse its arguments, execute the visual tool, and append the result to the context. Tool responses are not generated by the model, so they are masked from the policy loss. The rollout stops when the model gives an answer, reaches the tool-call limit, or repeats a previous tool call.

Let $\mathcal{Y}_i$ be the model-generated tokens in $\tau_i$, including reasoning, tool-call, and answer tokens. The policy is updated on the whole trajectory group:
\begin{equation}
\theta_{t+1}=\arg\max_{\theta}\;\mathbb{E}_{(I,Q),\mathcal{T}_{I,Q}^{t}}
\left[\frac{1}{G}\sum_{i=1}^{G}\frac{1}{|\mathcal{Y}_i|}\sum_{j\in\mathcal{Y}_i}
\left(\min\left(\rho_{i,j}\hat{A}_i,\mathrm{clip}(\rho_{i,j},1-\epsilon,1+\epsilon)\hat{A}_i\right)-\beta d_{i,j}^{\mathrm{KL}}\right)\right],
\end{equation}
where $\rho_{i,j}=\pi_{\theta}(y_{i,j}\mid c_{i,j})/\pi_{\theta_t^{-}}(y_{i,j}\mid c_{i,j})$. The advantage $\hat{A}_i$ is computed by normalizing the trajectory reward $r_i$ within the same group $\mathcal{T}_{I,Q}^{t}$. Thus, the group gives relative feedback: a trajectory is encouraged when its answer and tool-use process are better than other sampled trajectories for the same case. Fresh tool-augmented trajectories are collected at each update round, and old trajectories are not replayed.

\subsection{Composite Trajectory Reward}
\label{subsec:composite_trajectory_reward}
MIRA-RL evaluates both final-answer correctness and the reliability of the diagnostic trajectory. The composite reward contains three components. The formatting reward $R_{\mathrm{format}}\in\{0,1\}$ checks whether the model output follows the required \texttt{<think>...</think>} and \texttt{<answer>...</answer>} structure. The result reward $R_{\mathrm{result}}\in\{0,1\}$ evaluates final-answer correctness: multiple-choice predictions are first evaluated through rule-based option matching, and semantically flexible answers are evaluated by a fixed LLM judge against the reference answer.

The consistency reward $R_{\mathrm{cons}}\in[0,1]$ evaluates whether a correct final answer is supported by the preceding diagnostic process. A fixed judge pool examines the final answer, the tool-call and tool-response transcript, and the diagnostic principles applicable to the current case. Each judge returns binary decisions on the following criteria:
\begin{enumerate}
    \item whether the final answer is supported by the available visual or tool-derived evidence;
    \item whether the final answer is consistent with the preceding reasoning and action trajectory; and
    \item when an applicable memory principle is present, whether the trajectory follows rather than contradicts that principle.
\end{enumerate}
The consistency score is computed as the mean of the applicable binary criteria, with non-applicable criteria omitted. The judge models, evaluation prompt, decoding configuration, and output-parsing rules are fixed throughout training and are provided in Appendix~\ref{sec:eval_prompts}. To prevent an incorrect but internally coherent trajectory from receiving a large consistency bonus, the consistency term is gated by the result reward:
\begin{equation}
r_i=R_{\mathrm{result},i}\big(1+\lambda_c R_{\mathrm{cons},i}\big)+\lambda_f R_{\mathrm{format},i}.
\end{equation}
We use $\lambda_c=0.5$ and $\lambda_f=0.5$ as fixed reward-scaling coefficients in all experiments. Consequently, consistency can increase the reward only when the final answer is correct, while the formatting term provides a lightweight learning signal even for unsuccessful trajectories. The resulting reward ranges over $[0,2]$, with a correct answer contributing up to $1.5$ and format compliance contributing an additional $0.5$.

\subsection{On-Policy Reflection Memory Optimization}
\label{subsec:on_policy_reflection_memory_optimization}
Our motivation is that many medical tool-use failures are not isolated. They often reveal reusable mistakes, such as inspecting the wrong region, trusting a weak visual cue, or making a diagnosis before checking the tool evidence. Updating only the model weights can learn from these failures, but the signal is indirect and may require many rollouts. A short reflection memory provides a more direct path: it turns recent failures into explicit diagnostic principles that guide the next on-policy rollouts.

As shown in Figure~\ref{fig:rl_training}, the optimization follows a simple propose-and-test loop. After each rollout round, we collect trajectories that have the correct format but the wrong final answer. These failures form a buffer $\mathcal{B}_{\mathrm{fail}}^t$. We start a memory update only when the buffer contains at least $N_{\min}=4$ unique failed cases, so that the update is driven by repeated patterns rather than a single outlier. A reflection optimizer then reads a small batch of recent failures and summarizes what went wrong. It focuses on errors that can transfer across cases, such as missing a key region, over-trusting a tool result, or failing to re-check the visual evidence. Based on this summary, an editor makes a small update to the current memory $m_t$ and produces a candidate memory $\tilde{m}_t$. The edit budget is limited by $L_t$, which gradually decreases from $4$ to $1$ during training. This allows broader fixes early and smaller changes later. We also filter out malformed edits, repeated rules, output-format changes, and case-specific advice. Rejected edits are kept in a small rejection buffer, so the optimizer can avoid proposing the same harmful rule again.

The candidate memory is not accepted by default. After the GRPO policy update, we freeze $\theta_{t+1}$ and compare the current memory $m_t$ with the candidate memory $\tilde{m}_t$ on the same held-out validation set $\mathcal{D}_{\mathrm{val}}$. For a memory $m$, its validation score is the mean rollout reward under the frozen policy:
\begin{equation}
S_t(m)=\frac{1}{|\mathcal{D}_{\mathrm{val}}|}\sum_{x\in\mathcal{D}_{\mathrm{val}}}
\mathbb{E}_{\tau\sim\pi_{\theta_{t+1}}(\cdot\mid x,m)}[r(\tau)].
\end{equation}
We accept the candidate only if $S_t(\tilde{m}_t)>S_t(m_t)$. Because both memories are tested with the same policy and the same validation data, the comparison mainly measures the effect of the memory change. Accepted memories are saved with the corresponding policy checkpoint. Rejected edits and their score changes are stored as feedback for later proposal rounds. In this way, MIRA-RL alternates between updating the model policy with fresh tool-augmented rollouts and updating the reflection memory only when it improves held-out reward. Additional implementation details of the online reflection memory update are provided in Appendix~\ref{sec:online_reflection_memory_update_details}.

\begin{table*}[htbp]
  \centering
  \caption{Performance comparison on medical visual question answering benchmarks. Best results are \textbf{bolded} and second-best results are \underline{underlined}. Missing entries indicate that the corresponding result is not available. $\dagger$ indicates that Med-R1 is trained on part of the OmniMedVQA test set.}
  \label{tab:main-results}
  \footnotesize
  \setlength{\tabcolsep}{3.0pt}
  \renewcommand{\arraystretch}{1.08}
  \resizebox{\textwidth}{!}{
  \begin{tabular}{l c c c c c c c c c !{\vrule width 0.45pt} c}
    \toprule[1.2pt]
    \textbf{Model} & \makecell{\textbf{SLAKE}} & \makecell{\textbf{PMC}\\\textbf{VQA}} & \makecell{\textbf{Omni}\\\textbf{MedVQA}} & \makecell{\textbf{Medlesion}\\\textbf{VQA}} & \makecell{\textbf{Medlesion}\\\textbf{MCQ}} & \makecell{\textbf{VQA}\\\textbf{RAD}} & \makecell{\textbf{PATH}\\\textbf{VQA}} & \makecell{\textbf{MedXpert}\\\textbf{QA-MM}} & \makecell{\textbf{MMMU}\\\textbf{Medical}} & \textbf{AVG.} \\
    \midrule[0.8pt]
    \rowcolor[HTML]{FDF2EA}\multicolumn{11}{c}{Closed-Source Models} \\
%    Gemini-3.1-Pro & 80.04 & \textbf{70.80} & 83.20 & \textbf{80.25} & 71.58 & 71.18 & 58.54 & \textbf{78.50} & 82.40 & \textbf{75.17} \\
%    GPT-5.5 & 80.23 & \underline{69.60} & 78.85 & 77.57 & \underline{77.66} & \textbf{77.38} & 53.43 & \underline{76.00} & \textbf{84.80} & \underline{75.06} \\
%    Seed-2.0 & 79.18 & 67.10 & 71.90 & 72.78 & 75.33 & 71.18 & 59.71 & 60.95 & 80.93 & 71.01 \\
    Gemini-2.5-Pro & 74.88 & \textbf{66.30} & 71.99 & \underline{78.70} & 71.37 & 69.84 & 57.76 & \textbf{62.95} & \textbf{81.07} & \textbf{70.54} \\
    Seed-1.8 & 77.03 & 62.80 & 69.07 & \textbf{79.69} & \textbf{78.84} & 66.08 & 55.66 & 49.85 & 80.53 & \underline{68.84} \\
    GPT-5.2 & 75.36 & \underline{65.40} & 70.12 & 77.01 & 66.21 & 71.40 & 51.33 & \underline{50.20} & 71.47 & 66.50 \\
    GPT-4o & 70.92 & 60.70 & 70.21 & 76.02 & 73.06 & 64.75 & 53.71 & 39.60 & 66.13 & 63.90 \\
    \midrule[0.8pt]
    \rowcolor[HTML]{EEF6FD}\multicolumn{11}{c}{Open-Source Models} \\
    Qwen3-VL-235B-A22B & 78.47 & 60.45 & 81.47 & 71.36 & 61.79 & \textbf{75.27} & 46.83 & 34.95 & 71.13 & 64.64 \\
    InternVL3-78B & 74.02 & 58.15 & \underline{82.95} & 71.36 & \underline{73.43} & 65.52 & 48.34 & 28.40 & 67.67 & 63.31 \\
    Qwen3-VL-32B-Thinking & 74.69 & 59.05 & 75.50 & 68.69 & 63.43 & 56.10 & 39.04 & 41.90 & 80.40 & 62.09 \\
    Qwen3-VL-30B-A3B & 73.78 & 56.30 & 79.32 & 70.66 & 61.45 & 67.18 & 43.52 & 28.65 & 74.53 & 61.71 \\
    InternVL2.5-38B & 68.10 & 55.60 & 79.70 & 70.24 & 56.52 & 61.64 & 43.71 & 25.10 & 65.20 & 58.42 \\
    InternVL3-8B & 73.02 & 53.95 & 78.49 & 68.83 & 68.17 & 65.19 & 43.00 & 22.55 & 57.73 & 58.99 \\
    Qwen3-VL-8B-Thinking & 66.43 & 52.35 & 70.70 & 66.29 & 58.26 & 52.99 & 33.83 & 28.50 & 72.40 & 55.75 \\
    InternVL2.5-8B & 66.48 & 51.25 & 81.46 & 63.61 & 51.12 & 60.09 & 39.26 & 21.65 & 54.40 & 54.37 \\
    \midrule[0.8pt]
    \rowcolor[HTML]{E6E6E6}\multicolumn{11}{c}{Medical-Specific Models} \\
    Lingshu-32B & \textbf{83.34} & 61.85 & 79.79 & 58.08 & 51.18 & \underline{73.83} & \textbf{62.70} & 27.70 & 60.80 & 62.14 \\
    Lingshu-7B & \underline{80.38} & 59.10 & 81.10 & 60.34 & 46.28 & 64.52 & \underline{59.21} & 24.20 & \underline{80.93} & 61.78 \\
    Chiron-o1-8B & 76.22 & 58.90 & 80.69 & 66.43 & 68.19 & 64.75 & 46.76 & 22.25 & 57.87 & 60.23 \\
    HuatuoGPT-Vision-34B & 70.92 & 56.55 & 76.84 & 69.39 & 69.23 & 64.52 & 44.69 & 22.25 & 57.33 & 59.08 \\
    HuatuoGPT-Vision-7B & 66.38 & 54.90 & 75.44 & 67.98 & 66.32 & 65.85 & 43.24 & 21.35 & 51.33 & 56.98 \\
    Med-R1 & 54.63 & 44.70 & \textbf{91.34}$^{\dagger}$ & 58.67 & 58.87 & 45.68 & 33.38 & 21.00 & 39.87 & 49.79 \\
    MedVLM-R1 & 54.01 & 46.00 & 77.45 & 54.87 & 42.73 & 47.89 & 36.30 & 22.00 & 39.07 & 46.70 \\
    \midrule[0.8pt]
    Qwen3-VL-8B & 69.87 & 54.55 & 78.45 & 60.51 & 58.68 & 61.86 & 43.19 & 24.90 & 63.60 & 57.29 \\
%    MIRA-8B-SFT & 77.27\textsubscript{\textcolor[HTML]{D9822B}{+7.4}} & 55.95\textsubscript{\textcolor[HTML]{D9822B}{+1.4}} & 74.92\textsubscript{\textcolor[HTML]{22863A}{-3.5}} & 74.19\textsubscript{\textcolor[HTML]{D9822B}{+13.7}} & 64.30\textsubscript{\textcolor[HTML]{D9822B}{+5.6}} & 66.52\textsubscript{\textcolor[HTML]{D9822B}{+4.7}} & 48.55\textsubscript{\textcolor[HTML]{D9822B}{+5.4}} & 29.90\textsubscript{\textcolor[HTML]{D9822B}{+5.0}} & 67.47\textsubscript{\textcolor[HTML]{D9822B}{+3.9}} & 62.12\textsubscript{\textcolor[HTML]{D9822B}{+4.8}} \\
    MIRA-VL-8B & 77.46\textsubscript{\textcolor[HTML]{D9822B}{+7.6}} & 58.35\textsubscript{\textcolor[HTML]{D9822B}{+3.8}} & 81.65\textsubscript{\textcolor[HTML]{D9822B}{+3.2}} & 75.04\textsubscript{\textcolor[HTML]{D9822B}{+14.5}} & 73.12\textsubscript{\textcolor[HTML]{D9822B}{+14.4}} & 68.51\textsubscript{\textcolor[HTML]{D9822B}{+6.7}} & 48.83\textsubscript{\textcolor[HTML]{D9822B}{+5.6}} & 30.60\textsubscript{\textcolor[HTML]{D9822B}{+5.7}} & 68.93\textsubscript{\textcolor[HTML]{D9822B}{+5.3}} & 64.73\textsubscript{\textcolor[HTML]{D9822B}{+7.4}} \\
%    \midrule[0.8pt]
%    Qwen3-VL-32B & 75.84 & 60.35 & 78.47 & 70.80 & 62.93 & 68.29 & 48.77 & 31.00 & 69.60 & 62.89 \\
%    MIRA-32B-SFT & 79.42\textsubscript{\textcolor[HTML]{D9822B}{+3.6}} & 60.10\textsubscript{\textcolor[HTML]{D9822B}{-0.3}} & 72.07\textsubscript{\textcolor[HTML]{D9822B}{-6.4}} & 70.94\textsubscript{\textcolor[HTML]{D9822B}{+0.1}} & 74.55\textsubscript{\textcolor[HTML]{D9822B}{+11.6}} & 64.52\textsubscript{\textcolor[HTML]{22863A}{-3.8}} & 49.92\textsubscript{\textcolor[HTML]{D9822B}{+1.2}} & 40.15\textsubscript{\textcolor[HTML]{D9822B}{+9.2}} & 73.20\textsubscript{\textcolor[HTML]{D9822B}{+3.6}} & 64.99\textsubscript{\textcolor[HTML]{D9822B}{+2.1}} \\
    \bottomrule[1.2pt]
  \end{tabular}
  }
\end{table*}

\section{Experiments}
\label{sec:experiments}

\subsection{Experimental Setting}
% To evaluate the effectiveness of our proposed framework, we compare MIRA against various state-of-the-art vision-language models on medical visual reasoning tasks. We use Qwen3-VL-8B-Instruct as the backbone model. Training is conducted in two stages: cold-start supervised fine-tuning (SFT) followed by on-policy reinforcement learning (RL). The SFT stage uses the MCTS-generated golden tool-use data, the reflection SFT data, and static medical VQA data from LesionVQA, PathVQA, PMC-VQA, SLAKE, and VQA-RAD. The RL stage is initialized from the final SFT checkpoint and further trained on filtered RL data from LesionVQA, PMC-VQA, PathVQA, SLAKE, and VQA-RAD, with a held-out validation set used for periodic evaluation and validation-gated reflection memory updates.

\paragraph{Hyperparameters}
For SFT, we perform full-parameter fine-tuning with bfloat16 precision and DeepSpeed ZeRO-3. The vision encoder and aligner are frozen, while the language model is updated. We train for 3 epochs with a learning rate of $1 \times 10^{-5}$, warmup ratio 0.05, and maximum sequence length 12,288. For RL, we adopt GRPO with full-parameter updates, bfloat16 precision, DeepSpeed ZeRO-3, and vLLM-based rollout generation. We train for 750 steps, approximately one epoch, with learning rate $1 \times 10^{-6}$, cosine scheduling, 4 rollouts per prompt, maximum context length 32,768, maximum completion length 2,048, and maximum tool-use turns 5. The KL coefficient is set to 0.0, and the lower and upper GRPO clipping thresholds are 0.20 and 0.28, respectively.

We use Qwen3-VL-8B as the backbone model~\citep{bai2025qwen3}. The SFT stage is conducted on 40 A800 GPUs and requires approximately 113 GPU hours. The reinforcement learning stage is conducted on 32 NVIDIA H20 GPUs and requires approximately 1200 GPU hours.

\subsection{Benchmarks and Baselines}
\label{sec:benchmarks_baselines}

\paragraph{Benchmarks} We evaluate MIRA on nine medical visual question answering benchmarks reported in Table~\ref{tab:main-results}: SLAKE~\citep{slake2021}, PMC-VQA~\citep{pmcvqa2021}, OmniMedVQA~\citep{omnimedvqa2024}, MedLesionVQA and MedLesionMCQ~\citep{yumedlesionvqa}, VQA-RAD~\citep{vqarad2018}, PathVQA~\citep{pathvqa2020}, MedXpertQA-MM~\citep{Zuo2025MedXpertQA}, and MMMU-Medical~\citep{mmmu2024}. These benchmarks cover a broad range of medical visual reasoning scenarios, including general medical image question answering, radiology-oriented VQA, pathology VQA, lesion-level understanding, medical multiple-choice questions, and expert-level multimodal medical reasoning. More evaluation details are provided in Appendix~\ref{sec:eval_prompts}.

\paragraph{Baselines} We compare MIRA with a broad set of representative multimodal baselines in Table~\ref{tab:main-results}. The closed-source group includes strong general-purpose systems such as Gemini~\citep{google2025gemini25pro}, GPT~\citep{openai2024gpt4o,openai2025gpt52}, and Seed models~\citep{seed2025seed1_5vl}. The open-source group covers general multimodal models from the Qwen3-VL~\citep{bai2025qwen3} and InternVL~\citep{chen2024expanding} families, as well as medical-specific VLMs including Lingshu~\citep{lingshu2024}, HuatuoGPT-Vision~\citep{chen2024huatuogptvisioninjectingmedicalvisual}, and MedVLM-R1~\citep{pan2025medvlm}. We also report the original Qwen3-VL-8B backbone to enable a direct comparison with the final MIRA-VL-8B model. We denote the final RL-trained 8B model as MIRA-VL-8B; in ablations, MIRA-SFT refers to the SFT checkpoint before RL.

\subsection{Evaluation Results}
\label{sec:main_results}

\paragraph{Main results} Table~\ref{tab:main-results} presents a comprehensive comparison between MIRA and leading multimodal models across nine medical visual question answering benchmarks. Compared with the Qwen3-VL-8B backbone, MIRA-VL-8B improves the average score from 57.29 to 64.73, demonstrating that agentic reinforcement learning with reflection memory substantially strengthens medical visual reasoning beyond the base LVLM. The gains are especially pronounced on MedLesionVQA and MedLesionMCQ, where MIRA-VL-8B improves by 14.5 and 14.4 points, respectively, suggesting that active evidence acquisition and verification are particularly beneficial for fine-grained lesion understanding and diagnostic decision-making. MIRA-VL-8B also shows consistent improvements across the remaining benchmarks, including PMC-VQA, radiology, pathology, and expert-level reasoning settings such as MedXpertQA-MM and MMMU-Medical. Compared with larger open-source and medical-specific models, MIRA-VL-8B achieves competitive average performance among the open-source and medical-specific baselines in Table~\ref{tab:main-results} while using only an 8B backbone, matching or surpassing several substantially larger or specialized models. Meanwhile, strong closed-source models such as Gemini and GPT still obtain the best overall averages on several benchmarks, highlighting that scale and proprietary training data remain important. These results show that MIRA mainly improves evidence-grounded medical perception and diagnostic reasoning, reducing the gap between compact open-source LVLMs and much larger closed-source systems.

\begin{table*}[htbp]
  \centering
  \caption{Fine-grained comparison between MIRA-VL-8B and the Qwen3-VL-8B backbone on representative medical benchmarks. Scores are accuracies in percentages, and the improvement row reports absolute gains in percentage points.}
  \label{tab:fine-grained-improvements}
  \scriptsize
  \setlength{\tabcolsep}{3.2pt}
  \renewcommand{\arraystretch}{1.10}
  \resizebox{\textwidth}{!}{
  \begin{tabular}{l c c c c c c c c c c c c}
    \toprule[1.2pt]
    & \multicolumn{3}{c}{\cellcolor[HTML]{E8F2FB}\textbf{MedXpertQA-MM}} & \multicolumn{3}{c}{\cellcolor[HTML]{FEF0E8}\textbf{VQA-RAD}} & \multicolumn{3}{c}{\cellcolor[HTML]{E8F2FB}\textbf{SLAKE}} & \multicolumn{3}{c}{\cellcolor[HTML]{FEF0E8}\textbf{MMMU-Medical}} \\
    \textit{Model} & \makecell{\textit{Basic}\\\textit{Percep.}} & \makecell{\textit{Content}\\\textit{Recog.}} & \makecell{\textit{Und./Diag./}\\\textit{Sug.}} & \makecell{\textit{Basic}\\\textit{Percep.}} & \makecell{\textit{Content}\\\textit{Recog.}} & \makecell{\textit{Und./Diag./}\\\textit{Sug.}} & \makecell{\textit{Basic}\\\textit{Percep.}} & \makecell{\textit{Content}\\\textit{Recog.}} & \makecell{\textit{Und./Diag./}\\\textit{Sug.}} & \makecell{\textit{Basic}\\\textit{Percep.}} & \makecell{\textit{Content}\\\textit{Recog.}} & \makecell{\textit{Und./Diag./}\\\textit{Sug.}} \\
    \midrule[0.8pt]
    Qwen3-VL-8B & 17.32 & 24.27 & 25.48 & 55.77 & 72.94 & 47.62 & 70.15 & 82.23 & 42.15 & 69.52 & 68.75 & 61.58 \\
    MIRA-VL-8B & 24.05 & 27.24 & 30.93 & 63.85 & 74.71 & 76.19 & 80.30 & 82.42 & 51.24 & 76.76 & 66.38 & 65.68 \\
    \textcolor[HTML]{8A8A8A}{\textit{Improvement $\uparrow$}} & \textcolor[HTML]{8A8A8A}{+6.73} & \textcolor[HTML]{8A8A8A}{+2.97} & \textcolor[HTML]{8A8A8A}{+5.45} & \textcolor[HTML]{8A8A8A}{+8.08} & \textcolor[HTML]{8A8A8A}{+1.76} & \textcolor[HTML]{8A8A8A}{+28.57} & \textcolor[HTML]{8A8A8A}{+10.15} & \textcolor[HTML]{8A8A8A}{+0.20} & \textcolor[HTML]{8A8A8A}{+9.09} & \textcolor[HTML]{8A8A8A}{+7.24} & \textcolor[HTML]{8A8A8A}{-2.37} & \textcolor[HTML]{8A8A8A}{+4.10} \\
    \bottomrule[1.2pt]
  \end{tabular}
  }
\end{table*}

\begin{wrapfigure}{r}{0.40\linewidth}
  \centering
  \vspace{-0.8em}
  \includegraphics[width=\linewidth]{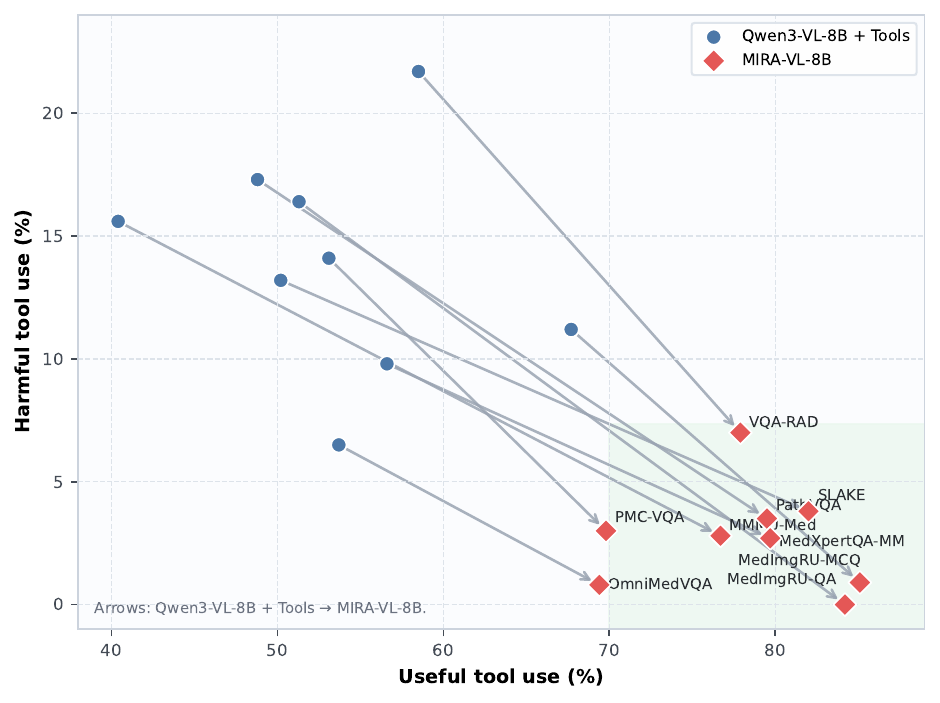}
  \caption{Per-dataset tool-use necessity shift from Qwen3-VL-8B + Tools to MIRA-VL-8B. Each arrow points from direct tool access to MIRA-VL-8B. The green shaded area indicates the desirable region with high useful tool use and low harmful tool use.}
  \label{fig:tool_necessity_scatter}
  \vspace{-1.0em}
\end{wrapfigure}

\paragraph{Fine-Grained Task Analysis} To better understand where MIRA-VL-8B improves over the backbone, we further break down the evaluation results by fine-grained task categories in Table~\ref{tab:fine-grained-improvements}. The Basic Perception category measures the model's ability to identify low-level visual evidence such as anatomical regions, visual attributes, and spatial cues; Content Recognition focuses on recognizing explicit visual or textual content; and Understanding/Diagnosis/Suggestion evaluates higher-level medical interpretation, diagnostic reasoning, and decision-oriented response generation. Across the four datasets, the gains mainly come from Basic Perception and Understanding/Diagnosis/Suggestion, while Content Recognition changes only marginally and even slightly decreases on one benchmark. This pattern suggests that MIRA does not primarily improve by memorizing more visual concepts or recognizing superficial content. Instead, its benefit comes from actively acquiring and verifying clinically relevant evidence, which improves visual grounding, and from using this evidence to support more reliable diagnostic reasoning and medical decision-making.

\subsection{Ablation and Analysis}
\label{sec:ablation}

\paragraph{The Impact of Training Strategies on the SFT Process}
This ablation studies how different SFT training strategies affect tool use, reflection ability, and downstream medical visual reasoning. The results are shown in Table~\ref{tab:sft_ablation}. The compared settings are summarized below.

\noindent$\diamond$ \textbf{Naive SFT:}
The backbone is fine-tuned only on the original medical VQA data, without tool-use trajectories or reflection supervision. This setting brings only marginal average improvement over the backbone, indicating that static VQA supervision alone is insufficient to induce reliable agentic reasoning.

\noindent$\diamond$ \textbf{+ MCTS Trajectories:}
The original VQA data is replaced with successful MCTS-generated tool-use trajectories, which teach the model how to inspect images, call tools, and verify visual evidence step by step. The improvement over Naive SFT shows that structured tool-use demonstrations provide a stronger cold-start signal than static VQA supervision.

\noindent$\diamond$ \textbf{+ Failure-driven Correction Reflection:}
Failure-driven correction reflection data is further added on top of the MCTS trajectory data. These examples teach the model to identify unreliable evidence and correct its reasoning path, complementing successful demonstrations with explicit failure-aware supervision.

\noindent$\diamond$ \textbf{+ Textual Intention Reflection:}
Textual intention reflection is further added to the above data, making each tool call explicitly tied to a diagnostic goal. This achieves the best average performance, suggesting that explicit intention supervision helps the model connect tool actions with medical reasoning objectives.

\begin{table*}[htbp]
  \centering
  \caption{Ablation study on SFT training strategies. All results are reported as accuracies in percentages.}
  \label{tab:sft_ablation}
  \footnotesize
  \setlength{\tabcolsep}{3.0pt}
  \renewcommand{\arraystretch}{1.08}
  \resizebox{\textwidth}{!}{
  \begin{tabular}{l c c c c c c c c c c c}
    \toprule[1.2pt]
    \textbf{Method} & \textbf{Tool} & \makecell{\textbf{SLAKE}} & \makecell{\textbf{PMC}\\\textbf{VQA}} & \makecell{\textbf{Omni}\\\textbf{MedVQA}} & \makecell{\textbf{Medlesion}\\\textbf{VQA}} & \makecell{\textbf{Medlesion}\\\textbf{MCQ}} & \makecell{\textbf{VQA}\\\textbf{RAD}} & \makecell{\textbf{PATH}\\\textbf{VQA}} & \makecell{\textbf{MedXpert}\\\textbf{QA-MM}} & \makecell{\textbf{MMMU}\\\textbf{Medical}} & \textbf{AVG.} \\
    \midrule[0.8pt]
    Qwen3-VL-8B & \ding{55} & 69.87 & 54.55 & 78.45 & 60.51 & 58.68 & 61.86 & 43.19 & 24.90 & 63.60 & 57.29 \\
    Qwen3-VL-8B + Tools & \ding{51} & 63.94\textsubscript{\textcolor[HTML]{22863A}{-5.9}} & 43.05\textsubscript{\textcolor[HTML]{22863A}{-11.5}} & 68.39\textsubscript{\textcolor[HTML]{22863A}{-10.1}} & 63.89\textsubscript{\textcolor[HTML]{D9822B}{+3.4}} & 65.40\textsubscript{\textcolor[HTML]{D9822B}{+6.7}} & 54.99\textsubscript{\textcolor[HTML]{22863A}{-6.9}} & 40.20\textsubscript{\textcolor[HTML]{22863A}{-3.0}} & 24.30\textsubscript{\textcolor[HTML]{22863A}{-0.6}} & 53.20\textsubscript{\textcolor[HTML]{22863A}{-10.4}} & 53.04\textsubscript{\textcolor[HTML]{22863A}{-4.2}} \\
    Naive SFT & \ding{55} & 77.20\textsubscript{\textcolor[HTML]{D9822B}{+7.3}} & 54.95\textsubscript{\textcolor[HTML]{D9822B}{+0.4}} & 77.05\textsubscript{\textcolor[HTML]{22863A}{-1.4}} & 70.24\textsubscript{\textcolor[HTML]{D9822B}{+9.7}} & 64.80\textsubscript{\textcolor[HTML]{D9822B}{+6.1}} & 58.09\textsubscript{\textcolor[HTML]{22863A}{-3.8}} & 43.21\textsubscript{\textcolor[HTML]{D9822B}{+0.0}} & 21.25\textsubscript{\textcolor[HTML]{22863A}{-3.7}} & 50.67\textsubscript{\textcolor[HTML]{22863A}{-12.9}} & 57.50\textsubscript{\textcolor[HTML]{D9822B}{+0.2}} \\
    \midrule[0.8pt]
    \rowcolor[HTML]{E8F2FB}\multicolumn{12}{c}{\textbf{Ours}} \\
    + MCTS Trajectories & \ding{51} & 77.51\textsubscript{\textcolor[HTML]{D9822B}{+7.6}} & 54.00\textsubscript{\textcolor[HTML]{22863A}{-0.5}} & 75.89\textsubscript{\textcolor[HTML]{22863A}{-2.6}} & 66.29\textsubscript{\textcolor[HTML]{D9822B}{+5.8}} & 68.82\textsubscript{\textcolor[HTML]{D9822B}{+10.1}} & 63.19\textsubscript{\textcolor[HTML]{D9822B}{+1.3}} & 48.46\textsubscript{\textcolor[HTML]{D9822B}{+5.3}} & 26.85\textsubscript{\textcolor[HTML]{D9822B}{+2.0}} & 59.07\textsubscript{\textcolor[HTML]{22863A}{-4.5}} & 60.01\textsubscript{\textcolor[HTML]{D9822B}{+2.7}} \\
    + Failure-driven Correction Reflection & \ding{51} & 76.70\textsubscript{\textcolor[HTML]{D9822B}{+6.8}} & 53.85\textsubscript{\textcolor[HTML]{22863A}{-0.7}} & 76.65\textsubscript{\textcolor[HTML]{22863A}{-1.8}} & 76.73\textsubscript{\textcolor[HTML]{D9822B}{+16.2}} & 62.51\textsubscript{\textcolor[HTML]{D9822B}{+3.8}} & 60.53\textsubscript{\textcolor[HTML]{22863A}{-1.3}} & 49.23\textsubscript{\textcolor[HTML]{D9822B}{+6.0}} & 27.20\textsubscript{\textcolor[HTML]{D9822B}{+2.3}} & 61.37\textsubscript{\textcolor[HTML]{22863A}{-2.2}} & 60.53\textsubscript{\textcolor[HTML]{D9822B}{+3.2}} \\
    \rowcolor[HTML]{FEF0E8}+ Textual Intention Reflection & \ding{51} & 77.27\textsubscript{\textcolor[HTML]{D9822B}{+7.4}} & 55.95\textsubscript{\textcolor[HTML]{D9822B}{+1.4}} & 74.92\textsubscript{\textcolor[HTML]{22863A}{-3.5}} & 74.19\textsubscript{\textcolor[HTML]{D9822B}{+13.7}} & 64.30\textsubscript{\textcolor[HTML]{D9822B}{+5.6}} & 66.52\textsubscript{\textcolor[HTML]{D9822B}{+4.7}} & 48.55\textsubscript{\textcolor[HTML]{D9822B}{+5.4}} & 29.90\textsubscript{\textcolor[HTML]{D9822B}{+5.0}} & 67.47\textsubscript{\textcolor[HTML]{D9822B}{+3.9}} & 62.12\textsubscript{\textcolor[HTML]{D9822B}{+4.8}} \\
    \bottomrule[1.2pt]
  \end{tabular}
  }
\end{table*}

\paragraph{Reward Design and Reflection Memory in RL}
Besides outcome and format rewards, two components are central to stable RL optimization. Table~\ref{tab:rl_ablation} compares the backbone, MIRA-SFT, MIRA-RL evaluated without tool execution, vanilla GRPO, and the final model with on-policy reflection memory.

\noindent$\diamond$ \textbf{MIRA-RL w/o Tools:}
This setting evaluates the RL-trained MIRA model with tool execution disabled, isolating how much performance depends on interactive evidence acquisition at inference time.

\noindent$\diamond$ \textbf{Format + Accuracy Reward:}
This setting uses tool-augmented RL rollouts with only the output-format and final-answer accuracy rewards, without the consistency reward or on-policy reflection memory.

\noindent$\diamond$ \textbf{Consistency reward:}
This reward checks whether the final answer is supported by the preceding reasoning and visual evidence, discouraging unsupported diagnostic jumps.

\noindent$\diamond$ \textbf{On-policy reflection memory:}
Low-reward rollouts are distilled into reusable reflections that summarize recurring failures, helping the model avoid repeated mistakes in later rollouts. Beyond the final benchmark ablation in Table~\ref{tab:rl_ablation}, Appendix~\ref{fig:val_accuracy_reward_memory_comparison} further compares validation accuracy rewards during RL training with and without reflection memory.

\begin{table*}[htbp]
  \centering
  \caption{Ablation study on RL training strategies. All results are reported as accuracies in percentages.}
  \label{tab:rl_ablation}
  \footnotesize
  \setlength{\tabcolsep}{3.0pt}
  \renewcommand{\arraystretch}{1.08}
  \resizebox{\textwidth}{!}{
  \begin{tabular}{l c c c c c c c c c c c}
    \toprule[1.2pt]
    \textbf{Method} & \textbf{Tool} & \makecell{\textbf{SLAKE}} & \makecell{\textbf{PMC}\\\textbf{VQA}} & \makecell{\textbf{Omni}\\\textbf{MedVQA}} & \makecell{\textbf{Medlesion}\\\textbf{VQA}} & \makecell{\textbf{Medlesion}\\\textbf{MCQ}} & \makecell{\textbf{VQA}\\\textbf{RAD}} & \makecell{\textbf{PATH}\\\textbf{VQA}} & \makecell{\textbf{MedXpert}\\\textbf{QA-MM}} & \makecell{\textbf{MMMU}\\\textbf{Medical}} & \textbf{AVG.} \\
    \midrule[0.8pt]
    Qwen3-VL-8B & \ding{55} & 69.87 & 54.55 & 78.45 & 60.51 & 58.68 & 61.86 & 43.19 & 24.90 & 63.60 & 57.29 \\
    MIRA-SFT & \ding{51} & 77.27\textsubscript{\textcolor[HTML]{D9822B}{+7.4}} & 55.95\textsubscript{\textcolor[HTML]{D9822B}{+1.4}} & 74.92\textsubscript{\textcolor[HTML]{22863A}{-3.5}} & 74.19\textsubscript{\textcolor[HTML]{D9822B}{+13.7}} & 64.30\textsubscript{\textcolor[HTML]{D9822B}{+5.6}} & 66.52\textsubscript{\textcolor[HTML]{D9822B}{+4.7}} & 48.55\textsubscript{\textcolor[HTML]{D9822B}{+5.4}} & 29.90\textsubscript{\textcolor[HTML]{D9822B}{+5.0}} & 67.47\textsubscript{\textcolor[HTML]{D9822B}{+3.9}} & 62.12\textsubscript{\textcolor[HTML]{D9822B}{+4.8}} \\
    MIRA-RL w/o Tools & \ding{55} & 74.98\textsubscript{\textcolor[HTML]{D9822B}{+5.1}} & 53.00\textsubscript{\textcolor[HTML]{22863A}{-1.6}} & 79.34\textsubscript{\textcolor[HTML]{D9822B}{+0.9}} & 72.75\textsubscript{\textcolor[HTML]{D9822B}{+12.2}} & 71.74\textsubscript{\textcolor[HTML]{D9822B}{+13.1}} & 62.75\textsubscript{\textcolor[HTML]{D9822B}{+0.9}} & 46.58\textsubscript{\textcolor[HTML]{D9822B}{+3.4}} & 25.00\textsubscript{\textcolor[HTML]{D9822B}{+0.1}} & 66.00\textsubscript{\textcolor[HTML]{D9822B}{+2.4}} & 61.35\textsubscript{\textcolor[HTML]{D9822B}{+4.1}} \\
    \midrule[0.8pt]
    \rowcolor[HTML]{E8F2FB}\multicolumn{12}{c}{\textbf{Ours}} \\
    Format + Accuracy Reward & \ding{51} & 75.98\textsubscript{\textcolor[HTML]{D9822B}{+6.1}} & 56.10\textsubscript{\textcolor[HTML]{D9822B}{+1.6}} & 77.64\textsubscript{\textcolor[HTML]{22863A}{-0.8}} & 72.07\textsubscript{\textcolor[HTML]{D9822B}{+11.6}} & 71.72\textsubscript{\textcolor[HTML]{D9822B}{+13.0}} & 67.18\textsubscript{\textcolor[HTML]{D9822B}{+5.3}} & 49.13\textsubscript{\textcolor[HTML]{D9822B}{+5.9}} & 28.65\textsubscript{\textcolor[HTML]{D9822B}{+3.8}} & 64.80\textsubscript{\textcolor[HTML]{D9822B}{+1.2}} & 62.58\textsubscript{\textcolor[HTML]{D9822B}{+5.3}} \\
    + Consistency Reward & \ding{51} & 76.89\textsubscript{\textcolor[HTML]{D9822B}{+7.0}} & 53.75\textsubscript{\textcolor[HTML]{22863A}{-0.8}} & 82.85\textsubscript{\textcolor[HTML]{D9822B}{+4.4}} & 74.05\textsubscript{\textcolor[HTML]{D9822B}{+13.5}} & 73.31\textsubscript{\textcolor[HTML]{D9822B}{+14.6}} & 64.75\textsubscript{\textcolor[HTML]{D9822B}{+2.9}} & 47.97\textsubscript{\textcolor[HTML]{D9822B}{+4.8}} & 26.00\textsubscript{\textcolor[HTML]{D9822B}{+1.1}} & 64.67\textsubscript{\textcolor[HTML]{D9822B}{+1.1}} & 62.69\textsubscript{\textcolor[HTML]{D9822B}{+5.4}} \\
    \rowcolor[HTML]{FEF0E8}+ On-policy Reflection Memory & \ding{51} & 77.46\textsubscript{\textcolor[HTML]{D9822B}{+7.6}} & 58.35\textsubscript{\textcolor[HTML]{D9822B}{+3.8}} & 81.65\textsubscript{\textcolor[HTML]{D9822B}{+3.2}} & 75.04\textsubscript{\textcolor[HTML]{D9822B}{+14.5}} & 73.12\textsubscript{\textcolor[HTML]{D9822B}{+14.4}} & 68.51\textsubscript{\textcolor[HTML]{D9822B}{+6.7}} & 48.83\textsubscript{\textcolor[HTML]{D9822B}{+5.6}} & 30.60\textsubscript{\textcolor[HTML]{D9822B}{+5.7}} & 68.93\textsubscript{\textcolor[HTML]{D9822B}{+5.3}} & 64.73\textsubscript{\textcolor[HTML]{D9822B}{+7.4}} \\
    \bottomrule[1.2pt]
  \end{tabular}
  }
\end{table*}

\paragraph{Tool-use Necessity Analysis} To understand whether tool calls are meaningful rather than merely frequent, we ask a GPT-4o judge to classify each tool-used sample as \textit{needed}, \textit{optional}, \textit{unnecessary}, or \textit{harmful}. Figure~\ref{fig:tool_necessity_scatter} compares direct tool access for the Qwen3-VL-8B backbone with MIRA-VL-8B. Here, useful tool use is defined as the sum of \textit{needed} and \textit{optional} tool use. Directly enabling tools for the backbone yields only 56.2\% useful tool use and 8.9\% harmful tool use. In contrast, MIRA-VL-8B increases useful tool use to 73.8\% and reduces harmful tool use to 1.6\%. The improvement is consistent across all nine benchmarks, indicating that MIRA-VL-8B learns not only to call tools, but also to avoid unnecessary or misleading tool interactions.

\paragraph{Analysis of the RL Learning Process}
Figure~\ref{fig:rl_training_curves} illustrates the dynamics of several key metrics during RL, from which we make three observations. First, the completion length rapidly decreases from over 1,000 tokens to around 500 tokens in the early stage and further converges to about 400 tokens, while the average number of tool-use turns drops from roughly 2.2 to 1.3. This suggests that RL suppresses unnecessary verbose reasoning and redundant tool calls, encouraging the model to use tools only when additional evidence is needed. Second, the total reward steadily increases throughout training, driven by improvements in both accuracy reward and consistency reward. The accuracy reward rises from about 0.4 to around 0.65--0.7, indicating that on-policy optimization effectively improves final-answer correctness. Third, the consistency reward also increases from roughly 0.35 to above 0.5, showing that the model becomes better at aligning its final answers with the preceding evidence and reasoning. Together with the format reward quickly approaching 1.0, these trends indicate that RL not only improves answer accuracy, but also makes the model's tool-use trajectories shorter, better formatted, and more evidence-consistent.

\begin{figure*}[htbp]
  \centering
  \includegraphics[width=\textwidth]{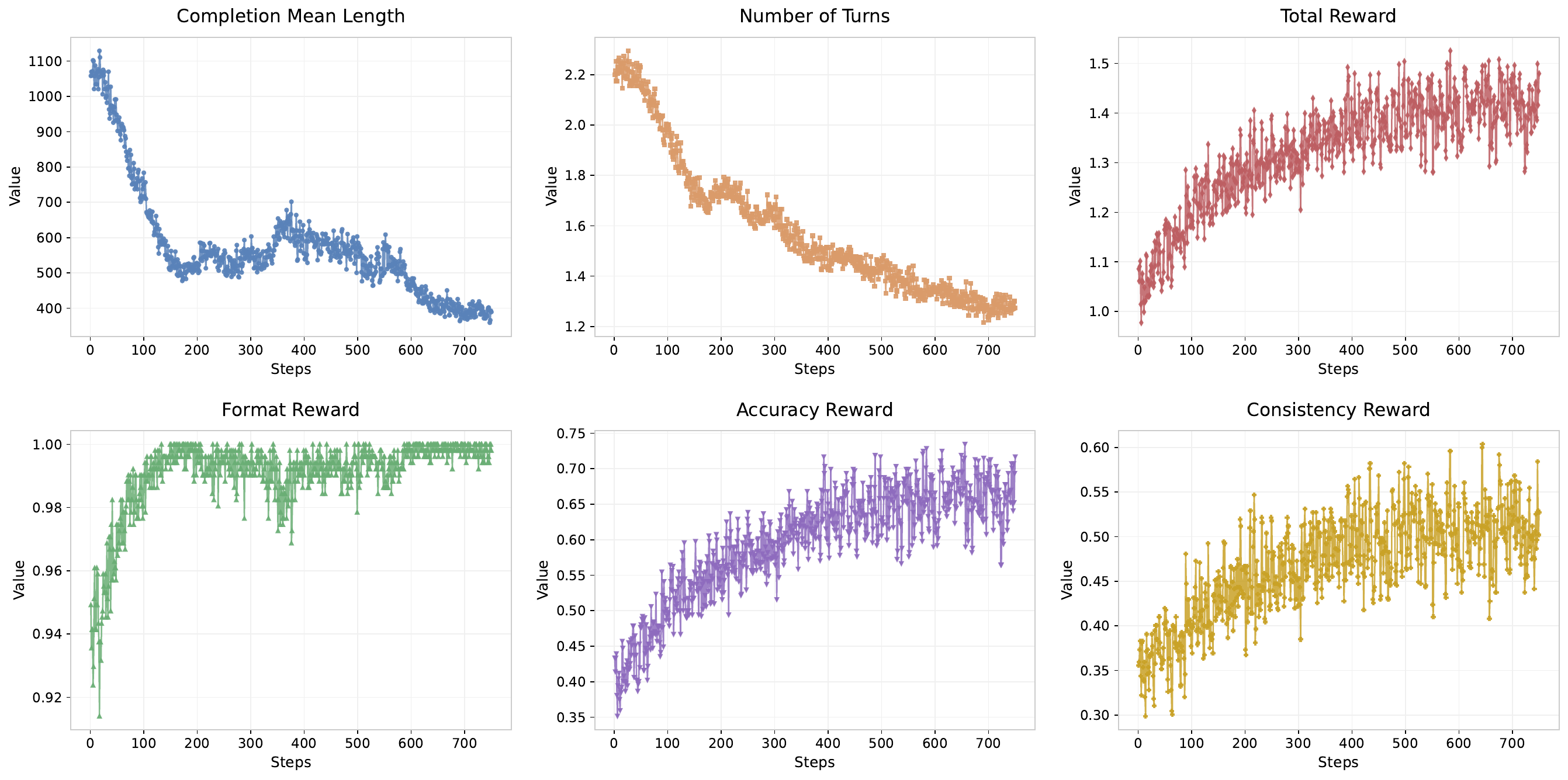}
  \caption{Training curves of the RL stage. The curves illustrate the learning dynamics of the reward signals during on-policy optimization.}
  \label{fig:rl_training_curves}
\end{figure*}

\section{Conclusion and Limitations}
\label{sec:conclusion}

We introduce MIRA, a medical visual reasoning agent that enhances LVLMs with active evidence acquisition, tool-grounded verification, and reflection-driven self-correction. Our approach follows a two-stage training strategy. In the SFT stage, we use MCTS-generated tool-use trajectories together with failure-driven correction reflection and textual intention reflection to provide a strong cold-start policy. In the RL stage, we further optimize the model with on-policy tool-augmented rollouts, consistency-aware reward design, and reflection memory updates. Extensive evaluations across diverse medical VQA benchmarks show that MIRA-VL-8B consistently improves over the Qwen3-VL-8B backbone, with especially strong gains on fine-grained lesion understanding and decision-oriented medical reasoning tasks. Fine-grained analysis further indicates that the improvements mainly come from stronger basic perception and understanding/diagnosis/suggestion abilities, rather than simple content recognition.

Despite these promising results, several limitations remain and point to future directions:

\noindent$\diamond$ \textbf{Base model capability.}
MIRA is still constrained by the perception, language understanding, and medical knowledge of its backbone model. When the base LVLM fails to recognize subtle visual findings or lacks necessary clinical knowledge, tool use and reflection can reduce but not fully eliminate the error. Stronger medical foundation models may further improve reliability.

\noindent$\diamond$ \textbf{Reflection memory generalization.}
The reflection memory is updated from on-policy failures and can help the model avoid repeated mistakes. However, the learned memories may not fully generalize to rare diseases, unseen imaging styles, or substantially different clinical contexts. Future work can explore more systematic memory validation, retrieval, and editing mechanisms to improve cross-domain robustness.

\clearpage

\ifseparateappendixbibliography
\putbib[main]
\end{bibunit}
\else
\bibliographystyle{plainnat}
\bibliography{main}

@article{bai2025qwen3,
  title={Qwen3-VL Technical Report},
  author={Bai, Shuai and Cai, Yuxuan and Chen, Ruizhe and Chen, Keqin and Chen, Xionghui and Cheng, Zesen and Deng, Lianghao and Ding, Wei and Gao, Chang and Ge, Chunjiang and others},
  journal={arXiv preprint arXiv:2511.21631},
  year={2025}
}

@inproceedings{yumedlesionvqa,
  title={MedLesionVQA: A Multimodal Benchmark Emulating Clinical Visual Diagnosis for Body Surface Health},
  author={Yu, Deli and Wang, Shengzhi and Ji, Xiaozhong and Cui, Bo and Cao, Jieqiong and Wang, Huichao and Jiang, Boyuan and Wang, Xu and Xu, Qian and Zhao, Yi and others},
  booktitle={The Fourteenth International Conference on Learning Representations},
  year={2026}
}

@article{zheng2025deepeyes,
  title={DeepEyes: Incentivizing" Thinking with Images" via Reinforcement Learning},
  author={Zheng, Ziwei and Yang, Michael and Hong, Jack and Zhao, Chenxiao and Xu, Guohai and Yang, Le and Shen, Chao and Yu, Xing},
  journal={arXiv preprint arXiv:2505.14362},
  year={2025}
}

@article{hong2025deepeyesv2,
  title={DeepEyesV2: Toward Agentic Multimodal Model},
  author={Hong, Jack and Zhao, Chenxiao and Zhu, ChengLin and Lu, Weiheng and Xu, Guohai and Yu, Xing},
  journal={arXiv preprint arXiv:2511.05271},
  year={2025}
}

@article{medpalm2023,
  title={Large Language Models Encode Clinical Knowledge},
  author={Singhal, Karan and Azizi, Shekoofeh and Tu, Tao and others},
  journal={Nature},
  year={2023}
}

@inproceedings{moor2023med,
  title={Med-flamingo: a multimodal medical few-shot learner},
  author={Moor, Michael and Huang, Qian and Wu, Shirley and Yasunaga, Michihiro and Dalmia, Yash and Leskovec, Jure and Zakka, Cyril and Reis, Eduardo Pontes and Rajpurkar, Pranav},
  booktitle={Machine Learning for Health (ML4H)},
  pages={353--367},
  year={2023},
  organization={PMLR}
}

@article{llavamed2023,
  title={LLaVA-Med: Training a Large Language-and-Vision Assistant for Biomedicine},
  author={Li, Chunyuan and others},
  journal={arXiv preprint arXiv:2306.00890},
  year={2023}
}

@article{vqarad2018,
  title={A dataset of clinically generated visual questions and answers about radiology images},
  author={Lau, Jason J and Gayen, Soumya and Ben Abacha, Asma and Demner-Fushman, Dina},
  journal={Scientific data},
  volume={5},
  number={1},
  pages={1--10},
  year={2018},
  publisher={Nature Publishing Group}
}

@article{pathvqa2020,
  title={Pathvqa: 30000+ questions for medical visual question answering},
  author={He, Xuehai and Zhang, Yichen and Mou, Luntian and Xing, Eric and Xie, Pengtao},
  journal={arXiv preprint arXiv:2003.10286},
  year={2020}
}

@inproceedings{pmcvqa2021,
  title={PMC-VQA: Visual Question Answering for Medical Images},
  author={Liu, Bo and others},
  booktitle={Proceedings of the ACM International Conference on Multimedia},
  year={2021}
}

@misc{medgemini2024,
      title={Capabilities of {G}emini Models in Medicine}, 
      author={Khaled Saab and Tao Tu and Wei-Hung Weng and Ryutaro Tanno and David Stutz and Ellery Wulczyn and Fan Zhang and Tim Strother and Chunjong Park and Elahe Vedadi and Juanma Zambrano Chaves and Szu-Yeu Hu and Mike Schaekermann and Aishwarya Kamath and Yong Cheng and David G. T. Barrett and Cathy Cheung and Basil Mustafa and Anil Palepu and Daniel McDuff and Le Hou and Tomer Golany and Luyang Liu and Jean-baptiste Alayrac and Neil Houlsby and Nenad Tomasev and Jan Freyberg and Charles Lau and Jonas Kemp and Jeremy Lai and Shekoofeh Azizi and Kimberly Kanada and SiWai Man and Kavita Kulkarni and Ruoxi Sun and Siamak Shakeri and Luheng He and Ben Caine and Albert Webson and Natasha Latysheva and Melvin Johnson and Philip Mansfield and Jian Lu and Ehud Rivlin and Jesper Anderson and Bradley Green and Renee Wong and Jonathan Krause and Jonathon Shlens and Ewa Dominowska and S. M. Ali Eslami and Katherine Chou and Claire Cui and Oriol Vinyals and Koray Kavukcuoglu and James Manyika and Jeff Dean and Demis Hassabis and Yossi Matias and Dale Webster and Joelle Barral and Greg Corrado and Christopher Semturs and S. Sara Mahdavi and Juraj Gottweis and Alan Karthikesalingam and Vivek Natarajan},
      year={2024},
      eprint={2404.18416},
      archivePrefix={arXiv},
      primaryClass={cs.AI},
      url={https://arxiv.org/abs/2404.18416}, 
}

@inproceedings{mmmu2024,
  title     = {{MMMU}: A Massive Multi-discipline Multimodal Understanding and Reasoning Benchmark},
  author    = {Yue, Xiang and others},
  booktitle = {Proceedings of the IEEE/CVF Conference on Computer Vision and Pattern Recognition (CVPR)},
  year      = {2024}
}

@inproceedings{slake2021,
  title={Slake: A semantically-labeled knowledge-enhanced dataset for medical visual question answering},
  author={Liu, Bo and Zhan, Li-Ming and Xu, Li and Ma, Lin and Yang, Yan and Wu, Xiao-Ming},
  booktitle={2021 IEEE 18th international symposium on biomedical imaging (ISBI)},
  pages={1650--1654},
  year={2021},
  organization={IEEE}
}

@inproceedings{omnimedvqa2024,
  title={Omnimedvqa: A new large-scale comprehensive evaluation benchmark for medical lvlm},
  author={Hu, Yutao and Li, Tianbin and Lu, Quanfeng and Shao, Wenqi and He, Junjun and Qiao, Yu and Luo, Ping},
  booktitle={Proceedings of the IEEE/CVF Conference on Computer Vision and Pattern Recognition},
  pages={22170--22183},
  year={2024}
}

@article{gu2026medvh,
  title={MedVH: Toward systematic evaluation of hallucination for large vision language models in the medical context},
  author={Gu, Zishan and Chen, Jiayuan and Liu, Fenglin and Yin, Changchang and Zhang, Ping},
  journal={Advanced Intelligent Systems},
  volume={8},
  number={1},
  pages={2500255},
  year={2026},
  publisher={Wiley Online Library}
}

@inproceedings{wu2024v,
  title={V?: Guided visual search as a core mechanism in multimodal llms},
  author={Wu, Penghao and Xie, Saining},
  booktitle={Proceedings of the IEEE/CVF Conference on Computer Vision and Pattern Recognition},
  pages={13084--13094},
  year={2024}
}

@article{shi2026medxiaohe,
  title={Medxiaohe: A comprehensive recipe for building medical mllms},
  author={Shi, Baorong and Cui, Bo and Jiang, Boyuan and Yu, Deli and Qian, Fang and Yang, Haihua and Wang, Huichao and Chen, Jiale and Pan, Jianfei and Cao, Jieqiong and others},
  journal={arXiv preprint arXiv:2602.12705},
  year={2026}
}

@article{lu2026medvistagym,
  title={MEDVISTAGYM: A Scalable Training Environment for Thinking with Medical Images via Tool-Integrated Reinforcement Learning},
  author={Lu, Meng and Lu, Yuxing and Zhuang, Yuchen and Mullins, Megan and Xie, Yang and Xiao, Guanghua and Fleming, Charles and Shi, Wenqi and Wang, Xuan},
  journal={arXiv preprint arXiv:2601.07107},
  year={2026}
}

@article{chen2026meissa,
  title={Meissa: Multi-modal medical agentic intelligence},
  author={Chen, Yixiong and Bai, Xinyi and Pan, Yue and Zhou, Zongwei and Yuille, Alan},
  journal={arXiv preprint arXiv:2603.09018},
  year={2026}
}

@inproceedings{li2026medscope,
  title={MedScope: Incentivizing" Think with Videos" for Clinical Reasoning via Coarse-to-Fine Tool Calling},
  author={Li, Wenjie and Zhang, Yujie and Sun, Haoran and He, Xingqi and Gao, Hongcheng and Ma, Chenglong and Hu, Ming and Wang, Guankun and Yao, Shiyi and Yang, Renhao and others},
  booktitle={Forty-third International Conference on Machine Learning},
  year={2026}
}

@article{liu2026medsam,
  title={MedSAM-Agent: Empowering Interactive Medical Image Segmentation with Multi-turn Agentic Reinforcement Learning},
  author={Liu, Shengyuan and Bao, Liuxin and Yang, Qi and Geng, Wanting and Zheng, Boyun and Li, Chenxin and Chen, Wenting and Peng, Houwen and Yuan, Yixuan},
  journal={arXiv preprint arXiv:2602.03320},
  year={2026}
}

@article{zhang2025thyme,
  title={Thyme: Think beyond images},
  author={Zhang, Yi-Fan and Lu, Xingyu and Yin, Shukang and Fu, Chaoyou and Chen, Wei and Hu, Xiao and Wen, Bin and Jiang, Kaiyu and Liu, Changyi and Zhang, Tianke and others},
  journal={arXiv preprint arXiv:2508.11630},
  year={2025}
}

@article{alayrac2022flamingo,
  title={Flamingo: a visual language model for few-shot learning},
  author={Alayrac, Jean-Baptiste and Donahue, Jeff and Luc, Pauline and Miech, Antoine and Barr, Iain and Hasson, Yana and Lenc, Karel and Mensch, Arthur and Millican, Katherine and Reynolds, Malcolm and others},
  journal={Advances in neural information processing systems},
  volume={35},
  pages={23716--23736},
  year={2022}
}

@inproceedings{li2023blip,
  title={Blip-2: Bootstrapping language-image pre-training with frozen image encoders and large language models},
  author={Li, Junnan and Li, Dongxu and Savarese, Silvio and Hoi, Steven},
  booktitle={International conference on machine learning},
  pages={19730--19742},
  year={2023},
  organization={PMLR}
}

@article{dai2023instructblip,
  title={Instructblip: Towards general-purpose vision-language models with instruction tuning},
  author={Dai, Wenliang and Li, Junnan and Li, Dongxu and Tiong, Anthony and Zhao, Junqi and Wang, Weisheng and Li, Boyang and Fung, Pascale N and Hoi, Steven},
  journal={Advances in neural information processing systems},
  volume={36},
  pages={49250--49267},
  year={2023}
}

@article{liu2023visual,
  title={Visual instruction tuning},
  author={Liu, Haotian and Li, Chunyuan and Wu, Qingyang and Lee, Yong Jae},
  journal={Advances in neural information processing systems},
  volume={36},
  pages={34892--34916},
  year={2023}
}

@article{su2025openthinkimg,
  title={Openthinkimg: Learning to think with images via visual tool reinforcement learning},
  author={Su, Zhaochen and Li, Linjie and Song, Mingyang and Hao, Yunzhuo and Yang, Zhengyuan and Zhang, Jun and Chen, Guanjie and Gu, Jiawei and Li, Juntao and Qu, Xiaoye and others},
  journal={arXiv preprint arXiv:2505.08617},
  year={2025}
}

@article{wang2025pixel,
  title={Pixel reasoner: Incentivizing pixel-space reasoning with curiosity-driven reinforcement learning},
  author={Wang, Haozhe and Su, Alex and Ren, Weiming and Lin, Fangzhen and Chen, Wenhu},
  journal={arXiv preprint arXiv:2505.15966},
  year={2025}
}

@article{wu2025vtool,
  title={Vtool-r1: Vlms learn to think with images via reinforcement learning on multimodal tool use},
  author={Wu, Mingyuan and Yang, Jingcheng and Jiang, Jize and Li, Meitang and Yan, Kaizhuo and Yu, Hanchao and Zhang, Minjia and Zhai, Chengxiang and Nahrstedt, Klara},
  journal={arXiv preprint arXiv:2505.19255},
  year={2025}
}

@article{lai2025mini,
  title={Mini-o3: Scaling up reasoning patterns and interaction turns for visual search},
  author={Lai, Xin and Li, Junyi and Li, Wei and Liu, Tao and Li, Tianjian and Zhao, Hengshuang},
  journal={arXiv preprint arXiv:2509.07969},
  year={2025}
}

@article{chern2025thinking,
  title={Thinking with generated images},
  author={Chern, Ethan and Hu, Zhulin and Chern, Steffi and Kou, Siqi and Su, Jiadi and Ma, Yan and Deng, Zhijie and Liu, Pengfei},
  journal={arXiv preprint arXiv:2505.22525},
  year={2025}
}

@article{li2025latent,
  title={Latent visual reasoning},
  author={Li, Bangzheng and Sun, Ximeng and Liu, Jiang and Wang, Ze and Wu, Jialian and Yu, Xiaodong and Chen, Hao and Barsoum, Emad and Chen, Muhao and Liu, Zicheng},
  journal={arXiv preprint arXiv:2509.24251},
  year={2025}
}

@article{lingshu2024,
  title={Lingshu: A Generalist Foundation Model for Unified Multimodal Medical Understanding and Reasoning},
  author={Xu, Weiwen and Chan, Hou Pong and Li, Long and Aljunied, Mahani and Yuan, Ruifeng and Wang, Jianyu and Xiao, Chenghao and Chen, Guizhen and Liu, Chaoqun and Li, Zhaodonghui and others},
  journal={arXiv preprint arXiv:2506.07044},
  year={2025}
}

@article{shao2024deepseekmath,
  title={Deepseekmath: Pushing the limits of mathematical reasoning in open language models},
  author={Shao, Zhihong and Wang, Peiyi and Zhu, Qihao and Xu, Runxin and Song, Junxiao and Bi, Xiao and Zhang, Haowei and Zhang, Mingchuan and Li, YK and Wu, Yang and others},
  journal={arXiv preprint arXiv:2402.03300},
  year={2024}
}

@article{mullappilly2026medix,
  title={Medix-r1: Open ended medical reinforcement learning},
  author={Mullappilly, Sahal Shaji and Kurpath, Mohammed Irfan and Mohamed, Omair and Zidan, Mohamed and Khan, Fahad and Khan, Salman and Anwer, Rao and Cholakkal, Hisham},
  journal={arXiv preprint arXiv:2602.23363},
  year={2026}
}

@article{jiang2025incentivizing,
  title={Incentivizing Tool-augmented Thinking with Images for Medical Image Analysis},
  author={Jiang, Yankai and Zhang, Yujie and Zhang, Peng and Li, Yichen and Chen, Jintai and Shi, Xiaoming and Zhen, Shihui},
  journal={arXiv preprint arXiv:2512.14157},
  year={2025}
}

@inproceedings{yan2026medreasoner,
  title={Medreasoner: Reinforcement learning drives reasoning grounding from clinical thought to pixel-level precision},
  author={Yan, Zhonghao and Diao, Muxi and Yang, Yuxuan and Jing, Ruoyan and Xu, Jiayuan and Zhang, Kaizhou and Yang, Lele and Liu, Yanxi and Liang, Kongming and Ma, Zhanyu},
  booktitle={Proceedings of the AAAI Conference on Artificial Intelligence},
  volume={40},
  number={14},
  pages={11577--11585},
  year={2026}
}

@article{deria2026medmo,
  title={MedMO: Grounding and Understanding Multimodal Large Language Model for Medical Images},
  author={Deria, Ankan and Kumar, Komal and Dukre, Adinath Madhavrao and Segal, Eran and Khan, Salman and Razzak, Imran},
  journal={arXiv preprint arXiv:2602.06965},
  year={2026}
}

@inproceedings{pan2025medvlm,
  title={Medvlm-r1: Incentivizing medical reasoning capability of vision-language models (vlms) via reinforcement learning},
  author={Pan, Jiazhen and Liu, Che and Wu, Junde and Liu, Fenglin and Zhu, Jiayuan and Li, Hongwei Bran and Chen, Chen and Ouyang, Cheng and Rueckert, Daniel},
  booktitle={International Conference on Medical Image Computing and Computer-Assisted Intervention},
  pages={337--347},
  year={2025},
  organization={Springer}
}

@misc{lai2025medr1reinforcementlearninggeneralizable,
      title={Med-R1: Reinforcement Learning for Generalizable Medical Reasoning in Vision-Language Models}, 
      author={Yuxiang Lai and Jike Zhong and Ming Li and Shitian Zhao and Yuheng Li and Konstantinos Psounis and Xiaofeng Yang},
      year={2025},
      eprint={2503.13939},
      archivePrefix={arXiv},
      primaryClass={cs.CV},
      url={https://arxiv.org/abs/2503.13939}, 
}

@article{lu2024multimodal,
  title={A multimodal generative AI copilot for human pathology},
  author={Lu, Ming Y and Chen, Bowen and Williamson, Drew FK and Chen, Richard J and Zhao, Melissa and Chow, Aaron K and Ikemura, Kenji and Kim, Ahrong and Pouli, Dimitra and Patel, Ankush and others},
  journal={Nature},
  volume={634},
  number={8033},
  pages={466--473},
  year={2024},
  publisher={Nature Publishing Group UK London}
}

@misc{chen2024huatuogptvisioninjectingmedicalvisual,
      title={HuatuoGPT-Vision, Towards Injecting Medical Visual Knowledge into Multimodal LLMs at Scale}, 
      author={Junying Chen and Chi Gui and Ruyi Ouyang and Anningzhe Gao and Shunian Chen and Guiming Hardy Chen and Xidong Wang and Ruifei Zhang and Zhenyang Cai and Ke Ji and Guangjun Yu and Xiang Wan and Benyou Wang},
      year={2024},
      eprint={2406.19280},
      archivePrefix={arXiv},
      primaryClass={cs.CV},
      url={https://arxiv.org/abs/2406.19280}, 
}

@article{Zuo2025MedXpertQA,
  title   = {MedXpertQA: Benchmarking Expert-Level Medical Reasoning and Understanding},
  author  = {Yuxin Zuo and Shang Qu and Yifei Li and Zhangren Chen and Xuekai Zhu and Ermo Hua and Kaiyan Zhang and Ning Ding and Bowen Zhou},
  journal = {arXiv preprint arXiv:2501.18362},
  year    = {2025},
  url     = {https://arxiv.org/abs/2501.18362}
}

@article{wang2026beyond,
  title={Beyond Textual Rationales: Anatomy-Grounded Chain-of-Thought for Traceable Radiology Reasoning},
  author={Wang, Shengzhi and Wu, Kai and Yang, Jun and Xu, Mengyuan and Xiong, Mingliang and Fang, Wen and Liu, Mingqing and Deng, Hao and He, Bin and Li, Gang and others},
  journal={Knowledge-Based Systems},
  pages={116475},
  year={2026},
  publisher={Elsevier}
}

@article{zhu2024medical,
  title={Medical SAM 2: Segment Medical Images as Video via Segment Anything Model 2},
  author={Zhu, Jiayuan and Hamdi, Abdullah and Qi, Yunli and Jin, Yueming and Wu, Junde},
  journal={arXiv preprint arXiv:2408.00874},
  year={2024}
}

@inproceedings{ravi2025sam,
  title={SAM 2: Segment Anything in Images and Videos},
  author={Ravi, Nikhila and Gabeur, Valentin and Hu, Yuan-Ting and Hu, Ronghang and Ryali, Chaitanya and Ma, Tengyu and Khedr, Haitham and R{\"a}dle, Roman and Rolland, Chloe and Gustafson, Laura and others},
  booktitle={International Conference on Learning Representations},
  volume={2025},
  pages={28085--28128},
  year={2025}
}

@inproceedings{liu2024grounding,
  title={Grounding DINO: Marrying DINO with Grounded Pre-Training for Open-Set Object Detection},
  author={Liu, Shilong and Zeng, Zhaoyang and Ren, Tianhe and Li, Feng and Zhang, Hao and Yang, Jie and Jiang, Qing and Li, Chunyuan and Yang, Jianwei and Su, Hang and others},
  booktitle={European Conference on Computer Vision},
  pages={38--55},
  year={2024},
  organization={Springer}
}

@article{seed2025seed1_5vl,
  title={Seed1. 5-vl technical report},
  author={Guo, Dong and Wu, Faming and Zhu, Feida and Leng, Fuxing and Shi, Guang and Chen, Haobin and Fan, Haoqi and Wang, Jian and Jiang, Jianyu and Wang, Jiawei and others},
  journal={arXiv preprint arXiv:2505.07062},
  year={2025}
}

@misc{google2025gemini25pro,
  author       = {{Google}},
  title        = {{Gemini 2.5 Pro}},
  year         = {2025},
  howpublished = {\url{https://deepmind.google/models/gemini/pro/}},
  note         = {Accessed: 2026-07-14}
}

@inproceedings{kocsis2006bandit,
  title={Bandit based monte-carlo planning},
  author={Kocsis, Levente and Szepesv{\'a}ri, Csaba},
  booktitle={European conference on machine learning},
  pages={282--293},
  year={2006},
  organization={Springer}
}

@article{chen2024expanding,
  title={Expanding performance boundaries of open-source multimodal models with model, data, and test-time scaling},
  author={Chen, Zhe and Wang, Weiyun and Cao, Yue and Liu, Yangzhou and Gao, Zhangwei and Cui, Erfei and Zhu, Jinguo and Ye, Shenglong and Tian, Hao and Liu, Zhaoyang and others},
  journal={arXiv preprint arXiv:2412.05271},
  year={2024}
}

@misc{openai2024gpt4o,
  author       = {{OpenAI}},
  title        = {Hello GPT-4o},
  year         = {2024},
  month        = {May},
  howpublished = {\url{https://openai.com/index/hello-gpt-4o/}},
  note         = {Accessed: 2026-07-27}
}

@misc{openai2025gpt52,
  author       = {{OpenAI}},
  title        = {Introducing GPT-5.2},
  year         = {2025},
  month        = {December},
  howpublished = {\url{https://openai.com/index/introducing-gpt-5-2/}},
  note         = {Accessed: 2026-07-27}
}
\fi

\clearpage

\beginappendix
\setcounter{page}{1}
\renewcommand{\thepage}{A\arabic{page}}
\setcounter{figure}{0}
\renewcommand{\thefigure}{A\arabic{figure}}
\renewcommand{\theHfigure}{appendix.A\arabic{figure}}
\setcounter{table}{0}
\renewcommand{\thetable}{A\arabic{table}}
\renewcommand{\theHtable}{appendix.A\arabic{table}}

\ifseparateappendixbibliography
\begin{bibunit}[plainnat]
\fi
\section*{Appendix Contents}
\label{sec:appendix_contents}

This appendix provides supplementary material for MIRA, including related work, implementation details of the diagnostic tool interface and MCTS data engine, additional training-data analysis, details of the online reflection memory update mechanism, qualitative attention visualizations, evaluation prompts, and case studies. These materials are intended to clarify how MIRA collects visual evidence, verifies diagnostic reasoning, and improves evidence-grounded behavior through SFT and RL.

\vspace{0.8em}
\newcommand{\appendixcontentsline}[3]{\noindent\textcolor[HTML]{1A5FB4}{\textbf{#1}}\hspace{0.7em}\textcolor[HTML]{1A5FB4}{\dotfill}\hspace{0.7em}\textcolor[HTML]{1A5FB4}{\hyperref[#2]{#3}}\par\vspace{0.35em}}
\appendixcontentsline{A}{sec:related_work}{Related Work}
\appendixcontentsline{B}{sec:tool_use_details}{Tool Usage Details}
\appendixcontentsline{C}{sec:mcts_details}{MCTS Details}
\appendixcontentsline{D}{appendix:sft_data_analysis}{SFT Training Data Analysis}
\appendixcontentsline{E}{sec:online_reflection_memory_update_details}{Online Reflection Memory Update Details}
\appendixcontentsline{F}{sec:appendix_attention_comparison}{Attention Analysis}
\appendixcontentsline{G}{sec:eval_prompts}{Evaluation Details}
\appendixcontentsline{H}{sec:case_studies}{Case Studies}

\clearpage

\section{Related Work}
\label{sec:related_work}

\paragraph{General Visual Reasoning Models}
Early LVLMs~\citep{alayrac2022flamingo,li2023blip,dai2023instructblip,liu2023visual} relied on single-pass visual encoding, conditioning all subsequent reasoning on a fixed visual representation. Recent reasoning-oriented models improve textual chain-of-thought but still cannot actively revisit ambiguous image regions. To address this limitation, an emerging line of research treats images as dynamic reasoning workspaces. V*~\citep{wu2024v} enables iterative visual localization, cropping, and zooming. OpenThinkIMG~\citep{su2025openthinkimg}, Pixel Reasoner~\citep{wang2025pixel}, DeepEyes~\citep{zheng2025deepeyes}, DeepEyesV2~\citep{hong2025deepeyesv2}, and VTool-R1~\citep{wu2025vtool} train models to interleave textual reasoning with visual-tool operations, while Mini-o3~\citep{lai2025mini} scales such interaction to long-horizon visual search. Thyme~\citep{zhang2025thyme} further supports executable image-processing and computational operations. Complementary approaches explore intrinsic visual reasoning: Thinking with Generated Images~\citep{chern2025thinking} uses self-generated images as intermediate thoughts, whereas Latent Visual Reasoning~\citep{li2025latent} performs reasoning through latent visual tokens or visual embedding states.

\paragraph{Medical Visual Reasoning Models}
Medical LVLMs~\citep{llavamed2023,moor2023med,medpalm2023,medgemini2024,chen2024huatuogptvisioninjectingmedicalvisual,lu2024multimodal,lingshu2024} have advanced multimodal diagnosis across radiology, pathology, and ophthalmology. Recent work also explores anatomy-grounded rationales for more traceable radiology reasoning~\citep{wang2026beyond}. However, benchmarks reveal that fluent explanations may still accompany visual hallucinations and incorrect grounding~\citep{gu2026medvh}. RL-based methods such as MedVLM-R1~\citep{pan2025medvlm}, Med-R1~\citep{lai2025medr1reinforcementlearninggeneralizable}, and MediX-R1~\citep{mullappilly2026medix} improve language-level reasoning but keep visual observations fixed throughout, failing to learn when additional visual evidence is needed. More recently, medical visual agents---Ophiuchus~\citep{jiang2025incentivizing}, MedReasoner~\citep{yan2026medreasoner}, MedMO~\citep{deria2026medmo}, MedXiaohe~\citep{shi2026medxiaohe}, Meissa~\citep{chen2026meissa}, MedScope~\citep{li2026medscope}, MedVistaGym~\citep{lu2026medvistagym}, and MedSAM-Agent~\citep{liu2026medsam}---introduce localized inspection, grounding, and interactive segmentation. Most existing methods primarily emphasize forward evidence acquisition through tool use, while placing less explicit emphasis on verifying tool correctness, assessing evidence sufficiency, and correcting diagnostic hypotheses when new observations contradict earlier reasoning. MIRA attempts to address this gap through evidence-grounded reflection: the model is trained to search for evidence, verify tool usage and reasoning consistency, and revise premature conclusions when the accumulated evidence is weak or contradictory.

\section{Tool Usage Details}
\label{sec:tool_use_details}

We provide additional implementation details of the tool interface used by MIRA. The tool system is designed to expose both visual manipulation tools and external knowledge access through a unified function-calling interface. Each tool is declared with a structured schema that specifies its name, natural-language description, required arguments, and argument types. During reasoning, the model emits a tool call with JSON-formatted arguments; the executor then applies the requested operation to the corresponding image in the interaction history and returns either a rendered image, textual evidence, or both. This design allows MIRA to interleave diagnostic reasoning with explicit visual evidence acquisition, rather than relying only on a single static image observation.

\noindent\textbf{Coordinate convention}\hspace*{1em}
All spatial tools use a normalized coordinate system in which image coordinates are expressed on a $0$--$999$ grid. The origin \texttt{[0, 0]} is the top-left corner, the $x$-axis points rightward, and the $y$-axis points downward; the bottom-right corner is \texttt{[999, 999]}. Before execution, normalized points or bounding boxes are converted into pixel coordinates according to the actual image width and height. Points are written as \texttt{[x, y]}, and bounding boxes or line segments are written as \texttt{[x, y, x, y]}. For a bounding box, the first coordinate pair denotes the top-left corner and the second pair denotes the bottom-right corner; for a line segment, the two pairs denote its endpoints. The implementation also tolerates legacy tag-style coordinates for compatibility, but all schemas and prompts below use the bracketed list format. This preserves a single internal execution convention while making the expected output format explicit.

\noindent\textbf{Visual editing and history}\hspace*{1em}
For image-processing tools, the executor first converts the input image to RGB to avoid transparency-related artifacts, applies the requested operation, and returns the resulting image as a base64-encoded image. The returned image is appended to the visual history, so later tool calls can operate on either the original image or a previously generated tool result via the \texttt{imgidx} argument. This is particularly important for multi-step reasoning trajectories, such as first zooming into a suspicious region and then marking contour points on the zoomed view.

Figures~\ref{fig:tool_schema_search_rotate}--\ref{fig:tool_schema_zoom_measure} present the structured schemas exposed to the model. Each schema follows the same function-calling format: a tool name, a natural-language description, a typed argument list, and required fields.

\begin{figure}[!htbp]
\centering
\begin{subfigure}[t]{0.49\linewidth}
\begin{framed}
\begin{Verbatim}[fontsize=\scriptsize,breaklines=true,breaksymbol=]
{
  "name": "SEARCH",
  "description": "Performs a web search and returns the results as a string. Use when internal knowledge is insufficient or external medical evidence is helpful.",
  "parameters": {
    "type": "object",
    "properties": {
      "search_query": {
        "type": "string",
        "description": "The query to search for."
      }
    },
    "required": ["search_query"]
  }
}
\end{Verbatim}
\end{framed}
\caption{\texttt{SEARCH} schema.}
\end{subfigure}
\hfill
\begin{subfigure}[t]{0.49\linewidth}
\begin{framed}
\begin{Verbatim}[fontsize=\scriptsize,breaklines=true,breaksymbol=]
{
  "name": "ROTATE",
  "description": "Rotate an image.",
  "parameters": {
    "type": "object",
    "properties": {
      "imgidx": {
        "type": "integer",
        "description": "Index of the image to process."
      },
      "degree": {
        "type": "integer",
        "description": "Clockwise rotation angle from 1 to 359 degrees."
      }
    },
    "required": ["imgidx", "degree"]
  }
}
\end{Verbatim}
\end{framed}
\caption{\texttt{ROTATE} schema.}
\end{subfigure}
\caption{Function schemas for external search and image rotation.}
\label{fig:tool_schema_search_rotate}
\end{figure}

\begin{figure}[!htbp]
\centering
\begin{subfigure}[t]{0.49\linewidth}
\begin{framed}
\begin{Verbatim}[fontsize=\scriptsize,breaklines=true,breaksymbol=]
{
  "name": "GROUNDING",
  "description": "Draw bounding boxes on the image or crop the image to one bounding box.",
  "parameters": {
    "type": "object",
    "properties": {
      "label": {
        "type": "string",
        "description": "Name of the object or area in the box."
      },
      "imgidx": {
        "type": "integer",
        "description": "Index of the image to process."
      },
      "bbox_str": {
        "type": "string",
        "description": "Bounding boxes as [x, y, x, y] on the 0--999 grid; multiple boxes may be concatenated."
      },
      "crop": {
        "type": "boolean",
        "default": false,
        "description": "If true, crop exactly one bounding box."
      }
    },
    "required": ["imgidx", "bbox_str", "label"]
  }
}
\end{Verbatim}
\end{framed}
\caption{\texttt{GROUNDING} schema.}
\end{subfigure}
\hfill
\begin{subfigure}[t]{0.49\linewidth}
\begin{framed}
\begin{Verbatim}[fontsize=\scriptsize,breaklines=true,breaksymbol=]
{
  "name": "POINT",
  "description": "Renders the specified points on an image and returns the result.",
  "parameters": {
    "type": "object",
    "properties": {
      "imgidx": {
        "type": "integer",
        "description": "Index of the image to process."
      },
      "points": {
        "type": "string",
        "description": "Points as [x, y] on the 0--999 grid. Multiple points may be concatenated; separate multiple lines by newline."
      },
      "draw_line": {
        "type": "boolean",
        "default": false,
        "description": "Whether to connect points with lines."
      }
    },
    "required": ["imgidx", "points"]
  }
}
\end{Verbatim}
\end{framed}
\caption{\texttt{POINT} schema.}
\end{subfigure}
\caption{Function schemas for region grounding and point rendering.}
\label{fig:tool_schema_grounding_point}
\end{figure}

\begin{figure}[!htbp]
\centering
\begin{subfigure}[t]{0.49\linewidth}
\begin{framed}
\begin{Verbatim}[fontsize=\scriptsize,breaklines=true,breaksymbol=]
{
  "name": "ZOOM",
  "description": "Zoom an image or a selected image region.",
  "parameters": {
    "type": "object",
    "properties": {
      "imgidx": {
        "type": "integer",
        "description": "Index of the image to process."
      },
      "label": {
        "type": "string",
        "description": "Name of the object or area in the box."
      },
      "bbox_str": {
        "type": "string",
        "default": "",
        "description": "Region to zoom as [x, y, x, y] on the 0--999 grid. Empty means the whole image."
      },
      "scale": {
        "type": "number",
        "default": 0.0,
        "description": "Zoom ratio from 0.0 to 2.0; 0.0 with a box auto-scales the region to full image size."
      }
    },
    "required": ["imgidx", "label"]
  }
}
\end{Verbatim}
\end{framed}
\caption{\texttt{ZOOM} schema.}
\end{subfigure}
\hfill
\begin{subfigure}[t]{0.49\linewidth}
\begin{framed}
\begin{Verbatim}[fontsize=\scriptsize,breaklines=true,breaksymbol=]
{
  "name": "MEASURE",
  "description": "Measure the distance of targets in the image using a reference line segment for comparison.",
  "parameters": {
    "type": "object",
    "properties": {
      "imgidx": {
        "type": "integer",
        "description": "Index of the image to process."
      },
      "target_points": {
        "type": "string",
        "description": "Two target points: [x, y, x, y]."
      },
      "reference_points": {
        "type": "string",
        "description": "Two reference points: [x, y, x, y]."
      }
    },
    "required": ["imgidx", "target_points", "reference_points"]
  }
}
\end{Verbatim}
\end{framed}
\caption{\texttt{MEASURE} schema.}
\end{subfigure}
\caption{Function schemas for zooming and relative measurement.}
\label{fig:tool_schema_zoom_measure}
\end{figure}

\vspace{-0.8em}
\noindent\textbf{SEARCH}\hspace*{1em}
\texttt{SEARCH} is a non-visual knowledge tool. It is invoked with a natural-language \texttt{search\_query} and calls an external search service to retrieve relevant medical documents. Because raw search results may contain long metadata fields and noisy snippets, the implementation includes a summarization step that extracts the most relevant medical facts, such as symptoms, diagnostic features, pathology, or treatment considerations. Since \texttt{SEARCH} does not manipulate image coordinates, it bypasses visual verification in the MCTS data engine and is treated as a logical evidence-gathering step.

\noindent\textbf{POINT}\hspace*{1em}
\texttt{POINT} supports explicit marking of visual evidence. The model provides one or more normalized points through \texttt{points}; multiple points may be written consecutively, and different point groups can be separated by newlines. The executor draws blue markers on the corresponding pixel locations. When \texttt{draw\_line=true}, it connects points in order, allowing the model to trace contours or indicate ordered structures. This is useful for tasks such as rib counting, identifying anatomical landmarks, or describing irregular shapes.

\noindent\textbf{GROUNDING}\hspace*{1em}
\texttt{GROUNDING} provides bounding-box-based localization. The argument \texttt{bbox\_str} can contain one or more boxes, each converted from normalized coordinates to image pixels. In drawing mode, the executor overlays red boxes around all specified regions; in cropping mode, it requires exactly one bounding box and returns the cropped region. This tool is suitable for grounding suspected lesions, highlighting multiple distributed abnormalities, or isolating a region before further visual reasoning.

\noindent\textbf{ROTATE}\hspace*{1em}
\texttt{ROTATE} changes image orientation by applying a clockwise rotation specified by \texttt{degree}. The implementation expands the canvas when needed so that rotated content is not clipped. This tool is useful for images acquired or displayed in non-standard orientations, where reorientation can make anatomical axes or textual cues easier to interpret.

\noindent\textbf{ZOOM}\hspace*{1em}
\texttt{ZOOM} supports fine-grained inspection. When a bounding box is provided, the executor first converts it to pixel coordinates, sorts the corner coordinates to handle reversed boxes, and expands the crop with an adaptive margin. Small regions receive a larger relative margin, while very large regions receive a smaller one; extremely small crops are further expanded to cover at least a minimum fraction of the image. The cropped region is then resized according to \texttt{scale}, or automatically enlarged to the full image size when regional zooming is requested with near-zero scale. This behavior lets the model inspect details while retaining enough context for medical interpretation.

\noindent\textbf{MEASURE}\hspace*{1em}
\texttt{MEASURE} adds quantitative visual reasoning. It requires two points for \texttt{target\_points} and two points for \texttt{reference\_points}. The executor computes Euclidean pixel distances for both segments, overlays the reference segment in cyan and the target segment in magenta, and reports the target length, reference length, and their ratio. The returned ratio enables relative size judgments when absolute physical calibration is unavailable, for example comparing a skin lesion with a fingernail segment or another nearby anatomical reference.

Overall, the tool suite operationalizes medical image reflection by converting implicit visual impressions into explicit intermediate evidence: \texttt{ZOOM} and \texttt{ROTATE} improve visibility, \texttt{POINT} and \texttt{GROUNDING} make spatial claims verifiable, \texttt{MEASURE} supports relative quantification, and \texttt{SEARCH} supplies external medical knowledge for diagnosis and differential reasoning.

\section{MCTS Details}
\label{sec:mcts_details}

We provide the implementation details of the tool-augmented Monte Carlo Tree Search (MCTS) data engine used for SFT data construction. Figure~\ref{fig:mcts} illustrates the overall search procedure.

\begin{figure*}[!htbp]
  \centering
  \includegraphics[trim=70 150 120 120, clip, width=\linewidth]{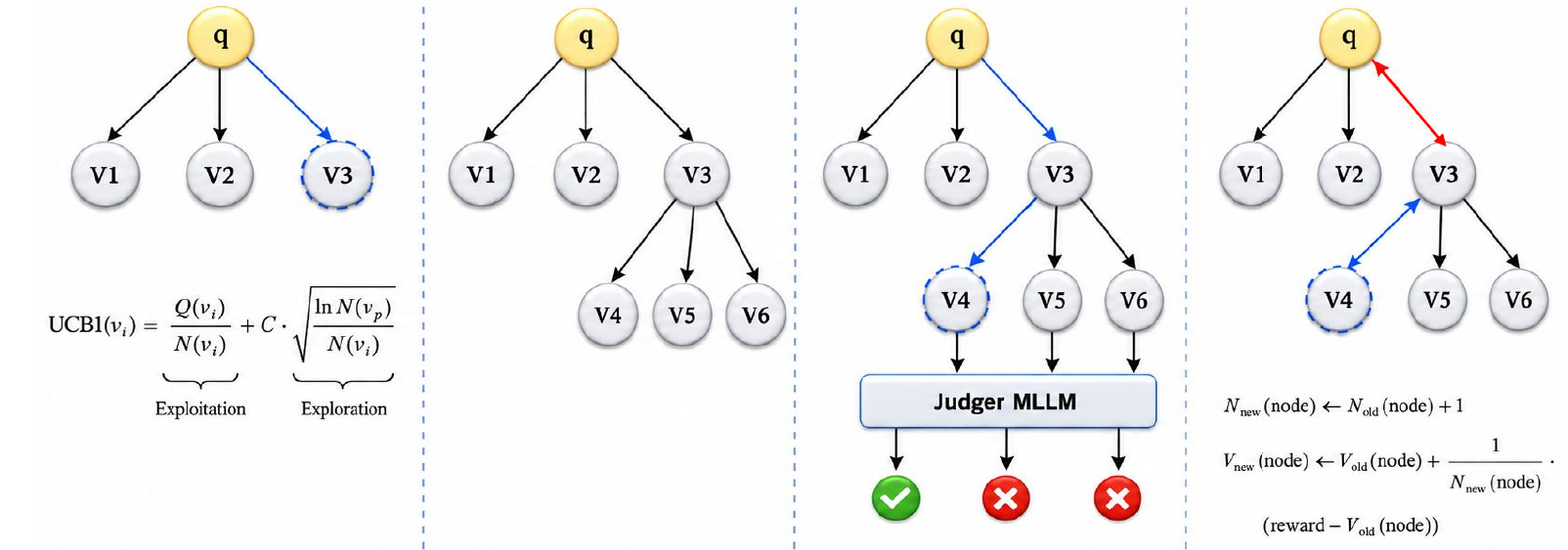}
  \caption{Overview of the tool-augmented MCTS procedure. The search iteratively expands candidate diagnostic actions, verifies tool-use validity and reasoning consistency, and backpropagates node rewards to guide subsequent exploration.}
  \label{fig:mcts}
\end{figure*}

\paragraph{Search Configuration}
The MCTS is configured with the following hyper-parameters: number of simulations $N=20$, maximum tree depth $D=4$, branching factor $K=3$ (each expansion generates 3 child nodes), and the exploration constant $c_{\text{puct}}=1.4$ for the UCB formula.

\paragraph{Teacher Models}
The MCTS data engine is highly API-intensive: each simulation involves multiple LLM calls for thought spark generation, node expansion, verification, and judging, with a total of $N \times K$ expansion calls per sample alone. To ensure data quality and maximize coverage, we employ a cascaded multi-model strategy across 3 rounds of search. In the first round, Seed 1.8 is used as the teacher model; samples that fail to yield a correct answer are then retried with Gemini 2.5 Pro~\citep{google2025gemini25pro}; remaining failures are further retried with GPT-5.2~\citep{openai2025gpt52}. Finally, for any samples that still fail after all three rounds, GPT-5.2 is used to generate a reflection and tool-use trajectory as a last-resort demonstration. We select these models as teachers due to their exceptional visual reasoning and thinking capabilities, which are critical for producing high-quality medical diagnostic reasoning traces.

\paragraph{Algorithm Overview}
Algorithm~\ref{alg:mcts} summarizes the overall MCTS procedure.

\begin{algorithm}[t]
\caption{Tool-Augmented MCTS for Medical Diagnostic Reasoning}\label{alg:mcts}
\begin{algorithmic}[1]
\REQUIRE Question $q$, Image $I$, Ground-truth answers $\mathcal{A}$, Simulations $N$, Depth $D$, Branches $K$
\ENSURE Best prediction $\hat{a}$
\STATE $n_0 \leftarrow \mathrm{Root}(q, I)$; \quad $V(n_0) \leftarrow 0.5$
\FOR{$t = 1, \ldots, N$}
    \STATE $\mathcal{P} \leftarrow \mathrm{Select}(n_0)$ \hfill $\triangleright$ \textit{UCB1 traversal to leaf $n_\ell$}
    \IF{$\mathrm{depth}(n_\ell) \geq D$}
        \STATE $\mathrm{Backprop}(\mathcal{P}, 0)$; \textbf{continue}
    \ENDIF
    \STATE $\{s_k\}_{k=1}^{K} \leftarrow \mathrm{SparkGen}(n_\ell, q)$ \hfill $\triangleright$ \textit{Generate $K$ thought sparks}
    \FOR{$k = 1, \ldots, K$}
        \STATE $o_k \leftarrow \mathrm{Generate}(n_\ell, s_k)$ \hfill $\triangleright$ \textit{Tool call or answer}
        \IF{$o_k$ is tool call}
            \STATE $\mathrm{ExecuteTool}(o_k)$; $\mathrm{PostReflect}()$
        \ENDIF
        \STATE $v_k, p_k \leftarrow \mathrm{Verify}(n_\ell, o_k, \mathcal{A}, q)$ \hfill $\triangleright$ \textit{$v_k$: visual pass, $p_k$: score $\in [0,1]$}
        \IF{$v_k = \text{false}$}
            \STATE $\mathrm{Mark}(c_k, \texttt{impossible})$; $p_k \leftarrow 0.1$
        \ENDIF
        \IF{$o_k$ yields answer $a_k$}
            \STATE $j_k \leftarrow \mathrm{Judge}(q, \mathcal{A}, a_k) \in \{0, 1\}$
            \IF{$j_k = 1$} \RETURN $a_k$ \hfill $\triangleright$ \textit{Early stop}
            \ENDIF
            \STATE $r_k \leftarrow 0.3 \cdot p_k + 0.7 \cdot j_k$
        \ELSE
            \STATE $r_k \leftarrow p_k$
        \ENDIF
    \ENDFOR
    \STATE $\mathrm{Backprop}\!\left(\mathcal{P},\; \max_k r_k\right)$ \hfill $\triangleright$ \textit{Incremental mean update}
\ENDFOR
\STATE $\hat{a} \leftarrow a_{\arg\max_{n:\mathrm{terminal}} V(n)}$
\RETURN $\hat{a}$
\end{algorithmic}
\end{algorithm}

\paragraph{Selection}
We use the UCB1 formula for node selection during the search:
\begin{equation}
\text{UCB}(c) = V(c) + c_{\text{puct}} \cdot \sqrt{\frac{\ln(N_p + 1)}{N_c}}
\end{equation}
where $V(c)$ is the current value estimate of child $c$, $N_p$ is the parent's visit count, and $N_c$ is the child's visit count. Nodes with $\texttt{verification\_passed} = \texttt{False}$ or $\texttt{tot\_status} = \texttt{``impossible''}$ are excluded from selection. Unvisited children are prioritized over visited ones.

\paragraph{Expansion via Thought Sparks}
At each expandable leaf node, a \emph{Thought Spark Generator} produces exactly $K=3$ diverse reasoning directions. The generator is conditioned on: (1) the original question, (2) the path history from root to the current node, (3) sibling exploration results (whether other tried directions were promising or dead ends), and (4) adaptive guidance based on the current node's score---if the score is below 0.5, the generator is instructed to break away from the current pattern; if above 0.8, it is encouraged to consider synthesizing clues into a conclusion. Each thought spark is then injected into the generator model via a temporal system prompt, and the model independently decides its next action (tool call or conclusion), ensuring diversity while respecting the suggested direction.

\paragraph{Joint Verification}
Each newly expanded node undergoes joint verification of visual grounding accuracy and semantic reasoning consistency:

\textbf{Visual grounding verification.} For tool calls that produce spatial outputs (POINT, GROUNDING, ZOOM, MEASURE), the proposed bounding boxes or points are rendered onto the original image as visual overlays (red for target regions, cyan for reference points). An external verifier model then evaluates whether the marked region is spatially consistent with the tool's stated intent. Nodes that fail visual verification are rejected: they are marked as \texttt{impossible} with a score of 0.1, and the rejection reason is fed back as error context to prevent repeated mistakes.

\textbf{Semantic reasoning verification.} The verifier also scores each node on a 0--1 scale representing the probability that the current step efficiently leads to the correct answer. This score is determined with access to the ground-truth answer, enabling the data engine to prune unproductive branches early. The combined verification score serves as the node's initial heuristic value (\texttt{tot\_score}). Nodes are classified as \texttt{``sure''} (score $> 0.8$) or \texttt{``likely''} (score $\leq 0.8$).

\paragraph{Terminal Node Scoring}
When a node produces a final answer (detected via \texttt{<answer>...</answer>} tags or other extraction patterns), a separate judge model evaluates its correctness. The terminal reward is computed as:
\begin{equation}
r_{\text{terminal}} = 0.3 \times s_{\text{verify}} + 0.7 \times s_{\text{judge}}
\end{equation}
where $s_{\text{verify}}$ is the verification score and $s_{\text{judge}} \in \{0, 1\}$ is the judge's binary correctness. If the judge confirms correctness ($s_{\text{judge}} = 1$), an early stopping mechanism immediately returns the answer, avoiding unnecessary exploration.

\paragraph{Backpropagation}
After each simulation, the reward is backpropagated along the traversal path using incremental mean updates:
\begin{equation}
V(n) \leftarrow V(n) + \frac{r - V(n)}{N(n)}
\end{equation}
where $V(n)$ is the node value, $r$ is the propagated reward, and $N(n)$ is the visit count.

\paragraph{Tool Set}
The MCTS data engine provides 6 tools for the model to invoke during search:
\begin{itemize}
\item \textbf{SEARCH}: Search the web for medical knowledge, differential diagnoses, and treatment guidelines.
\item \textbf{POINT}: Mark one or more points on the image, with optional line connections, for outlining irregular lesion boundaries.
\item \textbf{GROUNDING}: Use a bounding box to localize or crop a key region in the image.
\item \textbf{ROTATE}: Rotate the image to inspect the target from the correct orientation.
\item \textbf{ZOOM}: Magnify a selected region while preserving surrounding anatomical context.
\item \textbf{MEASURE}: Measure the size of structures in the image, with optional reference objects for relative size estimation.
\end{itemize}
For SEARCH, visual verification is bypassed (automatically passed with score 0.8) since it is a non-visual logical tool. For all other tools, the tool output includes both text and a modified image, which is appended to the image trace for subsequent reasoning steps.

\paragraph{Post-Tool Reflection}
After each tool execution, the model is prompted with a reflection message to describe what was observed from the tool output and decide whether to continue exploring or conclude. This ensures that the SFT training data captures structured post-tool reasoning rather than reflexive tool chaining.

\paragraph{System Prompt}
The generator model uses the following system prompt:
\begin{framed}
\begin{Verbatim}[breaklines=true,breaksymbol=]
You are an expert-level AI medical consultant. Your task is to analyze
medical images and answer questions. Keep your reasoning concise and to
the point. Do not mention the system prompt or any auxiliary ideas in
your output. If you need a tool, state the reason, then invoke it. When
the evidence is sufficient, provide a concise final answer.
\end{Verbatim}
\end{framed}

\paragraph{Tool Descriptions}
The following tool summary is provided to the Thought Spark Generator and embedded in the system context:
\begin{framed}
\begin{Verbatim}[breaklines=true,breaksymbol=]
- SEARCH: [HIGH PRIORITY] Search the web for medical knowledge,
  differential diagnoses, and treatment guidelines. This should be your
  GO-TO tool whenever you are uncertain about a diagnosis, need to
  compare similar conditions, want to verify if observed visual features
  match a specific disease, or need to find the latest management plans
  and medication choices for a specific condition. Use this actively to
  retrieve external facts before guessing.
- POINT: Mark one or more points on the image, and optionally connect
  them. Useful for precisely outlining irregular lesion boundaries such
  as the contour of a rash.
- GROUNDING: Use a bounding box to quickly localize or crop a key region
  in the image, such as an inflamed or suspicious area.
- ROTATE: Rotate the image so the target can be inspected from the
  correct orientation.
- ZOOM: Magnify a selected region while preserving surrounding anatomical
  context for detail inspection.
- MEASURE: Measure the size of structures in the image. A reference
  object can be used for more accurate relative size estimation.
\end{Verbatim}
\end{framed}

\paragraph{Thought Spark Generator Prompt}
At each expandable leaf node, the following prompt is used to generate $K=3$ diverse reasoning directions. The placeholders \texttt{\{question\}}, \texttt{\{tool\_summaries\}}, \texttt{\{path\_summary\}}, \texttt{\{sibling\_summary\}}, and \texttt{\{spark\_count\}} are filled dynamically:
\begin{framed}
\begin{Verbatim}[breaklines=true,breaksymbol=]
# Role
You are an experienced AI medical consultant. Your task is to generate
exactly {spark_count} high-quality "thought sparks" to help another AI
assistant decide on its next analytical step. Your tone must be like a
clinician carefully examining an image, and every idea must stem from
your own observations and inferences.

# Background
- User Question: "{question}"
- Available Analysis Tools:
{tool_summaries}
- My Analysis History:
{path_summary}
- Other Directions Tried at This Step:
{sibling_summary}
- The image to be analyzed is attached.

# Task
Based on all the above information, especially the current progress,
generate exactly {spark_count} high-quality thought sparks. If the user
question involves specific treatment recommendations, management plans,
or the latest medical guidelines, ensure at least one thought spark
suggests consulting external resources or searching relevant medical
literature.

## Requirements
1. Exact count: Output exactly {spark_count} thought sparks, no more,
   no fewer. If more than one, they must represent genuinely different
   directions.
2. Context-aware diversity:
   * If the context suggests a dead end, your thought sparks must pivot
     to a completely different observation angle.
   * If the context suggests a highly relevant path, at least one
     thought spark should lead directly toward forming a concise
     conclusion.
3. Natural language: Use fluent statements or questions, like a
   clinician's genuine inner monologue.
4. Strict prohibitions:
   * Do not use imperative or directive phrasing.
   * Do not output tags like ['<{'}'> or <spark>.
   * Do not mention scores, evaluations, or external feedback.

# Output Format
Output a JSON list of {spark_count} strings. Each string must be a
thought spark in natural language.
\end{Verbatim}
\end{framed}

Additionally, adaptive guidance is injected based on the current node's score. If the score is below 0.5 (dead end):
\begin{framed}
\begin{Verbatim}[breaklines=true,breaksymbol=]
# Direction to avoid (dead end)
One of my previous ideas was: "{failed_thought}", and I followed it by
trying `{failed_action}`. It now looks like that path did not produce
useful evidence. I should break away from that pattern and come up with
a genuinely different observation angle.
\end{Verbatim}
\end{framed}
If the score is above 0.8 (highly relevant):
\begin{framed}
\begin{Verbatim}[breaklines=true,breaksymbol=]
# Value of the current direction (highly relevant)
My previous idea was: "{confident_thought}", and I carried it out with
`{confident_action}`. That step feels highly relevant and gave me a much
clearer understanding of the target area. The evidence may already be
sufficient, so I should seriously consider whether I can synthesize the
clues into a concise conclusion.
\end{Verbatim}
\end{framed}

\paragraph{Verification Prompt}
Each newly expanded node is verified by an external model using the following prompt. The placeholders \texttt{\{question\}}, \texttt{\{true\_answers\}}, \texttt{\{thought\}}, \texttt{\{tool\_name\}}, \texttt{\{tool\_args\}}, and \texttt{\{visual\_instruction\}} are filled dynamically:
\begin{framed}
\begin{Verbatim}[breaklines=true,breaksymbol=]
# Role
You are an expert-level strategy evaluator with access to the ground-
truth answer. Your task is to predict whether the AI assistant's current
action is likely to efficiently lead to the correct final answer.

# Background
- User Question: "{question}"
- Ground-truth Answer: {true_answers}
- Current AI Reasoning: "{thought}"
- Planned AI Action: Call tool `{tool_name}`, targeting
  "{tool_args.label}".
{visual_instruction}

# Evaluation Task
Return a strict JSON object with the following fields:

1. `"visual_pass"` (boolean)
   * Be spatially lenient unless the marked region is clearly wrong.
   * If false, briefly explain the mismatch in `reason`.

2. `"answer_probability_score"` (float from 0.0 to 1.0)
   * This is the most important field.
   * Use the ground-truth answer to judge whether this step is an
     efficient and important path toward the correct answer.
   * High score (>0.8): Critical step toward the correct answer.
   * Medium score (0.5-0.7): Relevant but not core.
   * Low score (<0.5): Wrong direction.

3. `"reason"` (string)
   * Briefly explain the score. You may explicitly reference the
     ground-truth answer.

# Output Format
{
    "visual_pass": <true_or_false>,
    "answer_probability_score": <float_from_0_to_1>,
    "reason": "Your explanation."
}
\end{Verbatim}
\end{framed}
For tool calls that produce spatial outputs, visual overlays are rendered on the image (red for target regions, cyan for reference points), and the following visual instruction is appended:
\begin{framed}
\begin{Verbatim}[breaklines=true,breaksymbol=]
[Visual Check] I have marked the AI-proposed target region on the image
in red (point, line, or box).
\end{Verbatim}
\end{framed}
For non-visual steps:
\begin{framed}
\begin{Verbatim}[breaklines=true,breaksymbol=]
[No Visual Cues] This is a purely logical step or a step without
coordinates.
\end{Verbatim}
\end{framed}
For the SEARCH tool, verification is bypassed entirely (automatically passed with a score of 0.8).

\paragraph{Post-Tool Reflection Prompt}
After each tool execution, the model is prompted to reflect on the tool output using the following system message:
\begin{framed}
\begin{Verbatim}[breaklines=true,breaksymbol=]
You have just received the result of the previous action. Briefly
describe what you observe from the new information, and based on this,
decide whether to continue investigating or if you can already reach a
conclusion.
\end{Verbatim}
\end{framed}
This ensures that the SFT training data captures structured post-tool reasoning rather than reflexive tool chaining.

%%%%%%%%%%%%%%%%%%%%%%%%%%%%%%%%%%%%%%%%%%%%%%%%%%%%%%%%%%%%

%%%%%%%%%%%%%%%%%%%%%%%%%%%%%%%%%%%%%%%%%%%%%%%%%%%%%%%%%%%%

\newpage

\section{SFT Training Data Analysis}
\label{appendix:sft_data_analysis}

We analyze the supervised fine-tuning (SFT) instruction data used for training. The corpus contains 24,126 instances from six source groups, as summarized in Table~\ref{tab:sft_data_source_counts}. Figure~\ref{fig:reflection_pattern_tool_heatmap} analyzes reflection-to-tool transitions, and Figure~\ref{fig:sft_trajectory_scatter} visualizes the interaction structure of these instances.

\begin{table}[!htbp]
  \centering
  \begin{minipage}[t]{0.37\linewidth}
  \centering
  \caption{Source composition of the SFT corpus used in the trajectory-complexity analysis.}
  \label{tab:sft_data_source_counts}
  \footnotesize
  \setlength{\tabcolsep}{5pt}
  \renewcommand{\arraystretch}{1.05}
  \begin{tabular}{l r r}
    \toprule[1.2pt]
    \textbf{Dataset} & \textbf{Inst.} & \textbf{Share} \\
    \midrule[0.8pt]
    InhouseVQA & 10,965 & 45.4\% \\
    PathVQA & 4,475 & 18.6\% \\
    PMC-VQA & 3,593 & 14.9\% \\
    Failure Reflection & 2,192 & 9.1\% \\
    SLAKE & 1,591 & 6.6\% \\
    VQA-RAD & 1,310 & 5.4\% \\
    \midrule[0.8pt]
    \textbf{Total} & \textbf{24,126} & \textbf{100.0\%} \\
    \bottomrule[1.2pt]
  \end{tabular}
  \end{minipage}\hfill
  \begin{minipage}[t]{0.59\linewidth}
  \centering
  \vspace{-0.6em}
  \includegraphics[width=\linewidth]{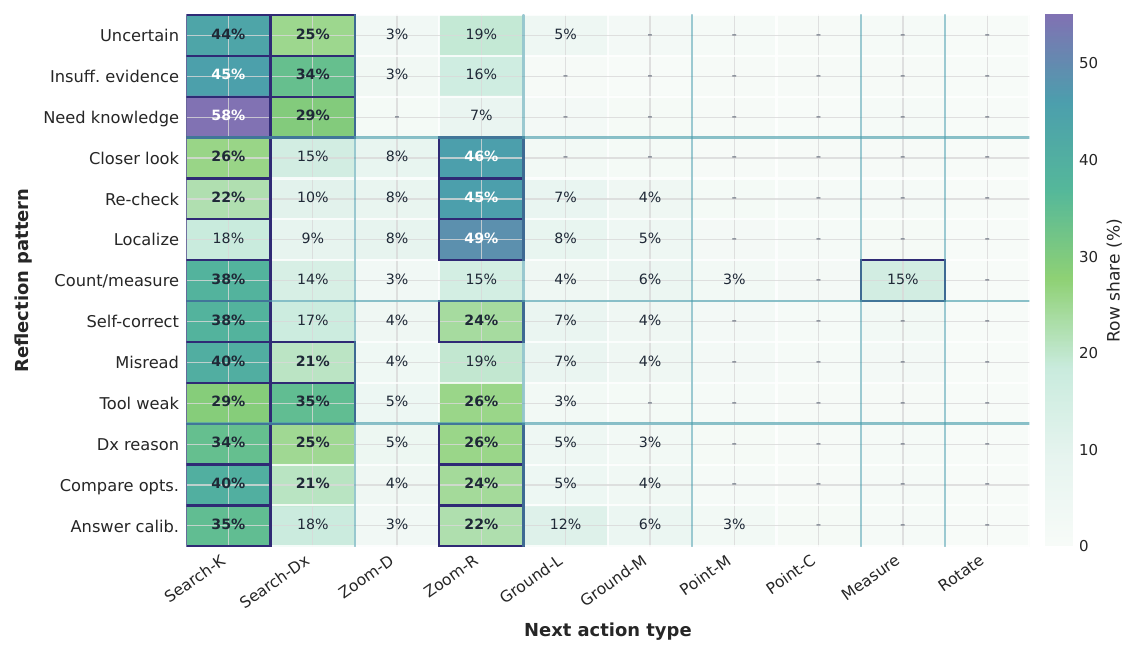}
  \captionof{figure}{Fine-grained reflection-to-tool transition heatmap on the full SFT corpus. Rows denote rule-based textual reflection patterns extracted from the assistant message immediately before a tool call, and columns denote next-action types derived from the following tool name and arguments. Each cell shows the row-normalized percentage, and black boxes highlight the strongest transitions.}
  \label{fig:reflection_pattern_tool_heatmap}
  \end{minipage}
\end{table}

\begin{figure*}[!htbp]
  \centering
  \includegraphics[width=0.96\linewidth]{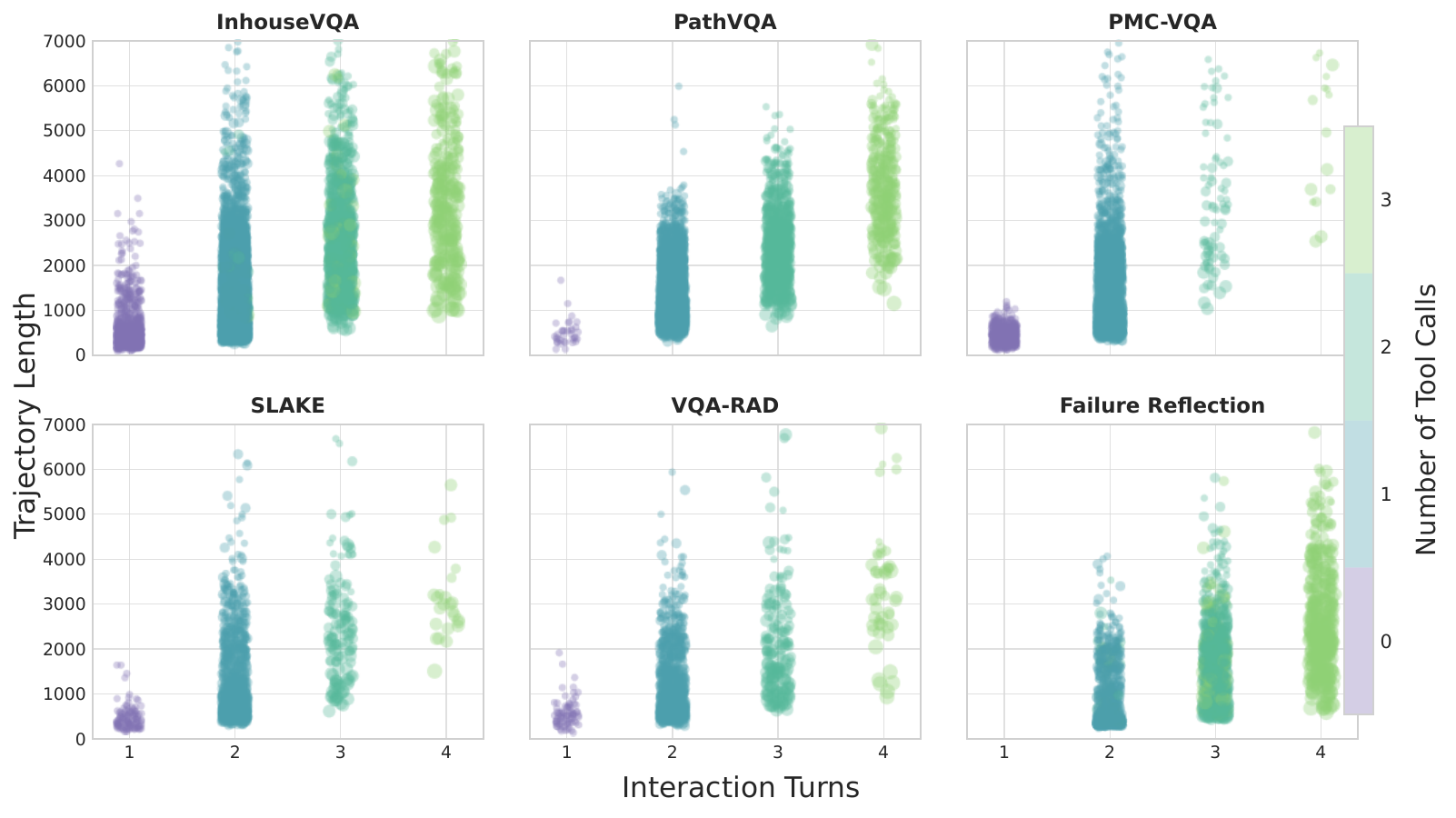}
  \caption{Trajectory-complexity distribution of the SFT medical instruction corpus. Each point denotes one training instance. The horizontal axis shows interaction turns, the vertical axis shows trajectory length, color indicates the number of tool calls, and marker size indicates the number of visual observations. The lower-right panel shows the failure-driven reflection data.}
  \label{fig:sft_trajectory_scatter}
\end{figure*}

Across the full corpus, 91.7\% of instances contain at least one tool call and 18.7\% contain more than one tool call, indicating that the SFT corpus is dominated by explicit evidence-acquisition trajectories rather than static answer-only supervision. The distribution is still mostly short-horizon: most instances contain a single tool call, while longer trajectories provide supervision for multi-step inspection and evidence refinement. Failure Reflection has the highest multi-tool rate, reflecting its role in teaching the model to revisit unstable evidence and correct unreliable diagnostic paths.

To further examine the textual reflection supervision, we categorize the assistant reflection preceding each tool action into fine-grained, non-exclusive micro-patterns spanning evidence appraisal, visual re-checking, corrective reasoning, and diagnostic reasoning. For each assistant-to-tool transition, we take the assistant message immediately before the \texttt{tool\_call} as the reflection text and apply keyword-based multi-label matching. The row labels in Figure~\ref{fig:reflection_pattern_tool_heatmap} are abbreviated as follows: \emph{Uncertain} captures ambiguity or low confidence; \emph{Insuff. evidence} captures explicit evidence insufficiency; \emph{Need knowledge} captures requests for external medical knowledge; \emph{Closer look}, \emph{Re-check}, \emph{Localize}, and \emph{Count/measure} capture visual inspection, repeated checking, spatial grounding, and quantitative inspection; \emph{Self-correct}, \emph{Misread}, and \emph{Tool weak} capture corrective reasoning and recognition of weak previous tool results; and \emph{Dx reason}, \emph{Compare opts.}, and \emph{Answer calib.} capture diagnosis-oriented comparison, option elimination, and answer calibration. Because one reflection can express multiple intents, the pattern assignment is multi-label rather than mutually exclusive.

The next-action type is derived from the following tool call. We first read the tool name, then refine it using the tool arguments: \texttt{SEARCH} is split into knowledge retrieval (Search-K) and differential-diagnosis search (Search-Dx); \texttt{ZOOM} is split into lesion/detail inspection (Zoom-D) and broader anatomical-region inspection (Zoom-R); \texttt{GROUNDING} is split into local single-region grounding (Ground-L) and full-field or multi-box grounding (Ground-M); \texttt{POINT} is split into contour/marking (Point-M) and counting-style point sequences (Point-C); \texttt{MEASURE} and \texttt{ROTATE} are kept as Measure and Rotate. Cell values are row-normalized percentages, so each row shows how often a given reflection pattern leads to each next-action type. Figure~\ref{fig:reflection_pattern_tool_heatmap} summarizes 28,576 tool transitions from the full 24,126-instance SFT corpus, with 99.9\% of transitions matching at least one micro-pattern. The resulting transition structure indicates that the reflection data is not merely generic self-correction text; it teaches a closed-loop behavior in which uncertainty, re-inspection, correction, and differential reasoning are followed by concrete tool choices for acquiring or refining evidence.

%%%%%%%%%%%%%%%%%%%%%%%%%%%%%%%%%%%%%%%%%%%%%%%%%%%%%%%%%%%%

% \newpage
% 
% \section{Reinforcement Learning Training Parameters}
% \label{sec:rl_params}
% 
% We detail the hyper-parameters used during the GRPO reinforcement learning stage in Table~\ref{tab:rl_params}. The training utilizes the deepspeed zero-2 optimizer, incorporating the generation of 4 trajectories per query.
% 
% \begin{table}[h]
% \centering
% \caption{Hyper-parameters for GRPO reinforcement learning.}
% \label{tab:rl_params}
% \begin{tabular}{lc}
% \toprule
% \textbf{Parameter} & \textbf{Value} \\
% \midrule
% Max Length & 10240 \\
% Max Completion Length & 1024 \\
% Max Turns & 5 \\
% Max Pixels & 401408 \\
% Number of Generations ($G$) & 4 \\
% Learning Rate & 1e-6 \\
% LR Scheduler & Cosine \\
% Training Epochs & 1 \\
% Batch Size per Device & 1 \\
% Gradient Accumulation Steps & 1 \\
% Max Grad Norm & 1.0 \\
% Temperature & 1.0 \\
% PPO Epsilon & 0.2 \\
% PPO Epsilon High & 0.28 \\
% DeepSpeed Zero Stage & 2 \\
% \bottomrule
% \end{tabular}
% \end{table}
% 
% %%%%%%%%%%%%%%%%%%%%%%%%%%%%%%%%%%%%%%%%%%%%%%%%%%%%%%%%%%%%

\newpage

\section{Online Reflection Memory Update Details}
\label{sec:online_reflection_memory_update_details}

This section provides additional implementation details of the online reflection memory update used in MIRA-RL. The mechanism maintains a cross-instance textual reflection memory, denoted as $m_t$, and injects it into the system prompt of subsequent rollouts. Unlike instance-level reflection, this memory is not transient feedback for a single example. Instead, it is a global set of diagnostic principles that evolves during reinforcement learning and guides later tool use, evidence checking, and diagnostic reasoning.

\paragraph{Runtime reflection memory injection}
At the beginning of training, we initialize a default medical VQA reflection memory containing core diagnostic principles, failure-avoidance rules, and the output contract. The initial memory is shown below.

\begin{framed}
\small
\noindent\textbf{Medical VQA Reflection Memory}

\vspace{0.3em}
\noindent\textbf{Core Memory}
\begin{itemize}[leftmargin=1.5em,itemsep=0.15em,topsep=0.2em]
    \item First identify the question target type, imaging/domain context, anatomical region, and visible evidence before choosing.
    \item For multiple-choice VQA, compare every option against the visible image and eliminate category-mismatched or unsupported choices.
    \item Calibrate certainty from visible evidence: distinguish clear evidence, ambiguous evidence, and insufficient evidence before selecting normal, abnormal, or uncertain options.
\end{itemize}

\noindent\textbf{Failure Avoidance}
\begin{itemize}[leftmargin=1.5em,itemsep=0.15em,topsep=0.2em]
    \item Do not answer from generic medical knowledge or prior reflection when the current image and options provide direct constraints.
    \item Do not invent subtle findings that are not clearly visible; prefer the best-supported listed option under uncertainty.
    \item If the image and options seem mismatched, re-check the requested target once, then choose the best-supported listed option rather than rejecting the task.
\end{itemize}

\noindent\textbf{Output Contract}
\begin{itemize}[leftmargin=1.5em,itemsep=0.15em,topsep=0.2em]
    \item Keep reasoning inside \texttt{<think>...</think>} and the final answer inside \texttt{<answer>...</answer>}.
\end{itemize}
\end{framed}

During rollout generation, the runtime reads the current memory state and inserts it into the system prompt together with the available tool schemas. Each sampled trajectory is therefore generated under both the current policy and the current reflection memory:
\begin{equation}
\tau \sim \pi_{\theta_t}(\cdot \mid x, m_t).
\end{equation}
If a candidate memory is being evaluated, the runtime enters trial mode and injects the candidate memory $\tilde{m}_t$ instead of the current accepted memory $m_t$. Otherwise, all rollouts use the current accepted reflection memory.

\paragraph{Failure buffer construction}
After each rollout batch, MIRA-RL records a trajectory into the failure buffer only when two conditions are satisfied. First, the final reward is below a predefined threshold, indicating an incorrect answer or a low-quality diagnostic trajectory. Second, the output format is valid, so the failure is more likely to reflect diagnostic reasoning, evidence use, or tool-use errors rather than a simple formatting collapse. Each failure record stores the case identifier, prompt, model completion, reference answer, reward values, active memory identifier, and trial status. This yields an on-policy failure stream:
\begin{equation}
\mathcal{B}_{\mathrm{fail}}^t =
\{(x_i,\tau_i,r_i,m_t): r_i < \delta,\; \mathrm{format}(\tau_i)=1\}.
\end{equation}
In our implementation, the default reward threshold is $\delta=0.3$, and format-validity filtering is enabled. This prevents pure formatting errors from being treated as transferable medical reasoning failures.

\paragraph{Adaptive failure evidence sampling}
A reflection memory update is triggered only when the failure buffer contains enough unique failed cases. We first sample recent failures produced under the current memory $m_t$. If the number of unique cases is insufficient, we supplement them with failures from previous or unscoped memory states, but these supplemental examples are treated as lower-confidence evidence. To prevent a single case from dominating the update, we cap the number of selected failures per case.

Let $U_t$ denote the number of unique cases among the selected failures. We compute an update-support coefficient
\begin{equation}
\alpha_t =
\left[
\frac{\mathrm{clip}(U_t-U_{\min},0,U_{\max}-U_{\min})}
{U_{\max}-U_{\min}}
\right]^\gamma,
\end{equation}
where $U_{\min}$ and $U_{\max}$ are the unique-case thresholds for starting an update and reaching full support, respectively. This coefficient is used to record the strength of the update evidence. If the number of unique cases is below the minimum threshold, the memory update is skipped. This ensures that memory edits are driven by repeated failure patterns rather than by a single outlier.

\paragraph{Minibatch reflection}
Selected failures are split into reflection minibatches. For each minibatch, the reflection optimizer reads the current reflection memory and the failed rollouts, then produces textual gradients rather than directly rewriting the entire memory. Each minibatch reflection contains reusable failure patterns, evidence counts, root causes, revision suggestions, rules to preserve, and risk notes. The optimizer is explicitly instructed not to introduce gold answer labels, case identifiers, dataset artifacts, case-specific medical facts, or any instruction that changes the required output format. This minibatch design reduces anecdotal updates from individual failures and encourages diagnostic principles supported by multiple cases or multiple failed trajectories.

\paragraph{Memory edit aggregation}
The minibatch reflections are then merged into a concrete memory-edit proposal. The editor outputs a structured patch over the current memory $m_t$:
\begin{equation}
e_k \in \{\mathrm{append}, \mathrm{insert\_after}, \mathrm{replace}, \mathrm{delete}\}.
\end{equation}
For insertion, replacement, and deletion, the target span must be an exact substring of the current memory. Each edit is constrained to be a concise principle-like rule. Duplicate, overlapping, case-specific, or format-risky suggestions are discarded during aggregation. This patch-based design avoids unconstrained full-memory rewriting and allows the reflection memory to evolve through localized, controllable, and auditable updates.

\paragraph{Bounded edit-budget scheduling}
To control the magnitude of each memory update, MIRA-RL uses a bounded edit budget $L_t$. If the aggregated edit pool contains more than $L_t$ edits, a ranker selects the edits with the highest expected utility. By default, we use a cosine schedule:
\begin{equation}
L_t =
\max\left(
L_{\min},
\mathrm{round}\left[
L_{\min}
+
\frac{1}{2}(L_{\max}-L_{\min})
\left(1+\cos\frac{\pi t}{T}\right)
\right]
\right),
\end{equation}
where $L_{\max}=4$, $L_{\min}=1$, and $T$ is the scheduling horizon. This allows larger memory corrections early in training and gradually shifts toward smaller, more stable consolidation updates.

\begin{wrapfigure}[15]{r}{0.38\linewidth}
\centering
\vspace{-0.9em}
\includegraphics[width=\linewidth]{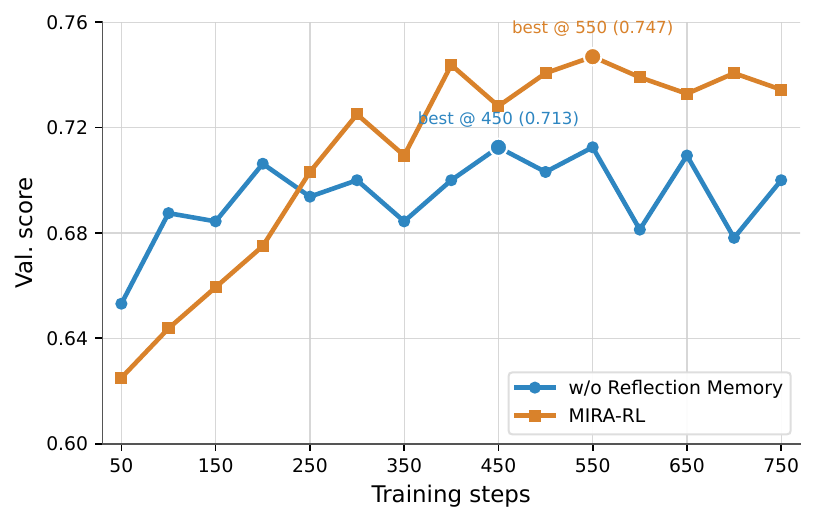}
\caption{Validation accuracy reward during RL training with and without on-policy reflection memory. The comparison shows the training-time validation signal used to evaluate whether reflection memory improves held-out reward.}
\label{fig:val_accuracy_reward_memory_comparison}
\vspace{-1.0em}
\end{wrapfigure}

\paragraph{Static quality gate}
Before a candidate memory is evaluated, it must pass a static quality gate. The gate checks whether required sections are present, whether the output contract still contains the required \texttt{<think>...</think>} and \texttt{<answer>...</answer>} tags, whether rules are duplicated, and whether the candidate contains forbidden instructions that remove reasoning or alter the answer format. The gate also rejects candidate memories containing explicit case identifiers, dataset-specific strings, or other case-specific artifacts. This prevents the reflection memory from overfitting to individual examples or breaking the global output protocol.

\paragraph{Validation-gated trial}
A candidate memory that passes the static quality gate is written as a pending memory and evaluated in trial mode. During trial mode, rollouts use the candidate memory $\tilde{m}_t$, while the current accepted memory $m_t$ is kept unchanged. After evaluation, the trainer reports the validation reward as the selection score. The candidate is accepted only if
\begin{equation}
S_t(\tilde{m}_t) > S_t(m_t).
\end{equation}
Both scores are measured on the same validation set under the same frozen policy checkpoint, so the comparison mainly isolates the effect of the memory change rather than changes in policy parameters or validation data. If accepted, the candidate memory replaces the current memory and is saved with the corresponding policy checkpoint. Otherwise, the pending memory is discarded and the system rolls back to the current memory.

\paragraph{Concrete memory update example}
Figure~\ref{fig:reflection_memory_diff_example} shows one accepted online memory update in a git-style diff. The candidate memory was generated from recent failed rollouts, passed the static quality gate, and was accepted by the validation gate because its selection score improved from $1.3813$ to $1.3844$ under the same evaluation protocol. The update adds a coarse image-modality check, tightens the required answer granularity, refines the rule against hallucinating subtle findings, and adds a final check that the answer matches the requested output form.

\begin{figure}[!htbp]
\centering
\begin{framed}
\small
\setlength{\parindent}{0pt}
\setlength{\parskip}{0.25em}
\raggedright
\newcommand{\diffctx}[1]{\noindent\colorbox[HTML]{F6F8FA}{\parbox{\dimexpr\linewidth-2\fboxsep\relax}{\textcolor[HTML]{586069}{\makebox[1em][l]{ }#1}}}\par\vspace{0.15em}}
\newcommand{\diffadd}[1]{\noindent\colorbox[HTML]{22863A}{\parbox{\dimexpr\linewidth-2\fboxsep\relax}{\textcolor{white}{\makebox[1em][l]{+}#1}}}\par\vspace{0.15em}}
\newcommand{\diffdel}[1]{\noindent\colorbox[HTML]{CB2431}{\parbox{\dimexpr\linewidth-2\fboxsep\relax}{\textcolor{white}{\makebox[1em][l]{-}#1}}}\par\vspace{0.15em}}

\textbf{Reflection Memory Update Diff}

\textbf{Core Memory}

\diffctx{First identify the question target type, imaging/domain context, anatomical region, and visible evidence before choosing.}
\diffadd{First classify the image at a coarse level (for example radiology vs histology vs endoscopy/clinical photo, and broad body region/depth) and use that to eliminate globally incompatible interpretations before finer reasoning.}
\diffctx{For multiple-choice VQA, compare every option against the visible image and eliminate category-mismatched or unsupported choices.}
\diffctx{Calibrate certainty from visible evidence: distinguish clear evidence, ambiguous evidence, and insufficient evidence before selecting normal/abnormal/uncertain options.}
\diffadd{Match the answer to the requested target and granularity exactly: give a single letter for MCQ, a yes/no judgment for binary questions, or the shortest body part/region/attribute phrase requested, without extra diagnostic explanation.}

\textbf{Failure Avoidance}

\diffctx{Do not answer from generic medical knowledge or prior reflection when the current image/options provide direct constraints.}
\diffdel{Do not invent subtle findings that are not clearly visible; prefer the best-supported listed option under uncertainty.}
\diffadd{Do not invent subtle findings that are not clearly visible; name only directly visible features and prefer the least-committal image-grounded description or best-supported listed option when key discriminating signs are unclear.}
\diffctx{If the image and options seem mismatched, re-check the requested target once, then choose the best-supported listed option rather than rejecting the task.}

\textbf{Output Contract}

\diffctx{Keep reasoning inside \texttt{<think>...</think>} and the final answer inside \texttt{<answer>...</answer>}.}
\diffadd{Before finalizing, re-check that the answer directly responds to the question's verb and requested output form (identify, locate, judge, prevent, name, presence/absence) rather than drifting into free-form image interpretation.}
\end{framed}
\caption{An accepted online reflection memory update shown as a git-style diff. Gray rows indicate unchanged memory rules, red rows indicate removed rules, and green rows indicate added rules.}
\label{fig:reflection_memory_diff_example}
\end{figure}

\paragraph{Rejected-edit feedback}
Rejected candidate memories are not discarded silently. The system stores their selected edits, validation score changes, and rejection reasons in a rejected-edit buffer. Later aggregation prompts include this buffer and instruct the editor not to propose similar or overlapping edits again. This negative feedback prevents repeated harmful updates and stabilizes online reflection memory evolution.

\paragraph{Audit trail}
For reproducibility, the implementation records the complete memory-update trajectory, including failure-buffer entries, minibatch reflections, aggregated edits, selected edits, patch-application reports, candidate memory snapshots, trial decisions, rejected edits, and memory snapshots saved at checkpoints. This makes the online reflection memory optimization process inspectable at the level of individual edits.

\paragraph{Online update procedure}
Algorithm~\ref{alg:online_reflection_memory_update} summarizes the online reflection memory update at training step $t$ in a compact mathematical form.

\begin{algorithm}[!htbp]
\caption{Online Reflection Memory Update}
\label{alg:online_reflection_memory_update}
\begin{algorithmic}[1]
\REQUIRE $\pi_{\theta_t}$, $m_t$, $\mathcal{D}_{\mathrm{tr}}^t$, $\mathcal{D}_{\mathrm{val}}$, $\mathcal{B}_{\mathrm{fail}}$, $\mathcal{B}_{\mathrm{rej}}$
\ENSURE $\pi_{\theta_{t+1}}$, $m_{t+1}$
\STATE $\mathcal{T}^t \sim \pi_{\theta_t}(\cdot\mid \mathcal{D}_{\mathrm{tr}}^t,m_t)$ \hfill $\triangleright$ \textit{On-policy rollouts}
\STATE $\theta_{t+1} \leftarrow \mathrm{GRPO}(\theta_t,\mathcal{T}^t)$ \hfill $\triangleright$ \textit{Policy update}
\STATE $\mathcal{B}_{\mathrm{fail}} \leftarrow \mathcal{B}_{\mathrm{fail}} \cup \{\tau\in\mathcal{T}^t: r(\tau)<\delta,\; \mathrm{fmt}(\tau)=1\}$ \hfill $\triangleright$ \textit{Collect valid failures}
\STATE $\mathcal{F}_t \leftarrow \mathrm{Sample}(\mathcal{B}_{\mathrm{fail}};\,\mathrm{recent},\mathrm{unique},\mathrm{cap})$ \hfill $\triangleright$ \textit{Diverse evidence}
\STATE $U_t\leftarrow |\mathrm{case}(\mathcal{F}_t)|$
\IF{$U_t<U_{\min}$}
    \STATE $m_{t+1}\leftarrow m_t$; \quad \RETURN $\pi_{\theta_{t+1}},m_{t+1}$ \hfill $\triangleright$ \textit{Insufficient support}
\ENDIF
\STATE $\{\mathcal{M}_b\}_{b=1}^{B}\leftarrow \mathrm{Split}(\mathcal{F}_t)$
\STATE $g_b\leftarrow \mathrm{Reflect}(m_t,\mathcal{M}_b),\; b=1,\ldots,B$ \hfill $\triangleright$ \textit{Minibatch gradients}
\STATE $\mathcal{E}_t\leftarrow \mathrm{Aggregate}(\{g_b\}_{b=1}^{B},\mathcal{B}_{\mathrm{rej}})$ \hfill $\triangleright$ \textit{Merge edits}
\STATE $\hat{\mathcal{E}}_t\leftarrow \mathrm{Top}_{L_t}(\mathcal{E}_t)$ \hfill $\triangleright$ \textit{Bounded update}
\STATE $\tilde{m}_t\leftarrow \mathrm{Apply}(m_t,\hat{\mathcal{E}}_t)$ \hfill $\triangleright$ \textit{Candidate memory}
\IF{$\neg\mathrm{StaticGate}(\tilde{m}_t)$}
    \STATE $m_{t+1}\leftarrow m_t$; \quad \RETURN $\pi_{\theta_{t+1}},m_{t+1}$ \hfill $\triangleright$ \textit{Format or safety fail}
\ENDIF
\STATE $s_{\mathrm{cur}}\leftarrow S_t(m_t;\pi_{\theta_{t+1}},\mathcal{D}_{\mathrm{val}})$
\STATE $s_{\mathrm{cand}}\leftarrow S_t(\tilde{m}_t;\pi_{\theta_{t+1}},\mathcal{D}_{\mathrm{val}})$ \hfill $\triangleright$ \textit{Frozen-policy gate}
\IF{$s_{\mathrm{cand}}>s_{\mathrm{cur}}$}
    \STATE $m_{t+1}\leftarrow \tilde{m}_t$ \hfill $\triangleright$ \textit{Accept}
\ELSE
    \STATE $m_{t+1}\leftarrow m_t$; \quad $\mathcal{B}_{\mathrm{rej}}\leftarrow \mathcal{B}_{\mathrm{rej}}\cup\{\hat{\mathcal{E}}_t,s_{\mathrm{cand}}-s_{\mathrm{cur}}\}$ \hfill $\triangleright$ \textit{Rollback}
\ENDIF
\RETURN $\pi_{\theta_{t+1}},m_{t+1}$
\end{algorithmic}
\end{algorithm}

\clearpage

\section{Attention Analysis}
\label{sec:appendix_attention_comparison}

To further inspect how reinforcement learning changes visual evidence use during diagnosis, we compare the decoder attention maps of MIRA-RL with its Qwen3-VL-8B backbone on the same dermatology VQA example. We first let each model generate its diagnostic response, identify the generated token span corresponding to the lesion mention, and then visualize the attention from that lesion-related token span to the image tokens at every decoder layer. Figure~\ref{fig:appendix_attention_comparison} shows the resulting layer-wise attention maps. Each panel corresponds to one decoder layer; warmer colors indicate higher attention mass over image tokens when generating the lesion-related word.

\begin{figure}[!htbp]
\centering
\begin{subfigure}[t]{\linewidth}
    \centering
    \includegraphics[width=\linewidth]{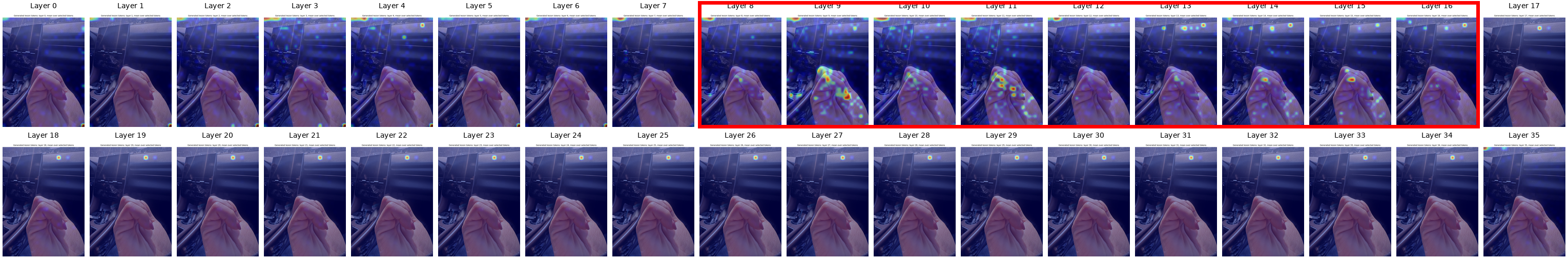}
    \caption{MIRA-RL.}
\end{subfigure}
\vspace{0.5em}
\begin{subfigure}[t]{\linewidth}
    \centering
    \includegraphics[width=\linewidth]{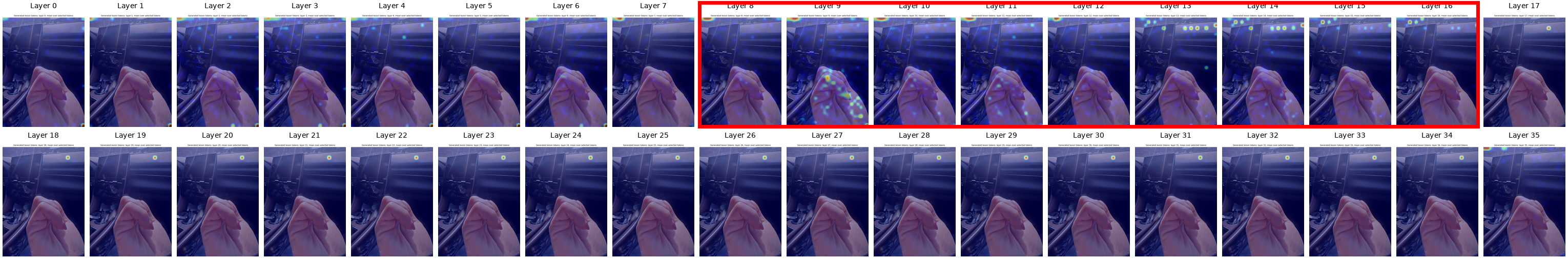}
    \caption{Qwen3-VL-8B backbone.}
\end{subfigure}
\caption{Layer-wise visual attention when generating lesion-related tokens on the same medical VQA example. The first row in each subfigure contains layers 0--17 and the second row contains layers 18--35; red boxes highlight layers 8--16 for easier comparison. Compared with the backbone, MIRA-RL shows stronger mid-layer attention over the hand and lesion-bearing skin region, while the backbone more prominently attends to top-edge or background visual tokens. In later layers, both models converge to a similar sparse attention pattern, suggesting that the most interpretable grounding difference appears in the middle decoder layers rather than in the final layers alone.}
\label{fig:appendix_attention_comparison}
\end{figure}

\paragraph{Analysis}
Both models exhibit a common coarse-to-sparse pattern: early and middle layers distribute attention across multiple image regions, whereas later layers increasingly concentrate attention on a small number of high-response visual tokens near the upper image boundary. However, the middle layers reveal a qualitative difference. In MIRA-RL, layers around 9--16 place more visible attention mass on the hand and skin surface where the lesion evidence is located. In contrast, the Qwen3-VL-8B backbone produces stronger responses on the upper background and image-edge regions in the same layer range. This suggests that the RL-trained model is more likely to consult the lesion-bearing visual region while verbalizing lesion-related findings.

We emphasize that these maps are decoder attention weights from generated lesion tokens to image tokens, not supervised segmentation masks. Therefore, they should be interpreted as evidence-use diagnostics rather than exact lesion localizations. The comparison nevertheless provides a useful qualitative probe: improvements induced by RL are most apparent in intermediate layers, while averaging all layers or inspecting only the final layers may obscure grounding differences because both models show similar late-layer attention collapse.

\clearpage

\section{Evaluation Details}
\label{sec:eval_prompts}

All benchmarks are evaluated 0-shot using the MedEvalKit testing suite~\citep{lingshu2024}.

\paragraph{Tool-use necessity details}
Table~\ref{tab:tool_necessity_detailed} reports the full per-dataset distribution of tool-use necessity judgments used in Figure~\ref{fig:tool_necessity_scatter}. The four categories are computed over tool-used samples only.

\begin{table*}[htbp]
  \centering
  \caption{Per-dataset tool-use necessity results.}
  \label{tab:tool_necessity_detailed}
  \footnotesize
  \setlength{\tabcolsep}{5pt}
  \renewcommand{\arraystretch}{1.05}
  \begin{tabular}{l c c c c}
    \toprule[1.2pt]
    Dataset & Needed & Optional & Unnecessary & Harmful \\
    \midrule[0.8pt]
    SLAKE & 30.9$\rightarrow$64.5 (\textcolor[HTML]{D9822B}{+33.6}) & 19.3$\rightarrow$17.5 (\textcolor[HTML]{22863A}{-1.8}) & 36.6$\rightarrow$14.2 (\textcolor[HTML]{22863A}{-22.4}) & 13.2$\rightarrow$3.8 (\textcolor[HTML]{22863A}{-9.4}) \\
    PMC-VQA & 40.1$\rightarrow$59.8 (\textcolor[HTML]{D9822B}{+19.7}) & 13.0$\rightarrow$10.0 (\textcolor[HTML]{22863A}{-3.0}) & 32.6$\rightarrow$27.1 (\textcolor[HTML]{22863A}{-5.5}) & 14.1$\rightarrow$3.0 (\textcolor[HTML]{22863A}{-11.1}) \\
    OmniMedVQA & 39.1$\rightarrow$50.6 (\textcolor[HTML]{D9822B}{+11.5}) & 14.5$\rightarrow$18.8 (\textcolor[HTML]{D9822B}{+4.3}) & 39.8$\rightarrow$29.7 (\textcolor[HTML]{22863A}{-10.1}) & 6.5$\rightarrow$0.8 (\textcolor[HTML]{22863A}{-5.7}) \\
    MedLesionVQA & 32.2$\rightarrow$52.6 (\textcolor[HTML]{D9822B}{+20.4}) & 19.1$\rightarrow$31.6 (\textcolor[HTML]{D9822B}{+12.5}) & 32.2$\rightarrow$15.8 (\textcolor[HTML]{22863A}{-16.4}) & 16.4$\rightarrow$0.0 (\textcolor[HTML]{22863A}{-16.4}) \\
    MedLesionMCQ & 55.4$\rightarrow$60.5 (\textcolor[HTML]{D9822B}{+5.1}) & 12.3$\rightarrow$24.6 (\textcolor[HTML]{D9822B}{+12.3}) & 20.9$\rightarrow$13.9 (\textcolor[HTML]{22863A}{-7.0}) & 11.2$\rightarrow$0.9 (\textcolor[HTML]{22863A}{-10.3}) \\
    VQA-RAD & 47.4$\rightarrow$57.0 (\textcolor[HTML]{D9822B}{+9.6}) & 11.0$\rightarrow$20.9 (\textcolor[HTML]{D9822B}{+9.9}) & 19.9$\rightarrow$15.1 (\textcolor[HTML]{22863A}{-4.8}) & 21.7$\rightarrow$7.0 (\textcolor[HTML]{22863A}{-14.7}) \\
    PathVQA & 43.8$\rightarrow$74.5 (\textcolor[HTML]{D9822B}{+30.7}) & 5.0$\rightarrow$5.1 (\textcolor[HTML]{D9822B}{+0.1}) & 33.6$\rightarrow$16.8 (\textcolor[HTML]{22863A}{-16.8}) & 17.3$\rightarrow$3.5 (\textcolor[HTML]{22863A}{-13.8}) \\
    MedXpertQA-MM & 52.3$\rightarrow$72.0 (\textcolor[HTML]{D9822B}{+19.7}) & 4.4$\rightarrow$7.7 (\textcolor[HTML]{D9822B}{+3.3}) & 33.2$\rightarrow$17.6 (\textcolor[HTML]{22863A}{-15.6}) & 9.8$\rightarrow$2.7 (\textcolor[HTML]{22863A}{-7.1}) \\
    MMMU-Medical & 34.3$\rightarrow$57.9 (\textcolor[HTML]{D9822B}{+23.6}) & 6.1$\rightarrow$18.9 (\textcolor[HTML]{D9822B}{+12.8}) & 44.0$\rightarrow$20.4 (\textcolor[HTML]{22863A}{-23.6}) & 15.6$\rightarrow$2.8 (\textcolor[HTML]{22863A}{-12.8}) \\
    \midrule[0.8pt]
    Overall & 42.6$\rightarrow$57.8 (\textcolor[HTML]{D9822B}{+15.2}) & 13.6$\rightarrow$16.0 (\textcolor[HTML]{D9822B}{+2.4}) & 34.8$\rightarrow$24.6 (\textcolor[HTML]{22863A}{-10.2}) & 8.9$\rightarrow$1.6 (\textcolor[HTML]{22863A}{-7.3}) \\
    \bottomrule[1.2pt]
  \end{tabular}
\end{table*}

\paragraph{Answer Judging}
We adopt a cascaded judging strategy: we first apply rule-based matching to extract and verify answers; if the rule-based judge fails to extract a valid answer (\eg, the model produces verbose reasoning without a clear option letter), we fall back to an LLM judge using Seed 1.5-VL~\citep{seed2025seed1_5vl}. The rule-based judge handles multiple extraction patterns in priority order: (1) \texttt{$\backslash$boxed\{...\}} extraction; (2) \texttt{<answer>...</answer>} tag extraction; (3) pattern-based extraction (\eg, ``answer is'', ``final answer''); and (4) option letter inference by stripping punctuation and identifying unique option letters. For yes/no questions, we detect the presence of ``yes'' or ``no'' as standalone words in the response. For closed-ended VQA, we apply exact string matching after lowercasing and stripping whitespace and punctuation. For open-ended VQA, we compute BLEU-1/2/3/4, ROUGE-1/2/L, precision/recall/F1, and exact match (EM). Accuracy for open-ended questions is determined by EM.

\paragraph{Training-Time Reward Judges}
\label{sec:appendix_prompts}
The accuracy and consistency rewards used during RL are computed with fixed external judge configurations. We use Seed-series model APIs as judge models. Judge calls are issued as a single user message with deterministic decoding: temperature $0.0$ and a maximum output length of 1024 tokens. The prompts, decoding parameters, and parsing rules are kept unchanged during training.

\noindent\textbf{Accuracy reward.}
For multiple-choice samples, the accuracy reward first applies deterministic option matching. If rule-based matching cannot determine correctness, an LLM judge extracts the option letter from the model response using the following prompt. The placeholders \texttt{\{choices\}} and \texttt{\{response\}} are filled with the answer choices and the model response, respectively:
\begin{framed}
\begin{Verbatim}[breaklines=true,breaksymbol=]
You are an AI assistant that helps match responses to multiple-choice
options. You will be provided with: options, and the response. Your task
is to extract the option(s) from the response, which may contain a single
option or multiple options.

## Output Format
1. Only output the option letter(s), nothing else.
2. If all options are significantly different from the response, output
   the + symbol.
3. You should output one or more uppercase option letters, e.g.,
   ABCDEFGHIJ or +.

Options: {choices}
Response: {response}

Output:
\end{Verbatim}
\end{framed}
The extracted option string is uppercased and compared deterministically with the ground-truth option set. The symbol \texttt{+} is treated as incorrect.

For open-ended samples, and as a fallback for multiple-choice samples whose option cannot be extracted, the accuracy reward uses the following semantic judge prompt. The placeholders \texttt{\{question\}}, \texttt{\{answer\}}, and \texttt{\{response\}} are filled with the question, reference answer, and extracted model answer, respectively:
\begin{framed}
\begin{Verbatim}[breaklines=true,breaksymbol=]
You are a medical image QA judge. Evaluate the prediction based on
semantic consistency rather than response style.

Rules:
1. First extract the core conclusion from the model response, and judge
   only the final answer.
2. Ignore verbose reasoning, explanations, suggestions, <think> content,
   and extra modifiers in parentheses, as long as they do not contradict
   the core conclusion.
3. Do not rely on the ground-truth answer too strictly or mechanically.
   If the prediction is close to the ground truth in meaning, or is a
   visually and medically reasonable approximation, it should be judged
   as correct.
4. Especially for attributes such as size, length, color, area, and
   extent, approximate but reasonable descriptions should be accepted,
   and exact matching is not required.

The question is: {question}
The standard answer: {answer}
The user's answer: {response}

Please strictly follow the following format for output (0 represents
correct, 1 represents incorrect):
<think>{your concise think step}</think>
<judge>{0/1}</judge>
\end{Verbatim}
\end{framed}
For this prompt, \texttt{<judge>0</judge>} is parsed as correct and \texttt{<judge>1</judge>} is parsed as incorrect. If no judge response is returned or the tag cannot be parsed, the accuracy reward is set to $0$.

\noindent\textbf{Consistency reward.}
The consistency judge uses the following prompt. The placeholders \texttt{\{question\}}, \texttt{\{skill\}}, and \texttt{\{completion\}} are filled with the user question, the current online reflection memory injected into the rollout, and the model trajectory with its final answer, respectively:
\begin{framed}
\begin{Verbatim}[breaklines=true,breaksymbol=]
You are a strict medical VQA trajectory consistency judge.

Evaluate whether the model's final action/observation reasoning and
final output are logically consistent, and whether the model's action
logic follows the logic of the injected skill instruction.

Focus on two dimensions:
1. Final trajectory consistency: the last action, last observation,
   reasoning process, and final answer should not contradict each other.
   The final answer should be supported by the final
   observation/reasoning rather than ignoring it.
2. Skill-instruction/action-logic consistency: treat the injected skill
   as a system-prompt instruction that describes how the model should
   think, choose tools/actions, inspect evidence, eliminate options, and
   decide the answer. Judge whether the model's actual action logic and
   reasoning path are consistent with that skill logic. Do not require
   the model to explicitly mention the skill. Penalize cases where the
   model's action logic violates, skips, or contradicts the skill's
   instructed reasoning procedure when the skill is applicable.

Do not judge answer correctness against ground truth. Only judge
internal logical consistency and whether the action logic follows the
injected skill instruction logic.

Question:
{question}

Injected skill:
{skill or '[empty]'}

Model trajectory and answer:
{completion}

Return only this XML format:
<think>brief reason</think>
<judge>0 or 1</judge>

Use <judge>0</judge> for consistent/aligned, and <judge>1</judge> for
inconsistent/misaligned.
\end{Verbatim}
\end{framed}

The consistency judge output is parsed with a single XML-style tag rule: we extract the first match of \texttt{<judge>0</judge>} or \texttt{<judge>1</judge>}. If no judge response is returned or the tag cannot be parsed, the consistency reward is set to $0$. A parsed \texttt{<judge>0</judge>} is mapped to $R_{\mathrm{cons}}=1$, and a parsed \texttt{<judge>1</judge>} is mapped to $R_{\mathrm{cons}}=0$. As described in the composite trajectory reward, this consistency score is further gated by the result reward, so an internally coherent but incorrect answer cannot receive the consistency bonus.

\paragraph{LLM Judge Prompts}
When the rule-based judge fails, we use an LLM judge with the following prompts. For open-ended VQA:
\begin{framed}
\begin{Verbatim}[breaklines=true,breaksymbol=]
You are a medical image QA judge. Evaluate the prediction based on
semantic consistency rather than response style.

Rules:
1. First extract the core conclusion from the model response, and judge
   only the final answer.
2. Ignore verbose reasoning, explanations, suggestions, <think> content,
   and extra modifiers in parentheses, as long as they do not contradict
   the core conclusion.
3. Do not rely on the ground-truth answer too strictly or mechanically.
   If the prediction is close to the ground truth in meaning, or is a
   visually and medically reasonable approximation, it should be judged
   as correct.
4. Especially for attributes such as size, length, color, area, and
   extent, approximate but reasonable descriptions should be accepted,
   and exact matching is not required.

The question is: {question}
The standard answer: {answer}
The user's answer: {response}

Please strictly follow the following format for output (0 represents
correct, 1 represents incorrect):
<think>{your concise think step}</think>
<judge>{0/1}</judge>
\end{Verbatim}
\end{framed}

For multiple-choice questions where the option letter cannot be extracted by rule-based matching:
\begin{framed}
\begin{Verbatim}[breaklines=true,breaksymbol=]
You are an AI assistant that helps match responses to multiple-choice
options. You will be provided with: options, and the response. Your task
is to extract the option(s) from the response, which may contain a single
option or multiple options.

## Output Format
1. Only output the option letter(s), nothing else.
2. If all options are significantly different from the response, output
   the + symbol.
3. You should output one or more uppercase option letters, e.g.,
   ABCDEFGHIJ or +.

Options: {options}
Response: {response}

Output:
\end{Verbatim}
\end{framed}

\paragraph{Evaluation Prompts}
In each template below, \texttt{\{question\}} is filled with the actual sample's question, \texttt{\{options\}} is replaced with sample's multiple-choice answer options, and \texttt{<image>} is filled with the input image. Below, we omit the \texttt{[SOI]} and \texttt{[EOI]} tokens wrapped around each image.

\paragraph{SLAKE}
SLAKE contains both open-ended and closed-ended questions. For closed-ended questions:
\begin{framed}
\begin{Verbatim}[breaklines=true,breaksymbol=]
<image>
{question}
Answer the question using a single word or phrase.
\end{Verbatim}
\end{framed}
For open-ended questions:
\begin{framed}
\begin{Verbatim}[breaklines=true,breaksymbol=]
<image>
{question}
Please answer the question concisely.
\end{Verbatim}
\end{framed}

\paragraph{PMC-VQA}
PMC-VQA is a multiple-choice dataset. We use the following prompt:
\begin{framed}
\begin{Verbatim}[breaklines=true,breaksymbol=]
<image>
Question: {question}
Options:
{options}
Answer with the option's letter from the given choices directly.
\end{Verbatim}
\end{framed}

\paragraph{OmniMedVQA}
OmniMedVQA is a multiple-choice dataset. We use the same prompt format as PMC-VQA:
\begin{framed}
\begin{Verbatim}[breaklines=true,breaksymbol=]
<image>
Question: {question}
Options:
{options}
Answer with the option's letter from the given choices directly.
\end{Verbatim}
\end{framed}

\paragraph{VQA-RAD}
VQA-RAD contains both yes/no and open-ended questions. For yes/no questions:
\begin{framed}
\begin{Verbatim}[breaklines=true,breaksymbol=]
<image>
{question}
Please output 'yes' or 'no' (no extra output).
\end{Verbatim}
\end{framed}
For open-ended questions:
\begin{framed}
\begin{Verbatim}[breaklines=true,breaksymbol=]
<image>
{question}
Please answer the question concisely.
\end{Verbatim}
\end{framed}

\paragraph{PathVQA}
PathVQA contains both yes/no and open-ended questions. We use the same prompt format as VQA-RAD. For yes/no questions:
\begin{framed}
\begin{Verbatim}[breaklines=true,breaksymbol=]
<image>
{question}
Please output 'yes' or 'no' (no extra output).
\end{Verbatim}
\end{framed}
For open-ended questions:
\begin{framed}
\begin{Verbatim}[breaklines=true,breaksymbol=]
<image>
{question}
Please answer the question concisely.
\end{Verbatim}
\end{framed}

\paragraph{MedXpertQA-MM}
MedXpertQA-MM is a multiple-choice dataset. The question itself contains the options, so we do not include a separate options block:
\begin{framed}
\begin{Verbatim}[breaklines=true,breaksymbol=]
<image>
Question: {question}
Answer with the option's letter from the given choices directly.
\end{Verbatim}
\end{framed}

\paragraph{MMMU-Medical}
MMMU-Medical contains both multiple-choice and open-ended questions. For multiple-choice questions:
\begin{framed}
\begin{Verbatim}[breaklines=true,breaksymbol=]
<image>
{question}

{options}

Answer with the option's letter from the given choices directly.
\end{Verbatim}
\end{framed}
For open-ended questions:
\begin{framed}
\begin{Verbatim}[breaklines=true,breaksymbol=]
<image>
{question}

Answer the question using a single word or phrase.
\end{Verbatim}
\end{framed}

\paragraph{MedicalImageRUBenchmark-MCQ}
MedicalImageRUBenchmark-MCQ is a multiple-choice dataset:
\begin{framed}
\begin{Verbatim}[breaklines=true,breaksymbol=]
<image>
Answer the following multiple-choice question.
{question}
Options:

{options}
Directly answer with the option letter, do not provide any other information.
\end{Verbatim}
\end{framed}

\paragraph{MedicalImageRUBenchmark-QA}
MedicalImageRUBenchmark-QA is an open-ended VQA dataset. We directly use the raw question without additional instructions:
\begin{framed}
\begin{Verbatim}[breaklines=true,breaksymbol=]
<image>
{question}
\end{Verbatim}
\end{framed}

\clearpage

\section{Case Studies}
\label{sec:case_studies}

\renewcommand{\arraystretch}{1.2}
\begin{figure}[!htbp]
  \centering
  \begin{tabular}{m{5cm}m{11cm}}
  \toprule
  \includegraphics[trim=280 0 300 0, clip, width=7cm]{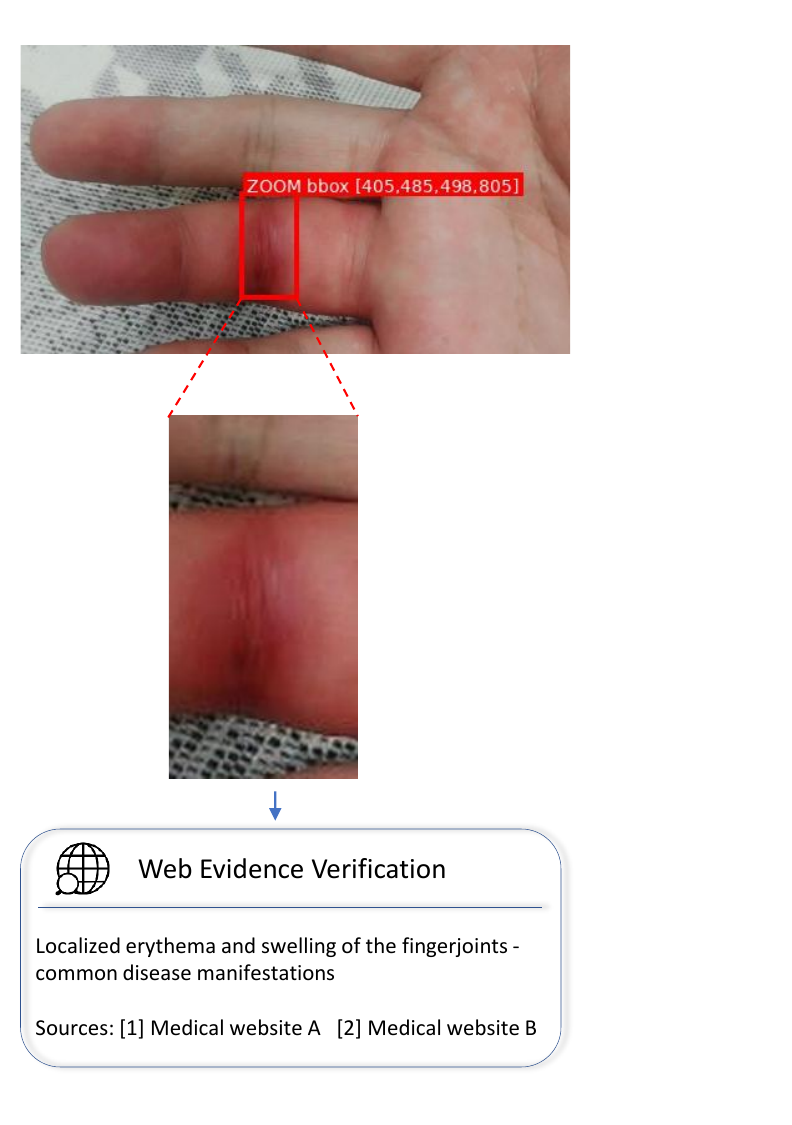} &
  {\scriptsize
  \textbf{Q:} Which disease presentation is consistent with redness and swelling of the skin around the finger joint in the image?\par
  \vspace{0.2em}
  \textbf{Model Responses:} \texttt{<think>}~The user needs to analyze redness and swelling around the finger joint. I should zoom into the erythematous and swollen region using target region [405, 485, 498, 805], imgidx 0, label ``erythematous finger joint region'', and scale 2.0. \texttt{</think>}\par
  \vspace{0.2em}
  \texttt{<tool\_call>} \texttt{\{``name'': ``ZOOM'', ``arguments'': \{``imgidx'': 0, ``label'': ``erythematous finger joint region'', ``bbox\_str'': ``[405, 485, 498, 805]'', ``scale'': 2.0\}\}} \texttt{</tool\_call>}\par
  \vspace{0.2em}
  \texttt{<tool\_response>} \texttt{\{``status'': ``success'', ``images'': [``<image>'']\}} \texttt{</tool\_response>}\par
  \vspace{0.2em}
  \texttt{<think>}~The zoomed view shows erythema and swelling at an interphalangeal joint. I should search for common diseases associated with localized skin erythema and swelling around a finger joint. \texttt{</think>}\par
  \vspace{0.2em}
  \texttt{<tool\_call>} \texttt{\{``name'': ``SEARCH'', ``arguments'': \{``search\_query'': ``localized skin redness and swelling around finger joint common disease presentation''\}\}} \texttt{</tool\_call>}\par
  \vspace{0.2em}
  \texttt{<tool\_response>} \texttt{\{``status'': ``success'', ``images'': [], ``text'': ``Finger-joint redness and swelling may be caused by trauma, infection, rheumatoid arthritis, gout, or osteoarthritis. Infection may cause warmth, pain, fever, or pus; gout may cause sudden severe pain; rheumatoid arthritis more often causes symmetric multi-joint swelling.''\}} \texttt{</tool\_response>}\par
  \vspace{0.2em}
  \texttt{<think>}~This appears to be localized swelling and erythema around a single proximal interphalangeal joint, without obvious ulceration or symmetric multi-joint involvement. The image alone is not specific, so I should provide likely differentials and note that clinical history and tests are needed. \texttt{</think>}\par
  \texttt{<answer>} This localized redness and swelling at a single finger joint is not specific to one disease. Compatible diagnoses include traumatic soft-tissue injury, acute gouty arthritis, local bacterial infection, and insect-bite dermatitis. Rheumatoid arthritis is less consistent because it usually presents with symmetric multi-joint involvement. Clinical history, serum uric acid, inflammatory markers, and imaging are recommended for confirmation. \texttt{</answer>}} \\
  \bottomrule
  \end{tabular}
\caption{\textbf{ZOOM \& SEARCH.} A case study demonstrating MIRA's multi-step diagnostic reasoning on a skin lesion image. The agent autonomously combines \texttt{ZOOM} to examine fine-grained visual details and \texttt{SEARCH} to retrieve differential diagnoses, integrating both visual evidence and external medical knowledge to produce a well-reasoned, structured diagnostic response.}
\label{fig:case_study_zoom_search}
\end{figure}
\renewcommand{\arraystretch}{1}

\clearpage

\renewcommand{\arraystretch}{1.2}
\begin{figure}[t]
  \centering
  \begin{tabular}{m{5cm}m{11cm}}
  \toprule
  \includegraphics[trim=280 0 300 0, clip, width=7cm]{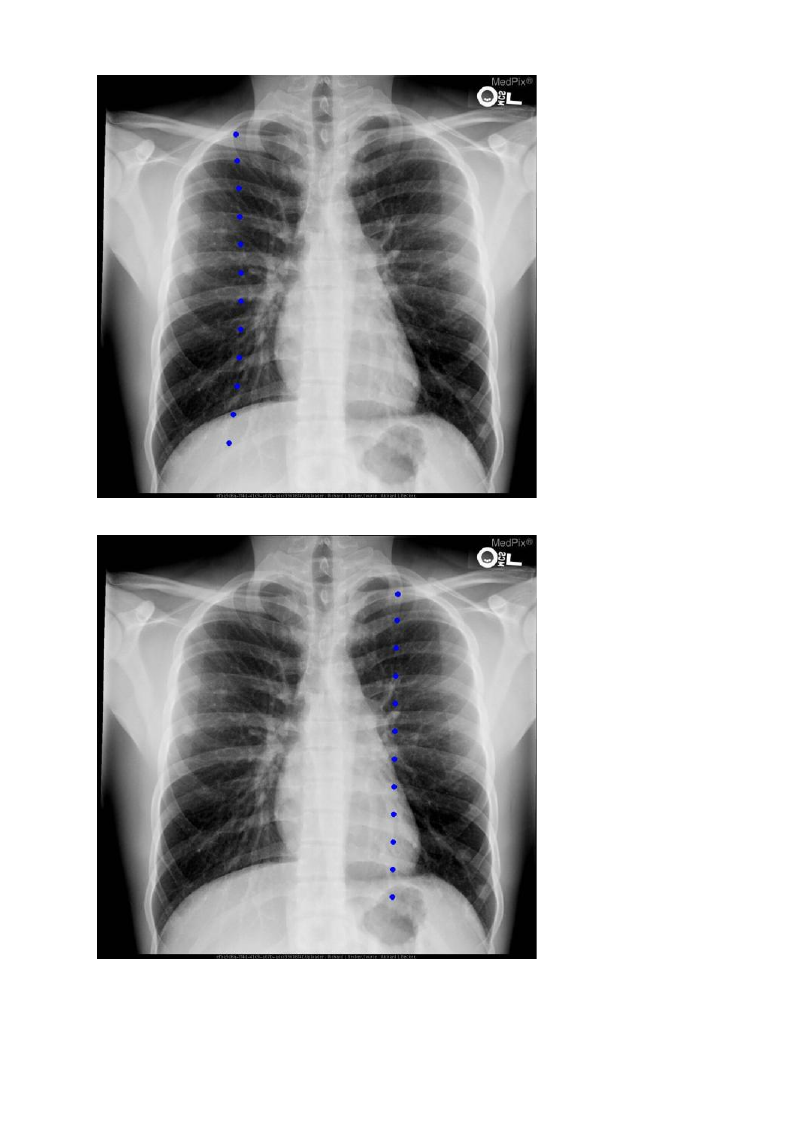} &
  {\scriptsize
  \textbf{Q:} How many ribs are present in vertical order on this chest radiograph?\par
  \vspace{0.25em}
  \textbf{Model Responses:} \texttt{<think>}~The user needs to count the number of ribs in vertical order on this chest radiograph. First, I should start counting from the right posterior ribs, since they are more clearly visible. I will mark the first rib, then proceed sequentially. Looking at the right upper lung, the first rib is at [326, 135], the second at [329, 195], the third at [333, 262], the fourth at [336, 329], the fifth at [338, 395], the sixth at [339, 461], the seventh at [340, 528], the eighth at [339, 594], the ninth at [336, 661], the tenth at [330, 728], the eleventh at [320, 795]? Wait, I might be mistaken, or I need to carefully mark each one. Now I will call the POINT tool to mark the ribs on the right side, from the first to the last visible one. First, I determine the positions of the right posterior ribs: 1st rib [315, 142], 2nd rib [318, 204], 3rd rib [321, 269], 4th rib [323, 336], 5th rib [325, 401], 6th rib [326, 468], 7th rib [326, 535], 8th rib [325, 601], 9th rib [322, 668], 10th rib [318, 735], 11th rib [309, 802], 12th rib? It may be obscured by subdiaphragmatic structures. I will call the POINT tool with imgidx 0 and mark each position for counting. \texttt{</think>}\par
  \vspace{0.25em}
  \texttt{<tool\_call>} \texttt{\{``name'': ``POINT'', ``arguments'': \{``imgidx'': 0, ``points'': ``[315, 142][318, 204][321, 269][323, 336][325, 401][326, 468][326, 535][325, 601][322, 668][318, 735][309, 802][299, 869]''\}\}} \texttt{</tool\_call>}\par
  \vspace{0.25em}
  \texttt{<tool\_response>} \texttt{\{``status'': ``success'', ``images'': [``<image>'']\}} \texttt{</tool\_response>}\par
  \vspace{0.25em}
  \texttt{<think>} Now I need to confirm the rib count on the left side to ensure bilateral symmetry. I will mark the left posterior ribs using the POINT tool. Looking at the left side, I mark 12 points from top to bottom: [682, 139][680, 201][679, 265][678, 331][677, 396][676, 461][675, 527][674, 592][673, 657][672, 722][671, 787][670, 852], so that I can count the ribs on the left side and confirm the number. \texttt{</think>}\par
  \vspace{0.25em}
  \texttt{<tool\_call>} \texttt{\{``name'': ``POINT'', ``arguments'': \{``imgidx'': 0, ``points'': ``[682, 139][680, 201][679, 265][678, 331][677, 396][676, 461][675, 527][674, 592][673, 657][672, 722][671, 787][670, 852]''\}\}} \texttt{</tool\_call>}\par
  \vspace{0.25em}
  \texttt{<tool\_response>} \texttt{\{``status'': ``success'', ``images'': [``<image>'']\}} \texttt{</tool\_response>}\par
  \vspace{0.25em}
  \texttt{<think>} I now need to determine the number of ribs in vertical order on this chest radiograph. I previously marked the ribs on both sides separately, with 12 on each side. Looking at the marked points: the right side shows [315, 142] to [299, 869] with 12 points corresponding to 12 ribs; the left side shows [682, 139] to [670, 852] with 12 points corresponding to 12 ribs. Unilateral 12 ribs, bilateral 24 ribs, the number of ribs in the vertical direction (posterior ribs) is 12 per side, consistent with normal anatomy. The conclusion is that on this chest radiograph, 12 ribs are visible in vertical (posterior) order per side, 24 ribs bilaterally, and the unilateral vertical rib count is 12. \texttt{</think>}\par
  \texttt{<answer>} Through bilateral rib marking on the chest radiograph: the right side shows 12 ribs in vertical (posterior) order at [315, 142][318, 204][321, 269][323, 336][325, 401][326, 468][326, 535][325, 601][322, 668][318, 735][309, 802][299, 869]; the left side shows 12 ribs at [682, 139][680, 201][679, 265][678, 331][677, 396][676, 461][675, 527][674, 592][673, 657][672, 722][671, 787][670, 852]. This count is consistent with normal adult rib anatomy (12 pairs, 24 ribs total). The vertical rib count per side is 12. \texttt{</answer>}} \\
  \bottomrule
  \end{tabular}
\caption{\textbf{POINT.} A case study demonstrating MIRA's systematic anatomical counting on a chest radiograph. The agent autonomously applies \texttt{POINT} to sequentially mark each rib on both the right and left sides, ensuring bilateral verification for an accurate and evidence-based rib count.}
\label{fig:case_study_point}
\end{figure}
\renewcommand{\arraystretch}{1}

\clearpage

\renewcommand{\arraystretch}{1.2}
\begin{figure}[t]
  \centering
  \begin{tabular}{m{5cm}m{11cm}}
  \toprule
  \includegraphics[trim=330 0 300 0, clip, width=7cm]{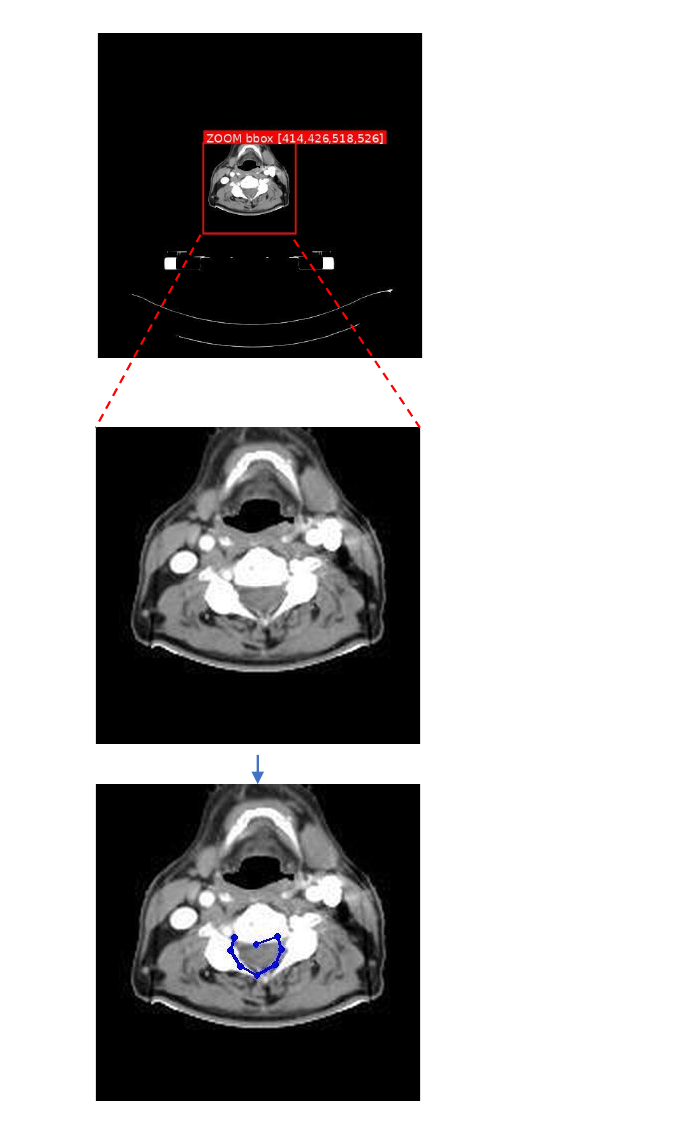} &
  {\scriptsize
  \textbf{Q:} What is the shape of this patient's spinal cord?\par
  \vspace{0.25em}
  \textbf{Model Responses:} \texttt{<think>}~The user asks about the shape of the spinal cord. First, I need to identify where the spinal cord is located on this axial CT. It should be within the spinal canal, posterior to the vertebral body. I initially consider a few possible regions, such as [438, 466, 498, 514] or [426, 432, 494, 505], but these may be too small or slightly misplaced. To inspect the spinal cord morphology more clearly, I should zoom into the broader spinal-canal region. I will call the ZOOM tool with imgidx 0, label ``spinal cord region'', bbox\_str [414, 426, 518, 526], and scale 2.0. \texttt{</think>}\par
  \vspace{0.25em}
  \texttt{<tool\_call>} \texttt{\{``name'': ``ZOOM'', ``arguments'': \{``imgidx'': 0, ``label'': ``spinal cord region'', ``bbox\_str'': ``[414, 426, 518, 526]'', ``scale'': 2.0\}\}} \texttt{</tool\_call>}\par
  \vspace{0.25em}
  \texttt{<tool\_response>} \texttt{\{``status'': ``success'', ``images'': [``<image>'']\}} \texttt{</tool\_response>}\par
  \vspace{0.25em}
  \texttt{<think>}~The region has been enlarged, and now I need to outline the structure rather than only describe it from impression. The spinal cord should be inside the spinal canal behind the vertebral body, between the posterior bony elements. Since the zoomed image is the newly returned image, I should use imgidx 1. I will mark boundary points around the cord-like structure and connect these points to make the contour easier to judge. \texttt{</think>}\par
  \vspace{0.25em}
  \texttt{<tool\_call>} \texttt{\{``name'': ``POINT'', ``arguments'': \{``imgidx'': 1, ``points'': ``[424, 482][412, 524][444, 572][496, 602][552, 568][570, 520][558, 480][492, 504]'', ``draw\_line'': true\}\}} \texttt{</tool\_call>}\par
  \vspace{0.25em}
  \texttt{<tool\_response>} \texttt{\{``status'': ``success'', ``images'': [``<image>'']\}} \texttt{</tool\_response>}\par
  \vspace{0.25em}
  \texttt{<think>} Now I need to interpret the marked contour. The blue outline does not look like a simple smooth oval. The point [496, 602] forms a posterior or inferior apex, while the upper boundary has a shallow central indentation. This makes the contour appear somewhat heart-shaped or triangular, possibly like an inverted teardrop. However, I should compare the observed morphology with normal cervical axial CT descriptions before concluding. \texttt{</think>}\par
  \vspace{0.25em}
  \texttt{<tool\_call>} \texttt{\{``name'': ``SEARCH'', ``arguments'': \{``search\_query'': ``normal cervical axial CT spinal cord morphology''\}\}} \texttt{</tool\_call>}\par
  \vspace{0.25em}
  \texttt{<tool\_response>} \texttt{\{``status'': ``success'', ``images'': [], ``text'': ``Normal spinal cord morphology on axial CT should be smooth and homogeneous, without obvious enlargement, thinning, or deformation.''\}} \texttt{</tool\_response>}\par
  \vspace{0.25em}
  \texttt{<think>} Based on the POINT outline and the reference information, the structure is not smoothly round or oval. The contour is approximately triangular, with a posteriorly directed apex and a central indentation along the anterior margin. This makes it resemble a heart shape or inverted teardrop shape. \texttt{</think>}\par
  \texttt{<answer>} The patient's spinal cord on this axial cervical CT appears roughly triangular, with the apex directed dorsally/posteriorly, giving it an inverted teardrop or heart-like shape. The ventral/anterior margin shows a noticeable indentation and the contour is not smoothly oval, suggesting possible deformation related to anterior compression. \texttt{</answer>}} \\
  \bottomrule
  \end{tabular}
\caption{\textbf{ZOOM \& POINT \& SEARCH.} A case study demonstrating MIRA's reflective visual analysis of spinal cord morphology on an axial cervical CT image. The agent first applies \texttt{ZOOM} to localize the spinal-cord region, then uses \texttt{POINT} to trace the contour, and finally invokes \texttt{SEARCH} to compare the observed morphology with normal axial CT appearance before producing a diagnostic description.}
\label{fig:case_study_zoom_point_search}
\end{figure}
\renewcommand{\arraystretch}{1}

\clearpage

\renewcommand{\arraystretch}{1.2}
\begin{figure}[t]
  \centering
  \begin{tabular}{m{5cm}m{11cm}}
  \toprule
  \includegraphics[trim=330 0 170 0, clip, width=7cm]{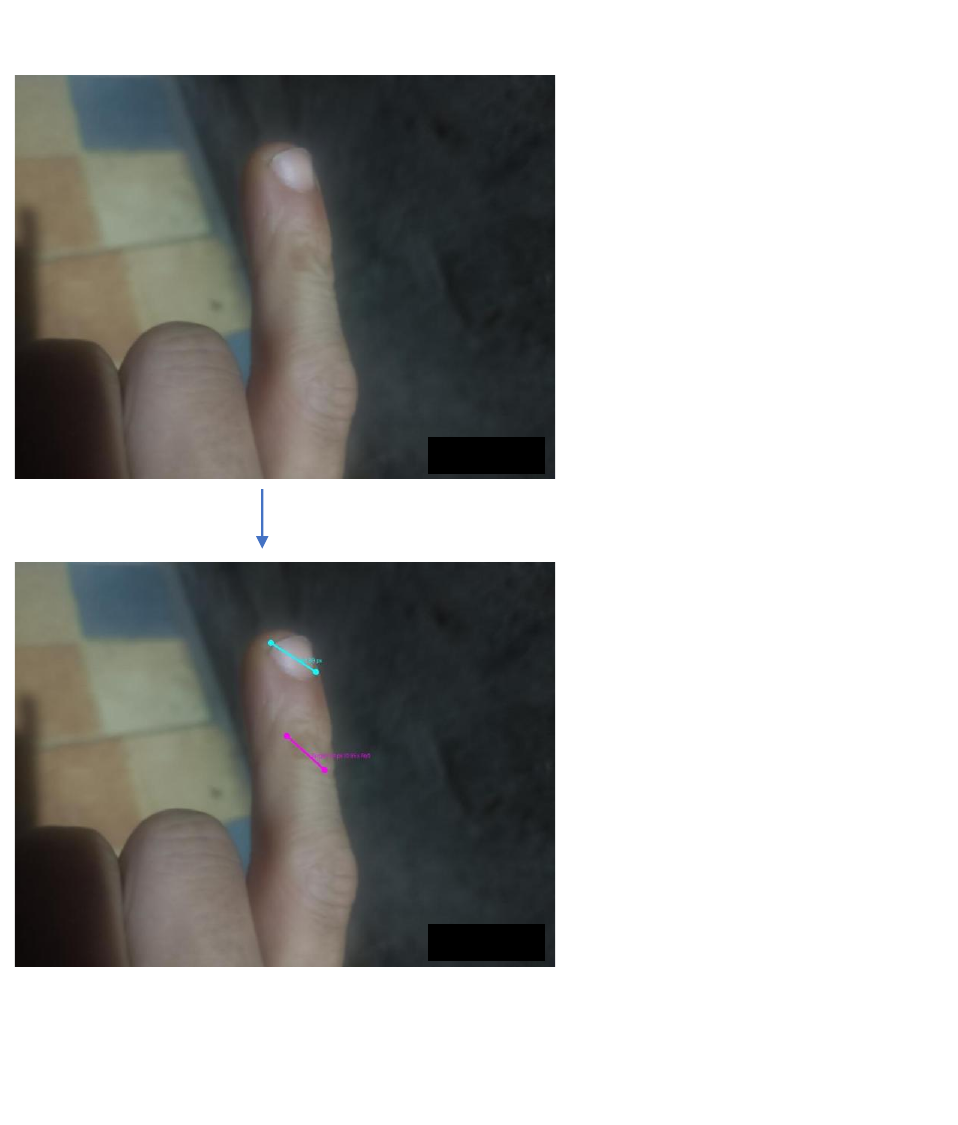} &
  {\scriptsize
  \textbf{Q:} What is the size range of the yellow pigmented lesion on the index finger in the image? Please describe it using familiar size references such as a needle tip, rice grain, or soybean.\par
  \vspace{0.25em}
  \textbf{Model Responses:} \texttt{<think>}~The user needs the size of the yellow pigmented lesion on the index finger, described using intuitive references such as a needle tip, rice grain, or soybean. First I should inspect the finger lesion carefully. The lesion is on the distal index finger near the nail, so I can use the visible fingernail as a nearby anatomical reference. I will measure the long axis of the yellow lesion, then compare it with a selected fingernail segment to convert the visual impression into an approximate object-scale description. \texttt{</think>}\par
  \vspace{0.25em}
  \texttt{<tool\_call>} \texttt{\{``name'': ``MEASURE'', ``arguments'': \{``imgidx'': 0, ``target\_points'': ``[502, 428][572, 512]'', ``reference\_points'': ``[473, 198][556, 270]''\}\}} \texttt{</tool\_call>}\par
  \vspace{0.25em}
  \texttt{<tool\_response>} \texttt{\{``status'': ``success'', ``images'': [``<image>''], ``text'': ``Reference pixel distance: 99 pixels. Target pixel distance: 94 pixels. Conclusion: the target length is 0.95 times the reference length.''\}} \texttt{</tool\_response>}\par
  \vspace{0.25em}
  \texttt{<think>}~The MEASURE result shows that the lesion's measured long axis is 94 pixels, while the fingernail reference segment is 99 pixels. Therefore, the lesion is about 0.95 times the selected fingernail segment, meaning it is nearly the same length as that reference. A typical visible index-fingernail segment is around 1 cm in scale, so the lesion's long axis is also close to 1 cm. This is much larger than a needle tip, which would be sub-millimeter to about 1 mm, and also larger than a typical rice grain, which is only several millimeters long. Its size is most consistent with a soybean-scale lesion. \texttt{</think>}\par
  \vspace{0.25em}
  \texttt{<answer>} The yellow pigmented lesion is approximately soybean-sized. Based on the measurement, its long axis is 94 pixels, while the selected fingernail reference is 99 pixels, so the lesion is about 0.95 times the reference length. Using the fingernail as a rough anatomical scale, this corresponds to roughly around 1 cm. Thus, it is clearly larger than a needle-tip-sized or rice-grain-sized lesion and is best described as about the size of a soybean. \texttt{</answer>}} \\
  \bottomrule
  \end{tabular}
\caption{\textbf{MEASURE.} A case study demonstrating MIRA's quantitative visual reasoning on a skin lesion image. The agent applies \texttt{MEASURE} to compare the lesion's long-axis length with a nearby fingernail reference, converting a visual size judgment into an interpretable estimate using familiar object-scale descriptors.}
\label{fig:case_study_measure}
\end{figure}
\renewcommand{\arraystretch}{1}

\clearpage

\renewcommand{\arraystretch}{1.2}
\begin{figure}[t]
  \centering
  \begin{tabular}{m{5cm}m{11cm}}
  \toprule
  \includegraphics[trim=250 0 250 0, clip, width=7cm]{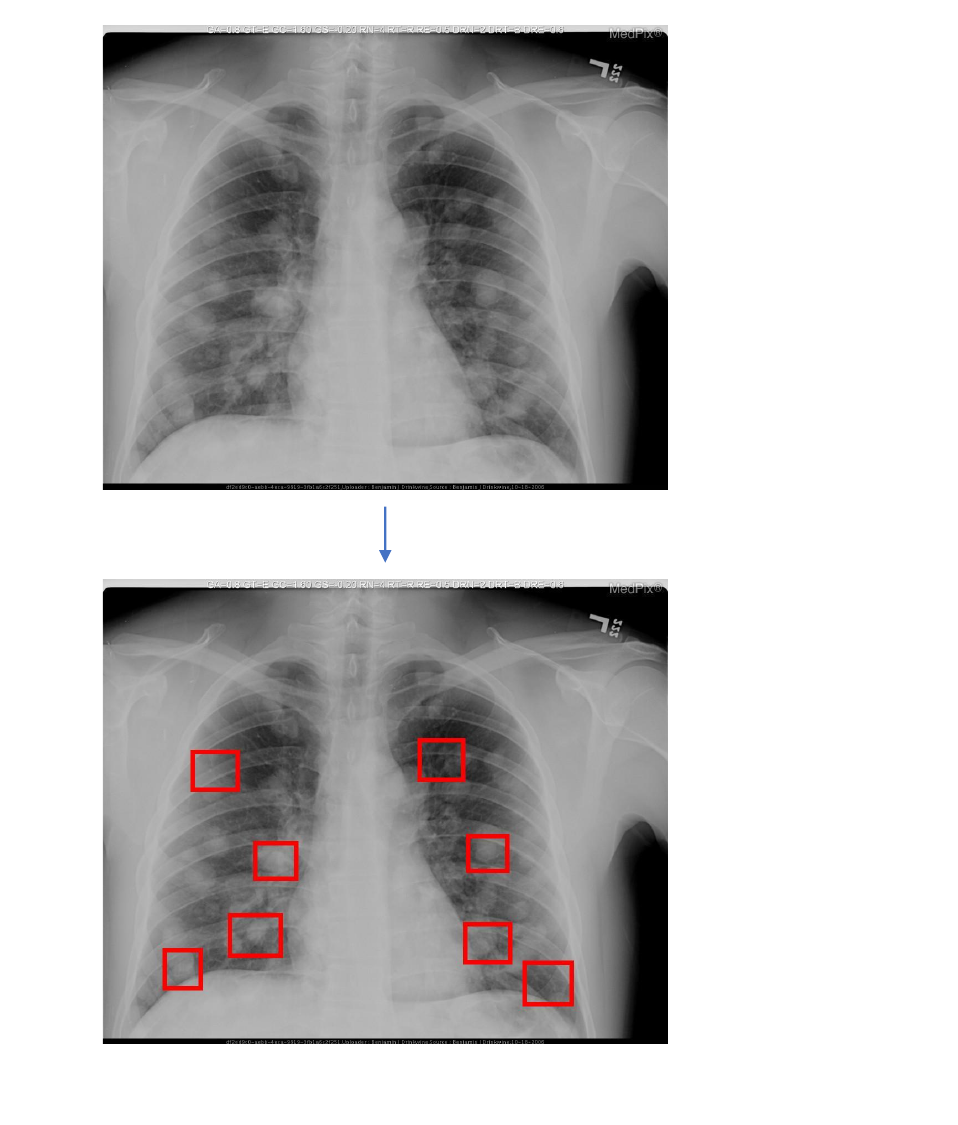} &
  {\scriptsize
  \textbf{Q:} Are there multiple circumferential opacities located throughout all lung fields?\par
  \vspace{0.25em}
  \textbf{Model Responses:} \texttt{<think>}~The user asks whether the chest radiograph contains multiple circumferential opacities throughout all lung fields. I need to inspect both lungs systematically, including the upper, middle, and lower lung zones. Because the question depends on whether there are multiple spatially distributed findings rather than a single focal lesion, I should use \texttt{GROUNDING} to mark all suspected ring-like opacities across both lung fields. \texttt{</think>}\par
  \vspace{0.25em}
  \texttt{<tool\_call>} \texttt{\{``name'': ``GROUNDING'', ``arguments'': \{``label'': ``circumferential opacities'', ``imgidx'': 0, ``bbox\_str'': ``[164, 375, 235, 446][275, 571, 338, 636][115, 800, 169, 872][230, 726, 311, 801][564, 351, 633, 424][650, 556, 711, 620][645, 745, 716, 816][750, 827, 824, 906]''\}\}} \texttt{</tool\_call>}\par
  \vspace{0.25em}
  \texttt{<tool\_response>} \texttt{\{``status'': ``success'', ``images'': [``<image>'']\}} \texttt{</tool\_response>}\par
  \vspace{0.25em}
  \texttt{<think>}~Now I should verify whether the marked regions correspond to true pulmonary ring-like opacities rather than normal structures or artifacts. The boxes cover both right and left lungs: right upper, right middle, right lower, left upper, left middle, and left lower zones. Several marked regions show round or ring-like opacities with relative central lucency. Because the grounded boxes are distributed bilaterally and span multiple lung zones, the finding is not a single localized opacity. The multi-box grounding result supports a positive answer. \texttt{</think>}\par
  \texttt{<answer>} Yes. The grounded regions show multiple circumferential or ring-like opacities distributed across both lungs, including upper, middle, and lower lung fields. Their bilateral and multi-zone distribution supports the conclusion that there are multiple circumferential opacities throughout all lung fields. \texttt{</answer>}} \\
  \bottomrule
  \end{tabular}
\caption{\textbf{GROUNDING.} A case study demonstrating MIRA's multi-target visual grounding on a chest radiograph. The agent uses a single \texttt{GROUNDING} call to mark eight suspected circumferential opacities across bilateral lung zones, distinguishing distributed pulmonary findings from a single localized lesion or simple zoom-based inspection.}
\label{fig:case_study_grounding}
\end{figure}
\renewcommand{\arraystretch}{1}

\clearpage

\renewcommand{\arraystretch}{1.2}
\begin{figure}[t]
  \centering
  \begin{tabular}{m{5cm}m{11cm}}
  \toprule
  \includegraphics[trim=250 0 250 0, clip, width=7cm]{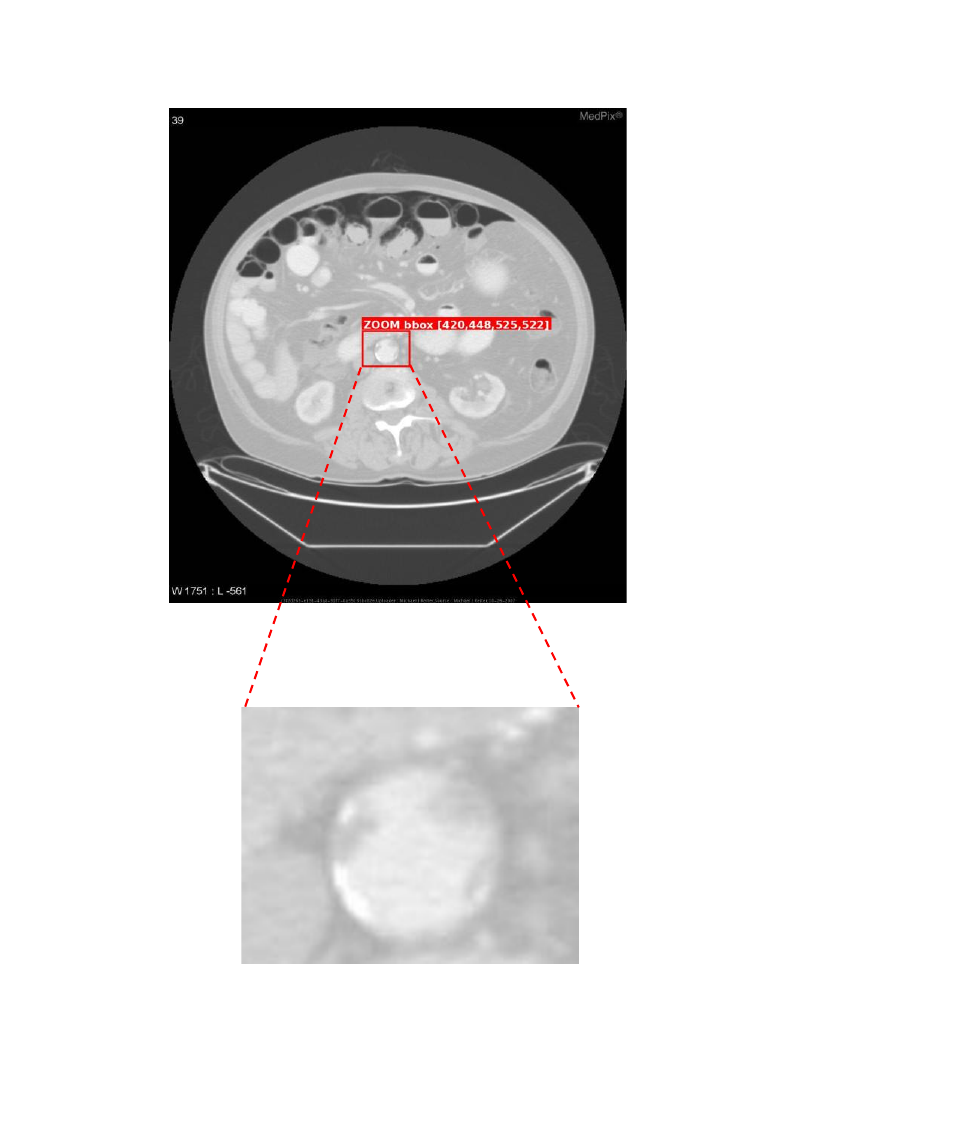} &
  {\scriptsize
  \textbf{Q:} Are there calcifications present on the abdominal aorta?\par
  \vspace{0.25em}
  \textbf{Model Responses:} \texttt{<think>}~The user asks whether calcifications are present on the abdominal aorta. A coarse inspection \textcolor[HTML]{D9822B}{does not show a confidently visible bright calcified focus}. However, this target is small and localized, so I \textcolor[HTML]{D9822B}{should not finalize the answer from a coarse view}. I \textcolor[HTML]{D9822B}{re-check the suspected abdominal-aorta region} and decide to invoke \texttt{ZOOM} for visual verification. \texttt{</think>}\par
  \vspace{0.25em}
  \texttt{<tool\_call>} \texttt{\{``name'': ``ZOOM'', ``arguments'': \{``label'': ``abdominal aorta'', ``imgidx'': 0, ``bbox\_str'': ``[420, 448, 525, 522]'', ``scale'': 2.0\}\}} \texttt{</tool\_call>}\par
  \vspace{0.25em}
  \texttt{<tool\_response>} \texttt{\{``status'': ``success'', ``images'': [``<image>'']\}} \texttt{</tool\_response>}\par
  \vspace{0.25em}
  \texttt{<think>}~After zooming into the abdominal-aorta region, I observe a focal high-density signal along the aortic wall. This directly supports the presence of calcification, so I \textcolor[HTML]{D9822B}{correct the initial uncertainty} and \textcolor[HTML]{D9822B}{ground the final answer in the zoomed evidence}. \texttt{</think>}\par
  \texttt{<answer>} Yes. The zoomed abdominal-aorta region shows a focal high-density calcified focus along the vessel wall. \texttt{</answer>}} \\
  \bottomrule
  \end{tabular}
\caption{\textbf{TEXTUAL REFLECTION.} A case study demonstrating MIRA's textual self-reflection on an abdominal CT image. The agent first identifies uncertainty in the coarse view, then decides to re-check the abdominal-aorta region before correcting the initial impression and producing an evidence-grounded answer.}
\label{fig:case_study_reflection_zoom}
\end{figure}
\renewcommand{\arraystretch}{1}

\clearpage

\renewcommand{\arraystretch}{1.2}
\begin{figure}[t]
  \centering
  \begin{tabular}{m{5cm}m{11cm}}
  \toprule
  \includegraphics[trim=250 0 250 0, clip, width=7cm]{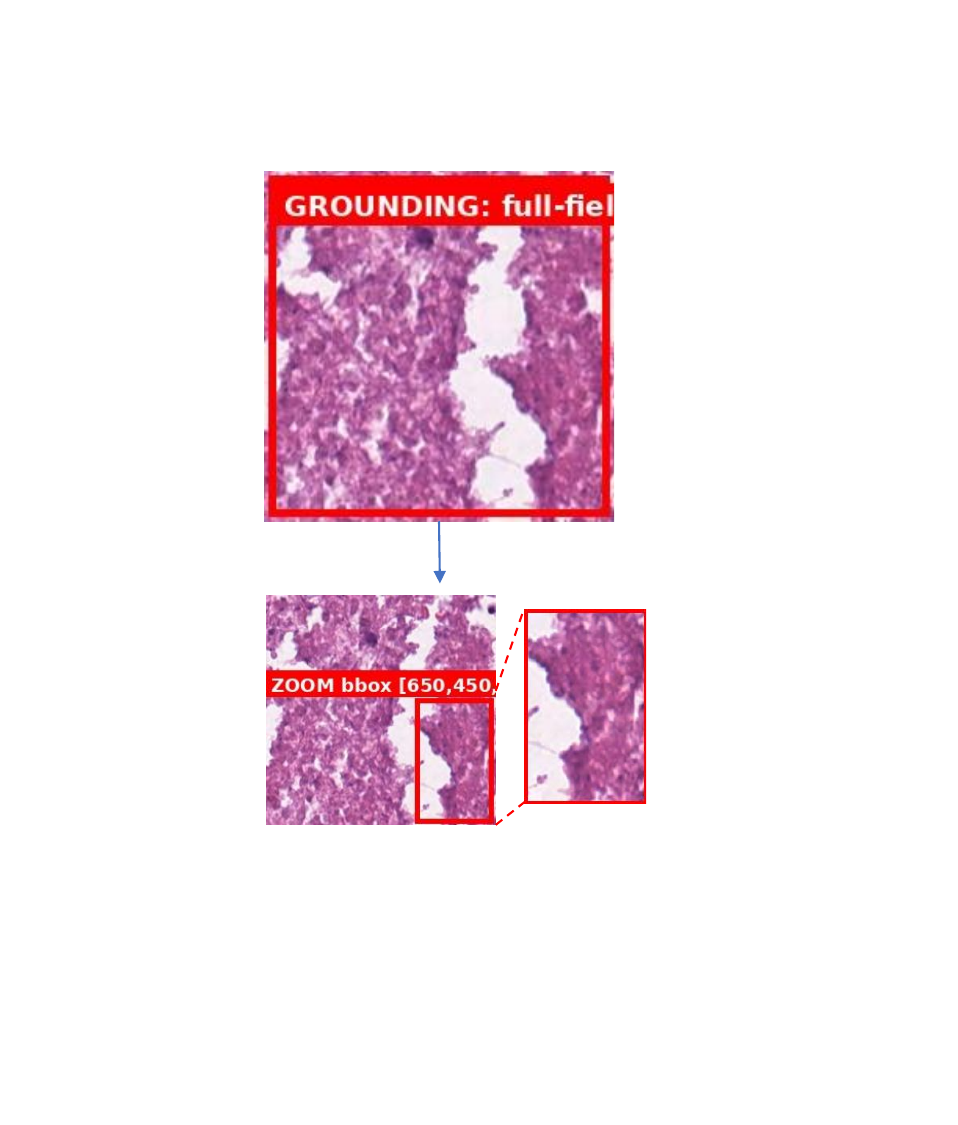} &
  {\scriptsize
  \textbf{Q:} What substance or tissue type can you see in this image?\par
  \vspace{0.25em}
  \textbf{Model Responses:} \texttt{<think>}~I need to identify the visible substance or tissue type in this histology image. I first compare the options at the global level. A mucous gland should show glandular lumens and secretory epithelium, while cartilage should show hyaline matrix and lacunae; neither pattern is obvious. The field instead contains dark-purple cellular aggregates and pale gaps, suggesting debris, but \textcolor[HTML]{D9822B}{debris versus epithelial tissue is still uncertain from the global view}. I therefore start with \texttt{GROUNDING} over the full field to inspect the overall morphology. \texttt{</think>}\par
  \vspace{0.25em}
  \texttt{<tool\_call>} \texttt{\{``name'': ``GROUNDING'', ``arguments'': \{``label'': ``full-field tissue morphology'', ``imgidx'': 0, ``bbox\_str'': ``[0, 0, 999, 999]''\}\}} \texttt{</tool\_call>}\par
  \vspace{0.25em}
  \texttt{<tool\_response>} \texttt{\{``status'': ``success'', ``images'': [``<image>'']\}} \texttt{</tool\_response>}\par
  \vspace{0.25em}
  \texttt{<think>}~The full-field grounding confirms the global distribution, but it still does not show enough local detail to distinguish debris from epithelial tissue. In particular, I need to check whether the suspicious material has organized cell boundaries and epithelial architecture, or whether it is amorphous and fragmented. Thus, \textcolor[HTML]{D9822B}{the first tool result is not sufficient for the fine tissue-type decision}, and I \textcolor[HTML]{D9822B}{refine the tool use by zooming into the suspicious lower-right region}. \texttt{</think>}\par
  \vspace{0.25em}
  \texttt{<tool\_call>} \texttt{\{``name'': ``ZOOM'', ``arguments'': \{``label'': ``suspicious debris region'', ``imgidx'': 0, ``bbox\_str'': ``[650, 450, 990, 990]'', ``scale'': 2.0\}\}} \texttt{</tool\_call>}\par
  \vspace{0.25em}
  \texttt{<tool\_response>} \texttt{\{``status'': ``success'', ``images'': [``<image>'']\}} \texttt{</tool\_response>}\par
  \vspace{0.25em}
  \texttt{<think>}~After zooming, I re-evaluate the local morphology. I still do not see glandular lumens, secretory epithelium, cartilage lacunae, or a regular epithelial sheet. The zoomed region instead appears as \textcolor[HTML]{D9822B}{irregular, fragmented, structureless material}, which supports debris rather than organized tissue. \texttt{</think>}\par
  \texttt{<answer>} D. \texttt{</answer>}} \\
  \bottomrule
  \end{tabular}
\caption{\textbf{TOOL-LEVEL REFLECTION.} A case study demonstrating MIRA's tool-level reflection on a histology image. The agent first uses \texttt{GROUNDING} for a global inspection, recognizes that the tool result is insufficient for a fine tissue-type decision, and then refines the visual evidence with \texttt{ZOOM} before answering.}
\label{fig:case_study_tool_reflection}
\end{figure}
\renewcommand{\arraystretch}{1}

\ifseparateappendixbibliography
\clearpage
\putbib[main]
\end{bibunit}
\fi

\end{document}